\documentclass{article}

\usepackage[preprint]{neurips_2026}

\usepackage[utf8]{inputenc} % allow utf-8 input
\usepackage[T1]{fontenc}    % use 8-bit T1 fonts
\usepackage[backref]{hyperref}       % hyperlinks
\hypersetup{
    colorlinks,
    linkcolor={purple},
    citecolor={blue!50!black}
}
\usepackage{url}            % simple URL typesetting
\usepackage{booktabs}       % professional-quality tables
\usepackage{amsfonts}       % blackboard math symbols
\usepackage{nicefrac}       % compact symbols for 1/2, etc.
\usepackage{microtype}      % microtypography
\usepackage{xcolor}         % colors
\usepackage[labelfont=bf]{caption}

\usepackage{tabularray}      % More adjustable Tables
\usepackage{fontawesome}     % To include lightbulbicon
\usepackage[most]{tcolorbox} % Color Boxes for findings
\usepackage{graphicx}        % For graphics
\usepackage{xspace}          % For logo styling
\usepackage{amsmath}         % For text in equations
\usepackage{amssymb}         % For method markers like triangle
\usepackage{cleveref}        % Simpler References
\usepackage{mathtools}       % Equation helpfull macros
\usepackage{subcaption}      % For the subcaptions
\usepackage{titletoc}        % Appendix table
\usepackage{multirow}        % Multirow tables
\usepackage{array}           % Fixes the 'm' column error
\usepackage{float}           % Allow in appendix figures to be under section

\tcbuselibrary{breakable, skins}

\newcommand{\mlogobig}{\raisebox{-4pt}{\includegraphics[height=19pt]{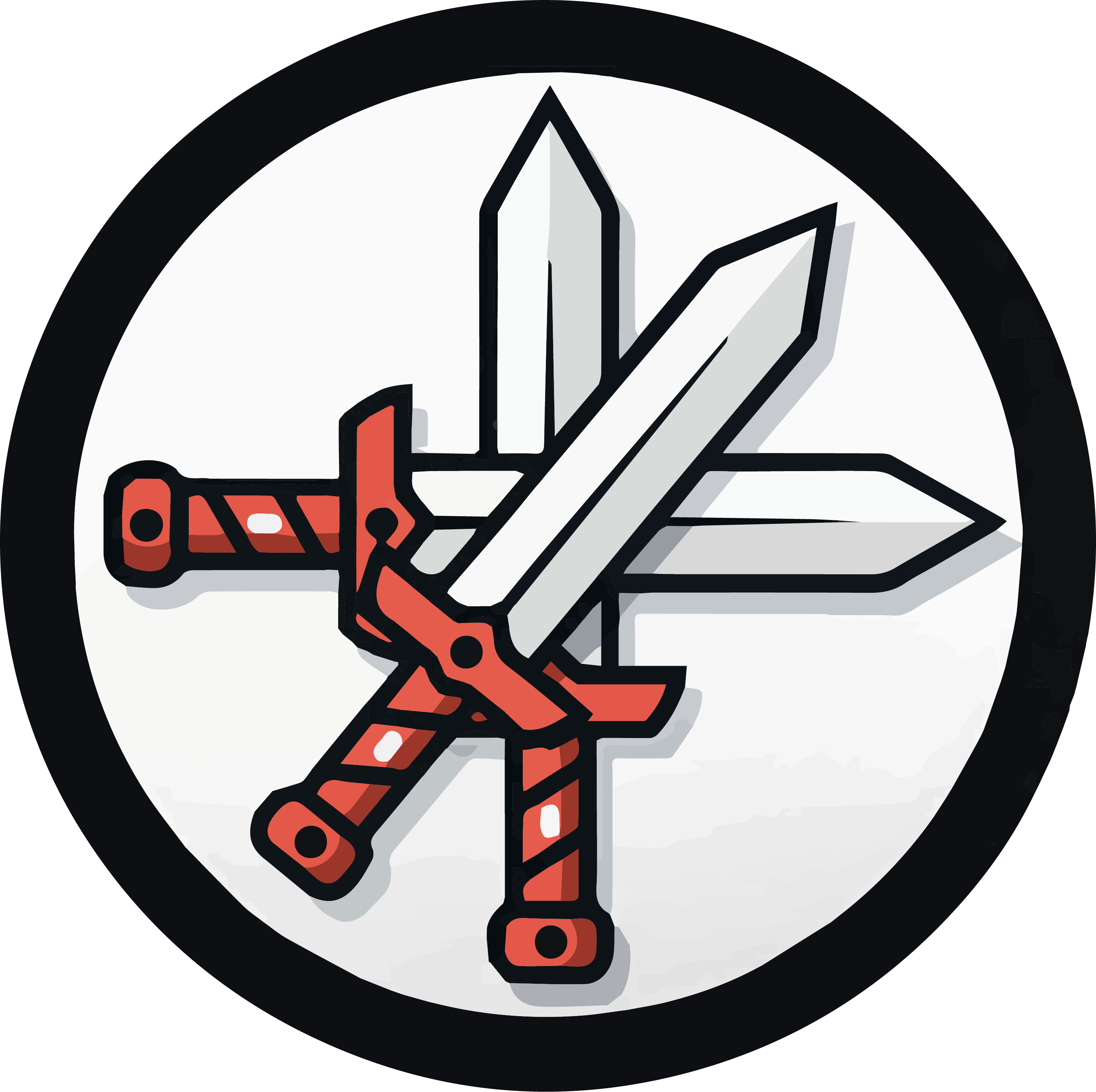}}}

\newcommand{\ourbenchmark}{\textsc{SwordBench}}
\newcommand{\ourbenchmarkx}{\textsc{SwordBench}\xspace}

\definecolor{customgreen}{RGB}{116, 154, 114}
\definecolor{lightgreen}{RGB}{240, 246, 232}
\newcommand{\lightbulbicon}{%
  \begin{tikzpicture}[baseline=-0.5ex]
    \draw[fill=white, draw=customgreen, thick] (0,0) circle (1.5ex);
    \node[scale=0.8, color=customgreen] at (0,0) {\faLightbulbO~};
  \end{tikzpicture}%
}
\newcounter{finding}
\newtcolorbox{customblockquote}{
  colframe=customgreen,
  colback=lightgreen,
  boxrule=0pt,
  leftrule=2pt, 
  left=1pt,  % Set to 0pt so the background color touches the left line
  right=3pt,
  top=4pt,
  bottom=2pt,
  arc=0pt,
  breakable,
  before skip=0.5\baselineskip,
  after skip=0.5\baselineskip,
  left skip=0pt,
  right skip=0pt,
  pad before break=0pt,
  pad after break=0pt,
  enhanced jigsaw,
  frame hidden,
   overlay={
    \draw[customgreen, line width=2pt] 
      (frame.north west) -- (frame.south west);
    \node[inner sep=0pt] at ([xshift=0pt, yshift=-1.5pt]frame.north west) {\lightbulbicon};
  },
  before upper={\refstepcounter{finding}\textbf{Finding \thefinding:}\ },
  boxsep=3pt
}
\makeatletter
\def\adl@drawiv#1#2#3{%
        \hskip.5\tabcolsep
        \xleaders#3{#2.5\@tempdimb #1{1}#2.5\@tempdimb}%
                #2\z@ plus1fil minus1fil\relax
        \hskip.5\tabcolsep}
\newcommand{\cdashlinelr}[1]{%
  \noalign{\vskip\aboverulesep
           \global\let\@dashdrawstore\adl@draw
           \global\let\adl@draw\adl@drawiv}
  \cdashline{#1}
  \noalign{\global\let\adl@draw\@dashdrawstore
           \vskip\belowrulesep}}
\makeatother

\newcommand{\Azero}{\ensuremath{\mathbb{A}_{\bar{c}}}}
\newcommand{\Aone}{\ensuremath{\mathbb{A}_c}}

\title{\ourbenchmark{}: Evaluating Orthogonality\\of Steering Image Representations}

\author{%
  Vladimir Zaigrajew \\
  Centre for Credible AI \\
  Warsaw University of Technology \\
  Warsaw, Poland \\
  \texttt{vladimir.zaigrajew.dokt@pw.edu.pl} \\
  \And
  Dawid Pludowski \\
  Centre for Credible AI \\
  Warsaw University of Technology \\
  Warsaw, Poland \\
  \texttt{dawid.pludowski.dokt@pw.edu.pl} \\
  \And
  Hubert Baniecki \\
  Centre for Credible AI \\ 
  Warsaw University of Technology\\
  University of Warsaw  \\
  Warsaw, Poland \\
  \And
  Przemyslaw Biecek \\
  Centre for Credible AI \\
  Warsaw University of Technology \\
  University of Warsaw \\
  Warsaw, Poland \\
}
\begin{document}

\maketitle

\begin{abstract}
  Steering or intervening on model representations at inference time to correct predictions is essential for AI interpretability and safety, yet existing evaluation protocols are limited to ambiguous language modeling tasks.
  To address this gap, we introduce \ourbenchmark{}, a benchmark for steering image representations of vision models across multiple backbones and concept removal tasks.
  Beyond a unified benchmarking suite, we propose new evaluation notions that uncover the second-order effects of orthogonalization among concept activation vectors for pragmatic steering.
  Specifically, \emph{cross-concept robustness} measures the stability of concept detection performance across inputs orthogonalized against alternative concepts, and \emph{collateral damage} quantifies whether steering inadvertently affects model performance on a downstream task for inputs lacking the bias.
  We find that although a linear support vector machine exhibits superior separability and orthogonality, it fails to achieve zero collateral damage, often trailing sparse autoencoders. 
  In simpler regimes, both standard baselines and optimization-based methods fail to achieve perfect steering. 
  The source code will be made available soon on GitHub.
\end{abstract}

%%%%%%%%%%%%%%%%%%%%%%%%%%%%%%%%%%%%%
\section{Introduction}
\label{sec:introduction}

Self-supervised learning achieves remarkable performance on various language~\citep{gemmateam2025gemma3}, vision~\citep{oquab2024dinov2}, and vision-language tasks~\cite{radford2021learning,zhai2023sigmoid}.
Unfortunately, models often rely on shortcut learning~\cite{li2023whac} and spurious correlations to make decisions~\cite{pahde2023reveal}, and are prone to adversarial attacks~\cite{wei2023jailbroken}, posing a key challenge for AI safety.
To this end, interpretability research develops methods that enable understanding the internal behavior of large models, or even correcting their predictions by \emph{steering}, i.e., by intervening on the model's representations at inference-time without model finetuning~\citep{subramani2022extracting,li2023inference,rimsky2024steering,qiu2024spectral,wu2025axbench}.

Steering has been extensively studied to control language model predictions at test-time without the unreliability of prompt engineering~\cite{wei2023jailbroken} or the misalignment risks of fine-tuning~\cite{betley2025emergent}, where it typically targets abstract concepts such as `honesty'~\cite{zou2023representation}. 
In this work, we make a natural connection that \emph{steering language models} practically mirrors \emph{concept removal in vision models}, where biases can be detected in the model's representation space~\cite{fel2023holistic,fel2023craft,pahde2025navigating}.
In fact, both lines of work rely on the fundamental notion of a \emph{concept activation vector}~\citep[CAV,][]{kim2018interpretability}; throughout this paper, we unify the terminology, referring to the intervention as \emph{steering} and to the vectors as CAVs.

The emerging utility of steering in alignment has spurred the development of numerous algorithms for computing CAVs, exposing a critical need for a unified benchmark to evaluate them.
Although a few such evaluation protocols have recently been proposed for steering language models~\cite{zou2023representation,bhalla2024towards}, notably \textsc{AxBench}~\citep{wu2025axbench}, there is no alternative for vision models, and they all share fundamental limitations.
Training CAVs for language models is inherently noisy because their multi-token representations are increasingly susceptible to spurious correlations~\cite{tan2024analysing}.
Moreover, evaluation metrics for generative models primarily rely on the \textit{LLM-as-a-judge}, an imperfect proxy for human assessment~\cite{wang2024large,li2025generation}.
Thus, we seek metrics that can assess CAVs without such confounding factors.

\begin{figure}[t]
    \centering
    \includegraphics[width=\textwidth]{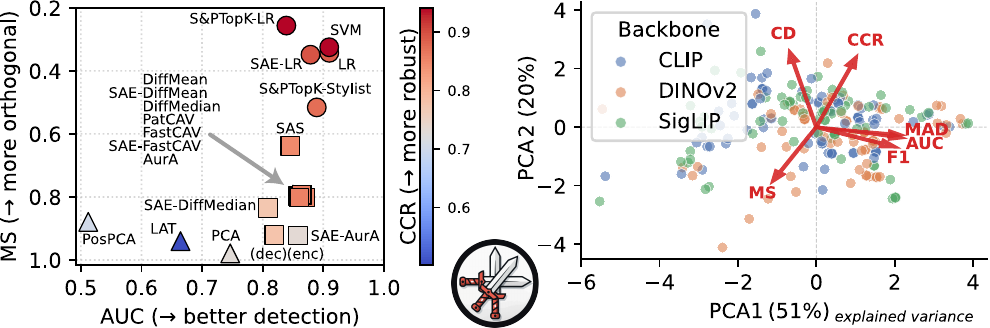}
    \caption{\textbf{\ourbenchmarkx{} multi-dataset concept activation vectors evaluation.}
    On the \emph{left}, we report vector metrics for SigLIP image representations CAVs on CelebA (averaged over 39 concepts). Improved disentanglement (lower MS) correlates with higher robustness (CCR), with methods forming three clusters ($\bullet,\blacksquare,\blacktriangle$) along the orthogonality and detection axes. On the \emph{right}, we compute a PCA biplot on ISIC using all evaluated backbones to visualize metric correlations, where arrows indicate metric directions. While robustness and orthogonality (CD) appear uncorrelated on ISIC, they show correlation in other datasets (Appendix~\ref{app:metric_analysis}).}
    \label{fig:main}
\end{figure}

\textbf{This work: A multidimensional evaluation of steering image representations.} 
We introduce \ourbenchmark{}, a benchmark for evaluating orthogonality, robustness and potential damage of steering image representations at scale (\Cref{fig:main}). 
\ourbenchmarkx{} shifts the domain from language models to vision transformers, focusing on steering modern self-supervised learning architectures such as CLIP~\citep{radford2021learning}, SigLIP \citep{zhai2023sigmoid} and DINOv2~\citep{oquab2024dinov2}.
Relying on \emph{final image representations} used in downstream tasks effectively mitigates the challenge of \textit{``How to steer across layers and tokens?''} and avoids the Hydra effect~\cite{mcgrath2023hydra} from benchmarking steering in language models.
Furthermore, \ourbenchmarkx{} offers ground-truth concept labels across four datasets for bias-removal: CelebA~\citep{liu2015deep}, ISIC~\citep{codella2019skin}, Waterbirds~\citep{sagawa2020distributionally} and Counteranimals~\citep{wang2024sober}; as well as synthetically concept-infused tasks, such as ImageNet-W~\citep{li2023whac} and ImageNet-C~\citep{hendrycks2019benchmarking}, which we adapt specifically to evaluate steering.

\textbf{Contributions.}
Beyond a unified collection of vision models, labeled datasets, standard CAV extraction methods, and performance metrics, we propose two novel evaluation notions:
\textbf{(1)~Cross-concept robustness} measures the stability of concept detection performance for inputs orthogonalized against alternative concepts apparent in data. 
\textbf{(2)~Collateral damage} measures whether concept erasing inadvertently affects model performance on a downstream task for inputs without the concept.
Combined, these two metrics uncover the undesirable \emph{second-order effects} of orthogonalization among CAVs for effective steering.
\textbf{(3)~Findings.} 
We use \ourbenchmarkx{} to deliver several insightful findings, including evidence that some linear methods achieve separability on par with nonlinear probes, and that standard classification metrics are insufficient proxies for robustness, concept entanglement, or collateral damage.
Additionally, we propose a \emph{controllable setup} for assessing steering effectiveness, demonstrating that while some methods achieve perfect steering, others cannot.
%
%%%%%%%%%%%%%%%%%%%%%%%%%%%%%%%%%%%%%
\section{Related Work}
\label{sec:related-work}

\textbf{Concept activation vectors.}
Concept-based explanations gained significant prominence with the formalism of the concept activation vector~\cite{kim2018interpretability}. 
This framework builds upon the \textit{linear representation hypothesis}~\cite{nanda2023emergent,park2024the}, which posits that semantic concepts can be encoded as linear directions within the intermediate model activations. 
While typically these directions are identified using supervised linear probes~\cite{alain2017understanding}, various alternative extraction methods have since been proposed~\cite{fel2023holistic,pahde2025navigating,schmalwasser2025fastcav}. 
Although the universality of this hypothesis is debated~\cite{crabbe2022concept,engels2025not}, it remains the foundational assumption for steering across vision~\cite{ghorbani2019towards, obadic2024contrastive} and language~\cite{clark2019does,belrose2023leace}. 
We empirically verify this hypothesis by comparing the performance of linear CAVs with that of nonlinear probes.

\textbf{Superposition and sparse autoencoders.}
A critical challenge of interpretability is \textit{superposition}~\cite{elhage2022toy} and \textit{polysemanticity}~\cite{bricken2023monosemanticity,kopf2025capturing}, in which individual neurons in representations simultaneously activate for multiple, unrelated concepts. 
Such an entanglement suggests that standard linear steering may inadvertently affect unrelated concepts. 
To address this, recent research on sparse autoencoders~\cite[SAEs][]{adly2024scaling,gao2024scaling,zaigrajew2025interpreting,pach2025sparse} aims to decompose dense representations into \emph{monosemantic}, sparse features without supervision. 
While, in theory, these disentangled features should yield `cleaner' CAVs than simple linear probes, empirical evidence suggests this is not often the case~\cite{bhalla2024towards,kantamneni2025are}. 
\ourbenchmarkx{} evaluates a range of SAE-based steering methods across both: standard CAVs~\citep[e.g. FastCAV,][]{schmalwasser2025fastcav} applied to SAE latent space, as well as recent SAE-specific algorithms~\citep{bayat2025steering} such as S\&PTopK~\cite{buarbualau2025rethinking}.

\textbf{Steering and bias mitigation.}
While CAVs began as tools for explanation, the field has shifted toward interventional methods~\citep{geiger2025causal}. 
In this paradigm, vectors are used interventionally, i.e., added to or subtracted from activation, to \textit{causally} alter the model's behavior~\cite{subramani2022extracting,zou2023representation,cywinski2025saeuron}. 
In the language domain, steering is widely used to manipulate behavioral outputs, such as refusal rates or writing style~\cite{o2024steering,rimsky2024steering,venhoff2025understanding,hedstrom2025tosteer,wu2025axbench}.
However, in the vision domain, a critical application is \textit{bias removal}~\cite{dreyer2024hope,bareeva2024reactive,pahde2025navigating,erogullari2025post}, where sensitive attributes, e.g,. race, are suppressed to improve the model. 
\ourbenchmarkx{} unifies these perspectives by evaluating the bias removal task, which is equivalent to the \textit{concept erasure} steering task used in the language domain~\cite{belrose2023leace}.

%%%%%%%%%%%%%%%%%%%%%%%%%%%%%%%%%%%%%
\section{\texorpdfstring{\mlogobig\xspace \textsc{\ourbenchmark{}}}{SwordBench}}
\label{sec:swordbench}

The ultimate goal of \ourbenchmarkx{} is to systemize the evaluation of steering image representations.
Moreover, \ourbenchmarkx{} shifts the focus from the often noisy, multi-token setting of language models to the more controlled benchmarking environment of vision models.
In what follows, we formalize the problem setup in \Cref{sec:methodology}, introduce benchmarked CAV extraction methods in Sec.~\ref{sec:methods}, define evaluation metrics in Sec.~\ref{sec:evaluation}, and detail the datasets used for evaluation in Sec.~\ref{sec:datasets}. 
Finally, we present the key results from our comprehensive analysis in Sec.~\ref{sec:results}, while several minor findings are deferred to the Appendix. 
For clarity, a summary of notation is provided in Appendix~\ref{app:notation}.

%%%%%%%%%%%%%%%%%%%%%%%%%%%%%%%%%%%%%
\subsection{Problem Formulation}
\label{sec:methodology}

\textbf{The global representation advantage.}
Self-supervised learning vision models (e.g., CLIP, DINOv2, SigLIP) are explicitly trained to compress the input into a single, global representation with the `[CLS]' token or mean-pooled last-layer tokens~\cite{radford2021learning, oquab2024dinov2, zhai2023sigmoid}.
This is unlike language models, where semantic information is distributed across a sequence of tokens, and identifying the optimal intervention target is often ambiguous. 
Such a structural clarity eliminates the uncertainty of selecting \textit{which} token represents the concept or \textit{where} in the sequence to intervene, thereby avoiding the potential \emph{``Hydra effect''} occurring when steering intermediate transformer layers~\cite{mcgrath2023hydra}.
By isolating the representation, \ourbenchmarkx{} allows for a rigorous assessment of the vector itself, effectively \emph{decoupling} the quality of the CAV from the complexity of circuit discovery. 
Importantly, our goal in this work is not to maximize concept discovery within model representation, but to establish a fair and standardized framework for evaluating steering methods.

\textbf{Rethinking evaluation.}
Evaluating \textit{bias removal} on natural data is inherently challenging due to the absence of ground-truth baselines and the lack of counterfactual  samples (with and without the concept). Resonating with the Model Science responsible benchmarking principles~\cite{modelscience2025}, which prioritize reproducible evaluation, \ourbenchmarkx{} evaluates removal success exclusively on controlled data using synthetic counterfactuals to ensure ground truth validity. 
However, to verify that these vectors generalize and are not merely artifacts of synthetic settings, we evaluate steering on popular non-synthetic datasets along two axes: 
(1) \emph{detection accuracy}, measuring how accurately the projected CAV identifies the presence of the target concept, and
(2) \emph{steering specificity}, showing the ability of CAV to erase the concept without \emph{degrading} performance on concept absent samples, i.e. causing \emph{collateral damage}.

\textbf{Concept activation vectors.}
Formally, let $\phi$ be a model mapping an input $\mathbf{x}$ to a hidden representation $h \in \mathbb{R}^d$, such that $h = \phi(\mathbf{x})$. 
Within this latent space, a \emph{unit norm} concept activation vector~\citep[CAV,][]{kim2018interpretability}, denoted as $\mathbf{v}_c \in \mathbb{R}^d$, represents the linear direction of a human-interpretable concept~$c$. 
Similarly to linear probes~\cite{alain2017understanding}, $\mathbf{v}_c$ can be derived from a binary dataset $\mathcal{D}_c = \{(h_k, y_k)\}_{k=1}^{2N}$. $\mathcal{D}_c$ is constructed from two (typically) balanced populations of size $N$: the concept-positive set $\Aone = \{h \mid y=1\}$ and the concept-negative set $\Azero = \{h \mid y=0\}$. 
$N$ negative samples are drawn from counterfactual pairs for synthetic data or from random images lacking the concept. 
Guided by our stability analysis (cf. ablations in \Cref{sec:results}), we set $N=500$ to ensure convergence, unless constrained by data scarcity.

\textbf{Model setup.}
We conduct experiments on three frozen vision backbones $\phi$: DINOv2 (ViT-B/14), CLIP (ViT-L/14), and SigLIP (ViT-B/16). 
We focus exclusively on the \textit{final-layer representations}, which compress global semantic information and serve as input to the downstream tasks; in our case, model classifiers $\psi$. 
For all predictive tasks, $\psi$ is a lightweight MLP (1--2 layers) trained on the frozen backbone. 
Model training details are provided in Appendix~\ref{app:downstream_model_training}.

\textbf{Downstream steering.}
The core of \ourbenchmarkx{} is evaluating the causal effect of interventions, specifically concept removal, on downstream tasks.
We measure this by quantifying the performance shift between the original $\psi(h)$ and steered $\psi(\bar{h})$ representations on a held-out validation set. 
We implement concept erasure by projecting out the component of the representation parallel to the CAV $\mathbf{v}_c$, effectively flattening the concept direction through \textbf{orthogonalization}: $\bar{h} = h - (\mathbf{v}_c^T h) \mathbf{v}_c$.
Crucially, we apply this intervention to samples \emph{lacking} the concept and to counterfactual pairs, which allows us to assess both the \emph{side effects} and the \emph{effectiveness} of steering. 
Ideally, steering should not alter performance on samples where the concept is not present, while for counterfactuals, the steered representation should match the performance of the corresponding natural concept absent samples.
We justify the use of orthogonalization as the steering approach for benchmarking in Appendix~\ref{app:steering_intervention}. 

%%%%%%%%%%%%%%%%%%%%%%%%%%%%%%%%%%%%%
\subsection{Concept Activation Vector (CAV) Extraction Methods}
\label{sec:methods}

We benchmark a diverse suite of CAV extraction methods, categorized into (\textbf{A}) \emph{standard linear} approaches and (\textbf{B}) \emph{sparse autoencoders}, some of which were benchmarked in \textsc{AxBench}~\cite{wu2025axbench}. 
In addition to CAVs, we include (\textbf{C}) \emph{nonlinear probes} to empirically validate the linear representation hypothesis and quantify the limitations of linear CAVs. 
In Appendix~\ref{app:cav_training_details}, we provide detailed hyperparameter setups for all the methods.

% \subsubsection{Standard Linear Methods}
\textbf{(A) Standard Linear Methods.}

Difference-in-Means \citep[\textbf{DiffMean},][]{li2023inference,marks2024geometry} is a foundational baseline in research on steering, which assumes the concept direction connects the centroids of the positive and negative class distributions: 
$\mathbf{v}_{\text{DiffMean}} \coloneqq \frac{1}{|\Aone|} \sum_{h \in \Aone} h - \frac{1}{|\Azero|} \sum_{h \in \Azero} h$.
Difference-in-Medians (\textbf{DiffMedian}) is a robust variant of DiffMean designed to mitigate the effect of outliers. 
It computes the element-wise median difference: 
$\mathbf{v}_{\text{DiffMedian}} \coloneqq \text{median}(\Aone) - \text{median}(\Azero)$.
We further train a linear Support Vector Machine (\textbf{SVM}) with $L_2$ regularization to find the maximum-margin hyperplane separating sets $\Aone$ and $\Azero$. 
The CAV $\mathbf{v}_{\text{SVM}}$ is the normal vector to this hyperplane.
Similarly, we train a Logistic Regression (\textbf{LR}) model with $L_2$ regularization, defining $\mathbf{v}_{\text{LR}}$ as the model's coefficient vector.

Fast Concept Activation Vectors \citep[\textbf{FastCAV},][]{schmalwasser2025fastcav} overcomes the computational costs of iterative training. 
It computes the direction by centering the positive activations relative to the global mean embedding ($\mu_{\text{global}}$) of inputs from the combined sets $\Aone \cup \Azero$: $\mathbf{v}_{\text{FastCAV}} \coloneqq \frac{1}{|\Aone|} \sum_{h \in \Aone} (h - \mu_{\text{global}})$.
Conversely, Pattern-Based CAV \citep[\textbf{PatCAV},][]{pahde2025navigating} addresses the directional divergence where classifier weights filter rather than represent the signal, recovering the concept direction directly. 
Instead of interpreting probe weights, it computes the signal pattern via a closed-form regression solution~\cite{haufe2014interpretation}: $\mathbf{v}_{\text{PatCAV}} \coloneqq \frac{\text{Cov}(h, y)}{\text{Var}(y)}$, where $y$ denotes the binary concept labels.

We evaluate two specific variants of Principal Component Analysis:
\textbf{PCA} computes the first principal component of the combined dataset $\Aone \cup \Azero$, capturing the dominant axis of variation across the entire distribution.
\textbf{PosPCA} computes the first principal component of only the positive set $\Aone$, isolating the variance structure specific to the concept expression.
Linear Artificial Tomography \citep[\textbf{LAT},][]{zou2023representation} extracts a latent direction by analyzing pairwise activation differences. 
We construct normalized difference vectors $\delta_{ij} = \frac{(h_i - h_j)}{\|h_i - h_j\|}$ for $N$ pairs such that $h_i \in \Aone$ and $h_j \in \Azero$. 
We strictly pair counterfactuals for synthetic datasets, whereas we use random pairing for real datasets.
Vector $\mathbf{v}_{\text{LAT}}$ is the first principal component of the difference matrix $\Delta \in \mathbb{R}^{N \times d}$, formed by stacking the vectors $\delta_{ij}$ as rows.

AUROC Adaptation \citep[\textbf{AurA},][]{suau2024whispering} constructs the CAV by first calculating the AUC of $h^{(j)}$ on $\Aone$ and~$\Azero$, treating each dimension $j \in \{1, \dots, d\}$ of the represetion $h$ as an individual component.
The $\mathbf{v}_{\text{AurA}}$ is then formed by zeroing components where $\text{AUC}(h^{(j)}) \le 0.5$ and assigning the remaining weights according to the Gini coefficient: $v_{\text{AurA}}^{(j)} \coloneqq 2 \cdot (\text{AUC}(h^{(j)}) - 0.5)$.

%%%%%%
% \subsubsection{SAE-Based Methods}
\textbf{(B) SAE-based Methods.}
SAEs disentangle polysemantic representations into monosemantic neurons. 
An SAE consists of an encoder $z = \text{ReLU}(W_{\text{enc}}h + b_{\text{enc}})$ resulting in sparse features $z \in \mathbb{R}^{d_{\text{SAE}}}$, and a decoder $\hat{h} = W_{\text{dec}}z$. 
We train SAEs for each vision backbone using TopK regularization~\cite{gao2024scaling} with $k=64$ and expansion factor $32$. 
Training details are provided in Appendix~\ref{app:sae_training}.

We implement \textbf{SAE-DiffMean/Median/FastCAV} by applying statistical aggregations, such as DiffMean, DiffMedian, FastCAV, directly on the sparse latent representations $z$ to obtain a latent direction $\mathbf{w}_z$. 
The final $\mathbf{v}_{\text{SAE}}$ is obtained by decoding this direction to the $h$ space: $\mathbf{v}_{\text{SAE}} \coloneqq W_{\text{dec}} \mathbf{w}_z$.
Sparse Activation Steering \citep[\textbf{SAS},][]{bayat2025steering} applies density-based filtering on both $\Aone$ and $\Azero$ to mitigate noise from rare or spurious features. 
We select relevant SAE neurons based on activation density and optimize the threshold via cross-validation to maximize AUC. 
We also exclude neurons active in both sets, assuming that shared features represent context and not the concept. 
Vector $\mathbf{v}_{\text{SAS}}$ is obtained via SAE-DiffMean as mentioned above, but only on the remaining neurons from~$z$.

For SAE Probe (\textbf{SAE-LR}), we train a logistic regression classifier (similar to LR) on the sparse representation $z$, using $L_1$ regularization instead to encourage sparsity. 
The learned weight vector $\mathbf{w}_z$ is projected back via the decoder $\mathbf{v}_{\text{SAE-LR}} \coloneqq W_{\text{dec}} \mathbf{w}_z$ as in previous SAE methods.

Selection and Projection TopK \citep[\textbf{S\&PTopK},][]{buarbualau2025rethinking} employs a two-stage CAV extraction process. 
First, we identify candidate neurons by training an LR with an $L_1$ penalty or by applying the Stylist~\cite{smeu2025robust} ranking method, selecting the indices $\mathcal{S}$ of the Top-$K$ neurons based on their corresponding LR coefficients (fixed at $K=16$ as originally proposed). 
Second, we re-train an LR with $L_2$ penalty exclusively on this subset to obtain refined coefficients~$\mathbf{w}_{\mathcal{S}}$.
Uniquely, the final CAV is formed by projecting these coefficients using the rows of the \textit{encoder} matrix: 
$\mathbf{v}_{\text{S\&PTopK}} \coloneqq \sum_{k \in \mathcal{S}} \mathbf{w}_{\mathcal{S}}^{(k)} \cdot W_{\text{enc}}[k]$.
 Finally, leveraging SAE's disentangled representation and monosemantic neurons, we adapt the AurA to the SAE latent space $z$ and call it \textbf{SAE-AurA}. 
After computing the SAE AurA vector $\mathbf{w}_{\text{SAE-AurA}}$, we decode it back to the $h$ space using either the decoder ($W_{\text{dec}}$) or the encoder ($W_{\text{enc}}$) weights, giving two approaches: 
$\mathbf{v}_{\text{SAE-AurA}} \coloneqq W \mathbf{w}_{\text{SAE-AurA}}$, where $W \in \{W_{\text{dec}},W_{\text{enc}}^T\}$.

% \subsubsection{Nonlinear Probes}
\textbf{(C) Nonlinear Probes.}
To establish an upper bound on concept detection and test the limits of the linear hypothesis, we evaluate two nonlinear classifiers. 
We use a support vector classifier with a radial basis function kernel (\textbf{SVM-RBF}) and \textbf{TabPFN}~\cite{hollmann2025accurate}, a state-of-the-art pre-trained transformer for tabular data. 
As a hyperparameter-free AutoML solution, it serves as a robust and reliable baseline.

%%%%%%%%%%%%%%%%%%
\subsection{Evaluation Metrics}
\label{sec:evaluation}
In \ourbenchmark{}, we decompose the evaluation into two distinct phases: 
(\textbf{A}) \textit{vector analysis}, assessing the intrinsic fidelity of the extracted CAVs, and 
(\textbf{B}) \textit{steering analysis}, measuring the causal impact of interventions.

% \subsubsection{Vector Metrics}
\textbf{(A) Vector Metrics} validate CAV accuracy via projection-based detection and concept entanglement.

Concept Detection (\textbf{AUC}) quantifies the CAV's ability to linearly separate the concept by computing the Area Under the ROC Curve using projection scores $s(h) = \mathbf{v}_c^T h$. 
For a concept $c$, this is defined as: $
    \text{AUC}(\mathbf{v}_c) \coloneqq \frac{1}{|\Azero| |\Aone|} \sum_{h_{\bar{c}} \in \Azero} \sum_{h_c \in \Aone} \mathbb{I}[ \mathbf{v}_c^T h_c > \mathbf{v}_c^T h_{\bar{c}} ],
$\\
where $\mathbb{I}[\cdot]$ is the indicator function. A higher AUC reflects better separation of concept samples.

Maximum Similarity (\textbf{MS}) detects if CAVs represent distinct semantic directions by measuring the highest pairwise cosine similarity between all concept vectors in a dataset for a given extraction method. 
For a set of unit norm concept vectors $\{\mathbf{v}_{1}, \dots, \mathbf{v}_{k}\}$, the metric for concept $c$ is:
\begin{equation}
    \text{MS}(\mathbf{v}_c) \coloneqq \max_{j \neq c} \mathbf{v}_c^T \mathbf{v}_j.
\end{equation}%
Lower values indicate better disentanglement from other concepts, mitigating untargeted steering.
 
Because MS ignores functional interference, we introduce a proxy metric to preemptively estimate robustness to collateral damage---Cross-Concept Robustness (\textbf{CCR}). 
We measure the stability of the target concept $c$ when removing any other concept $j$. 
First, we compute the representation $h_{c\perp j} = h_c - (\mathbf{v}_j^T h_c) \mathbf{v}_j$ orthogonalized with respect to a non-target $\mathbf{v}_j$. Next,
CCR is defined as the minimum normalized AUC of target $c$ retained after removing any concept $j$:
\begin{equation}
    \text{CCR}(\mathbf{v}_c) \coloneqq \min_{j \neq c} \frac{\text{AUC}(\mathbf{v}_c \mid h_{c\perp j})}{\text{AUC}(\mathbf{v}_c \mid h_c)}.
\end{equation}
High $\text{CCR}$ values indicate conceptual independence for $\mathbf{v}_c$, while lower scores reflect a dependency on other concepts. 
In future evaluations, this metric may serve as a proxy for multi-concept steering.

We also measure the Mean Activation Difference~\citep[\textbf{MAD},][]{kopf2024cosy} as detailed in Appendix~\ref{app:mad_metric}.
However, we omit reporting this metric due to its high correlation with AUC, a relationship shown in \Cref{fig:main} and further discussed in Appendix~\ref{app:metric_analysis}.

% \subsubsection{Steering Metrics}
\textbf{(B) Steering Metrics} quantify debiasing success and damage on model performance on the test set. 

Downstream Concept Detection (\textbf{F1 Score}) measures the CAVs concept detection performance on the downstream datasets via the F1 score: $
\text{F1}(y, \hat{y}) \coloneq \frac{2\sum^N_{i=1}\mathbb{I}[y_i=1,\hat{y}_i=1]}{\sum^N_{i=1}\mathbb{I}[y_i=1]+\sum^N_{i=1}\mathbb{I}[\hat{y}_i=1]},
$
where $\hat{y}_i$ is a binary prediction thresholded using Youden's J statistic on the training set.

We posit that a perfect concept-removal method should have zero effect on samples that do not possess the concept and propose Collateral Damage (\textbf{CD}) to quantify this. 
For the concept-absent subset ($\Azero$), we measure the drop in downstream accuracy between original and steered representations:
\begin{equation}
    \text{CD}(\mathbf{v}_c) \coloneqq {\text{Acc}\big(\psi(h) \!\mid\! h \in \Azero\big) - \text{Acc}\left(\psi(\bar{h}) \!\mid\! h \in \Azero\right)}.
\end{equation}
Ideally, $\text{CD} \approx 0$. 
A large value indicates that the CAV is entangled with other task-relevant features.

Furthermore, for synthetic datasets with exact counterfactual pairs (clean $h \in \Azero$, infused $h_{\text{concept}} \in \Aone$), we measure Steering Disparity (\textbf{SD}), which indicates whether removing the concept from $h_{\text{concept}}$ recovers the prediction of the clean counterpart:
\begin{equation}
    \text{SD}(\mathbf{v}_c) \coloneqq \frac{\text{Acc}\big(\psi(h) \!\mid\! h \in \Azero\big) - \text{Acc}\big(\psi(\bar{h}) \!\mid\! h \in \Aone\big)}{\text{Acc}\big(\psi(h) \!\mid\! h \in \Azero\big) - \text{Acc}\big(\psi(h) \!\mid\! h \in \Aone\big)}.
\end{equation}
% where $\text{Acc}(\cdot)_{\mathbb{A}}$ is an accuracy measured on samples from~$\mathbb{A}$. 
Ideally, $\text{SD} \approx 0$, indicating that steering has successfully inverted the concept addition. 

%%%%%%%%%%%%%%%%%%%%%
\subsection{Datasets}
\label{sec:datasets}
Steering success can be conflated with specific dataset properties. 
To analyze these factors, we adopt a robust evaluation strategy using 4 real-world bias-removal datasets and one synthetic control dataset. 
Full details on dataset splits, concept distributions, and sample visualizations are deferred to the Appendix~\ref{app:dataset_distribution}. 
To ensure reliable generalization, we train all CAVs on balanced samples from the training split and validate vector metrics on the validation split. 
For downstream tasks, model selection is driven by validation performance, while all final steering is performed on a test split.

\textbf{CelebA.} 
The CelebA dataset~\cite{liu2015deep} is a standard benchmark for bias removal~\cite{sagawa2020distributionally} and CAV evaluation~\cite{pahde2025navigating}. 
It consists of facial images annotated with $40$ binary attributes (concepts) such as \textit{male}, \textit{smiling}, or \textit{bald}.
Unlike other datasets, CelebA lacks a target task, so, similarly to the literature, we discard the \textit{male} concept from the concept set and treat it as a target classification task.

\textbf{ISIC.} 
To evaluate steering in a high-stakes medical domain, we use ISIC 2018~\cite{codella2019skin} for skin lesion classification with eight biases designed to simulate sampling artifacts~\cite{bissoto2020debiasing}.
This dataset is widely adopted for bias removal and CAV evaluation~\cite{wu2023discover,pahde2025navigating}.
Crucially, due to significant overlap between the training and validation sets, we evaluate CAV metrics on the test split and report results for the four concepts with sufficient sample support.

\textbf{Waterbirds.} 
Waterbirds~\cite{sagawa2020distributionally} is a well-established benchmark for spurious correlation analysis~\cite{wu2023discover}.
It specifically targets the natural correlation between a bird species (\emph{landbirds} or \emph{waterbirds}) and its background (\emph{water} or \emph{land}), which is balanced on the test split. Given the concepts' antipodal structure (see \Cref{tbl:results_waterbirds_clip}), we flag this dataset as low-quality and report its results only in the Appendix.

\textbf{Counteranimal.} To address environmental biases in models like CLIP~\cite{wang2024sober}, we include Counteranimal, which categorizes ImageNet~\cite{deng2009imagenet} images into \textit{common} (typical background, e.g., a polar bear on ice) and \textit{uncommon} (atypical background, e.g., a polar bear on grass) groups. We train CAVs to capture the direction of spurious background correlations and next test whether neutralizing these concepts enables background-invariant classification.

\textbf{ImageNet-W} (watermark) and \textbf{ImageNet-C} (corruptions). 
A critical challenge in evaluating real-world data is the lack of ground-truth concept counterfactuals to assess the steering success. 
To address this, we use ImageNet-W~\cite{li2023whac}, which modifies images by overlaying transparent textual \emph{watermarks}, and ImageNet-C~\citep{hendrycks2019benchmarking}, which applies one of 19 corruptions, such as glass blur or shot noise. 
This allows us to generate exact sample pairs $(\mathbf{x}_{\bar{c}}, \mathbf{x}_{c})$ and precisely measure the CAV's impact on the concept alone. 
Such an approach is particularly useful for measuring concept erasure on models sensitive to text, such as CLIP and SigLIP~\citep[see e.g.][]{baniecki2025explaining} and for evaluating classifier robustness under real-world conditions.
For computational efficiency, we limit our evaluation to a diverse subset of 10 classes, targeting the most degrading watermark–class pairs and corruptions (see Appendix~\ref{app:dataset_distribution}). 
% All results are reported across five seeds.

We also initially investigated \textbf{MetaShift}~\cite{liang2022metashift}, but decided to excluded it due to frequent label conflicts, as detailed in Appendix~\ref{app:dataset_distribution}.

%%%%%%%%%%%%%%%%%%%%%%%%%%%%%%%%%%%%%
\section{Results}
\label{sec:results}

We use \ourbenchmarkx{} to deliver several insightful findings regarding steering across multiple CAV methods, datasets, model backbones, and metrics. Extended analyses are provided in the Appendix.

\begin{customblockquote}\label{finding1}
    Linear SVM and logistic regression consistently achieve top concept detection accuracy, matching the performance of nonlinear probes across all datasets.
\end{customblockquote}
The primary proxy metric used prior to steering analysis is AUC.
Across all benchmarks, we observe that linear SVM and LR consistently yield the highest AUC and F1 scores (cf. Tables~\ref{tbl:results_dinov2}~\&~\ref{tbl:results_watermark}). 
Notably, these methods match the performance of complex nonlinear probes, such as TabPFN. 
We rigorously validate this observation on the CelebA and ISIC datasets in Apendix~\ref{app:linear_vs_nonlinear}, where we find no statistically significant difference between these two linear approaches and their nonlinear counterparts.

\begin{figure}[t]
    \centering
    \includegraphics[width=1\columnwidth]{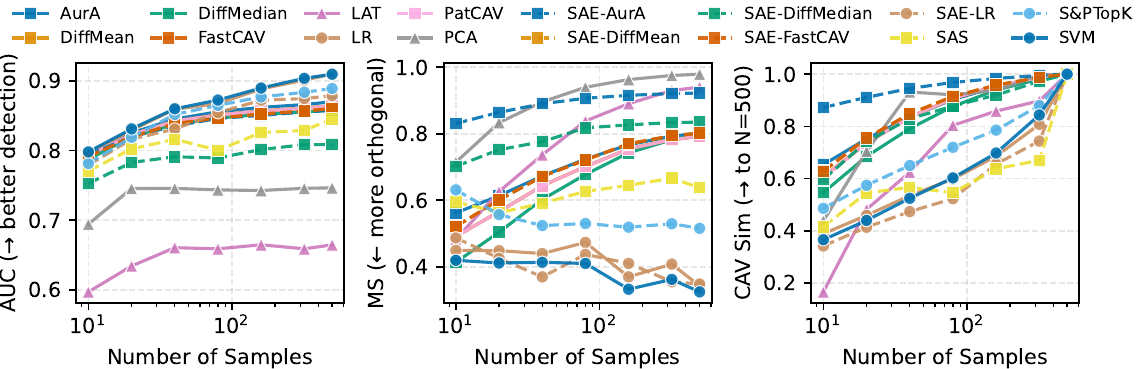}
    \caption{\textbf{Sample efficiency analysis on CelebA (SigLIP).} 
    We report AUC (left), MS (middle), and Cosine Similarity to the CAV trained on $N=500$ (right) as a function of training sample size $N \in \{10, \dots, 500\}$. 
    Notably, $\bullet$ methods show a decrease in MS as $N$ increases, indicating a more disentangled representation. See Appendix~\ref{app:extended_ablation_samples} for additional backbones and datasets results.
    }
    % \vspace{-0.75em}
    \label{fig:samples_ablations}
\end{figure}

\begin{customblockquote}\label{finding2}
    Optimization-based methods ($\bullet$) maximize vector separability, excelling in regimes with high concept co-occurrence often found in real-world data; however, they risk overfitting to redundant features in simpler settings.
    Conversely, while statistical~($\blacksquare$) may underperform in detection metrics or vector orthogonality, they identify truer global directions in cleaner datasets.
\end{customblockquote}
We can categorize methods into optimization-based ($\bullet$), statistical ($\blacksquare$), and environment-specific ($\blacktriangle$) based on their method type as presented in \Cref{fig:main}. 
Environment-specific methods ($\blacktriangle$) has weak real-data performance but perfect steering on the synthetic task~(\Cref{tbl:results_watermark}). 
Among statistical methods ($\blacksquare$), FastCAV, DiffMean, or PatCAV, are empirically and mathematically equivalent (see Appendix~\ref{app:stat_cav_equivalence}). We recognize AurA as the best method from this family as it achieves the highest overall performance on both synthetic and real-world data.
Comparing optimization-based~($\bullet$) and statistical~($\blacksquare$) methods reveals a critical trade-off: the former excel at disentanglement in complex, real-word data such as CelebA or ISIC, yielding better MS and CCR (\Cref{fig:samples_ablations}) and CD (\Cref{fig:cdd_heatmap}). 
In contrast, in simpler domains such as ImageNet-W/C, statistical ($\blacksquare$) methods often achieve greater steering success despite lower detection scores (\Cref{tbl:results_watermark}), as further evidenced in the Appendix. 
To summarize, $\bullet$ methods excel on entangled, real-world data but risk overfitting on cleaner datasets. On the other hand, $\blacksquare$ methods capture core semantics in simpler regimes but struggle with disentanglement.

\begin{table}[t]
    \centering
    \footnotesize
    \caption{\textbf{DINOv2 representations steering evaluation on CelebA and ISIC.} 
    Values show the mean $\pm$ two standard errors (2SE) across concepts. The final CD metric is reported as an absolute value.
    The best method (row) in each evaluation metric (column) is in \textbf{bold} and the second best is \underline{underlined}.
    For additional reference, two nonlinear probe baselines are denoted with {\color{gray}\textbf{gray}}.
    Extended results for the other models, the remaining datasets, and metrics, are in Appendix~\ref{app:extended_results}.
    }
    \label{tbl:results_dinov2}
    \vspace{0.5em}
    \begin{tblr}{
      colspec = {lcccccccccc},
      column{1} = {leftsep=0pt},
      vline{2,7} = {dashed},
      hline{8,17,19} = {dashed},
      row{5,9,11,13,16} = {bg=black!10},
      rowsep=1.5pt,
      colsep = 2.55pt
    }
    & \SetCell[c=5]{c} \textbf{CelebA} & & & & & \SetCell[c=5]{c} \textbf{ISIC} & & & & \\
    \textbf{Method} & AUC ($\uparrow$) & CCR ($\uparrow$) & MS ($\downarrow$) & F1 ($\uparrow$) & CD ($\downarrow$) & AUC ($\uparrow$) & CCR ($\uparrow$) & MS ($\downarrow$) & F1 ($\uparrow$) & CD ($\downarrow$) \\
    \hline
    Linear SVM & $\mathbf{.88}_{\pm.03}$ & $\mathbf{.97}_{\pm.01}$ & $\mathbf{.17}_{\pm.03}$ & $\mathbf{.56}_{\pm.06}$ & $\underline{0.02}_{\pm0.02}$ & $\mathbf{.91}_{\pm.09}$ & $\mathbf{.98}_{\pm.02}$ & $\mathbf{.07}_{\pm.04}$ & $\mathbf{.81}_{\pm.15}$ & $1.55_{\pm2.26}$ \\
    LR & $\mathbf{.88}_{\pm.03}$ & $\underline{.93}_{\pm.02}$ & $\underline{.30}_{\pm.05}$ & $\mathbf{.56}_{\pm.07}$ & $0.35_{\pm0.46}$ & $\mathbf{.91}_{\pm.09}$ & $\underline{.97}_{\pm.04}$ & $\underline{.09}_{\pm.03}$ & $\mathbf{.81}_{\pm.16}$ & $2.30_{\pm2.75}$ \\
    \hspace{5pt}\includegraphics[width=6pt]{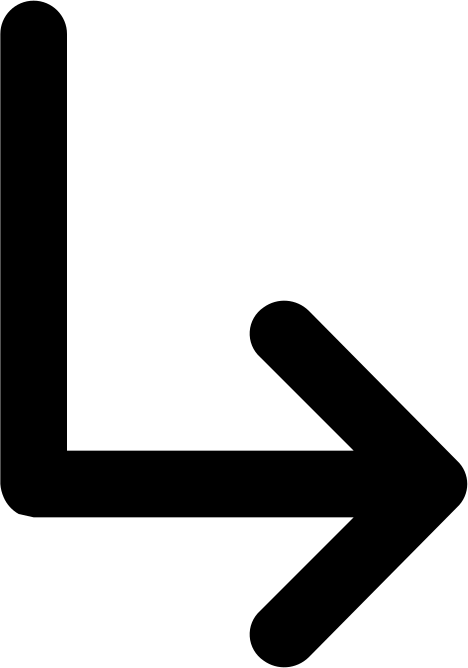} SAE & $.84_{\pm.03}$ & $.87_{\pm.03}$ & $.45_{\pm.08}$ & $.51_{\pm.06}$ & $0.71_{\pm0.84}$ & $.84_{\pm.11}$ & $\underline{.97}_{\pm.04}$ & $.12_{\pm.05}$ & $.69_{\pm.16}$ & $2.86_{\pm2.80}$ \\
    SAS & $.79_{\pm.04}$ & $.82_{\pm.03}$ & $.72_{\pm.06}$ & $.45_{\pm.07}$ & $3.70_{\pm1.54}$ & $.75_{\pm.12}$ & $.94_{\pm.03}$ & $.40_{\pm.18}$ & $.69_{\pm.13}$ & $2.92_{\pm6.50}$ \\
    S\&PTopK & $\underline{.85}_{\pm.03}$ & $.79_{\pm.05}$ & $.64_{\pm.06}$ & $\underline{.52}_{\pm.07}$ & $0.42_{\pm0.26}$ & $\underline{.87}_{\pm.12}$ & $.92_{\pm.08}$ & $.37_{\pm.09}$ & $\underline{.77}_{\pm.14}$ & $\mathbf{0.72}_{\pm2.03}$ \\
    DiffMean & $.83_{\pm.03}$ & $.82_{\pm.03}$ & $.82_{\pm.05}$ & $.49_{\pm.07}$ & $15.3_{\pm4.84}$ & $.86_{\pm.07}$ & $.93_{\pm.06}$ & $.50_{\pm.25}$ & $.73_{\pm.12}$ & $9.02_{\pm6.67}$ \\
    \hspace{5pt}\includegraphics[width=6pt]{figures_new/subdirectory_arrow_right.png} SAE & $.83_{\pm.03}$ & $.81_{\pm.03}$ & $.84_{\pm.05}$ & $.48_{\pm.07}$ & $16.2_{\pm5.00}$ & $.83_{\pm.08}$ & $.95_{\pm.04}$ & $.50_{\pm.25}$ & $.71_{\pm.12}$ & $9.52_{\pm10.6}$ \\
    DiffMedian & $.83_{\pm.03}$ & $.83_{\pm.03}$ & $.82_{\pm.05}$ & $.48_{\pm.07}$ & $14.8_{\pm4.66}$ & $.85_{\pm.07}$ & $.94_{\pm.06}$ & $.51_{\pm.30}$ & $.73_{\pm.11}$ & $10.1_{\pm5.77}$ \\
    \hspace{5pt}\includegraphics[width=6pt]{figures_new/subdirectory_arrow_right.png} SAE & $.78_{\pm.04}$ & $.76_{\pm.04}$ & $.85_{\pm.04}$ & $.44_{\pm.07}$ & $7.42_{\pm2.98}$ & $.74_{\pm.07}$ & $.96_{\pm.03}$ & $.48_{\pm.29}$ & $.67_{\pm.13}$ & $2.28_{\pm4.87}$ \\
    FastCAV & $.83_{\pm.03}$ & $.82_{\pm.03}$ & $.82_{\pm.05}$ & $.49_{\pm.07}$ & $15.3_{\pm4.84}$ & $.86_{\pm.07}$ & $.93_{\pm.06}$ & $.50_{\pm.25}$ & $.73_{\pm.12}$ & $9.02_{\pm6.67}$ \\
    \hspace{5pt}\includegraphics[width=6pt]{figures_new/subdirectory_arrow_right.png} SAE & $.83_{\pm.03}$ & $.81_{\pm.03}$ & $.84_{\pm.05}$ & $.48_{\pm.07}$ & $16.2_{\pm5.00}$ & $.83_{\pm.08}$ & $.95_{\pm.04}$ & $.50_{\pm.25}$ & $.71_{\pm.12}$ & $9.52_{\pm10.6}$ \\
    PatCAV & $.83_{\pm.03}$ & $.82_{\pm.03}$ & $.82_{\pm.05}$ & $.49_{\pm.07}$ & $15.3_{\pm4.84}$ & $.86_{\pm.07}$ & $.93_{\pm.06}$ & $.50_{\pm.25}$ & $.73_{\pm.12}$ & $9.02_{\pm6.67}$ \\
    AurA & $.83_{\pm.03}$ & $.81_{\pm.03}$ & $.85_{\pm.04}$ & $.49_{\pm.07}$ & $0.62_{\pm0.22}$ & $.86_{\pm.07}$ & $.93_{\pm.06}$ & $.62_{\pm.18}$ & $.73_{\pm.12}$ & $6.13_{\pm4.46}$ \\
    \hspace{5pt}\includegraphics[width=6pt]{figures_new/subdirectory_arrow_right.png} SAE & $.81_{\pm.03}$ & $.62_{\pm.04}$ & $1.00_{\pm.00}$ & $.47_{\pm.06}$ & $\mathbf{0.01}_{\pm0.01}$ & $.83_{\pm.09}$ & $.85_{\pm.12}$ & $.98_{\pm.01}$ & $.71_{\pm.12}$ & $\underline{1.23}_{\pm1.47}$ \\
    PCA & $.68_{\pm.04}$ & $.60_{\pm.05}$ & $.96_{\pm.02}$ & $.35_{\pm.06}$ & $8.26_{\pm3.88}$ & $.68_{\pm.16}$ & $.74_{\pm.26}$ & $.64_{\pm.71}$ & $.59_{\pm.18}$ & $12.9_{\pm4.11}$ \\
    LAT & $.56_{\pm.02}$ & $.52_{\pm.07}$ & $.96_{\pm.01}$ & $.28_{\pm.06}$ & $2.77_{\pm1.98}$ & $.63_{\pm.14}$ & $.43_{\pm.21}$ & $.51_{\pm.92}$ & $.58_{\pm.18}$ & $9.96_{\pm5.11}$ \\
    {\color{gray} SVM-RBF} & {\color{gray} $\mathbf{.88}_{\pm.03}$} & -- & -- & -- & -- & {\color{gray} $\mathbf{.91}_{\pm.01}$} & -- & -- & -- & -- \\
    {\color{gray} TabPFN} & {\color{gray} $\mathbf{.88}_{\pm.03}$} & -- & -- & -- & -- & {\color{gray} $\mathbf{.91}_{\pm.01}$} & -- & -- & -- & -- \\
    \hline
    \end{tblr}
\end{table}

\begin{customblockquote}\label{finding3}
    While most methods achieve zero collateral damage, only LAT ($\blacktriangle$) consistently achieves optimal steering disparity on ImageNet-W, while PCA achieves it on ImageNet-C. This highlights that, in simpler setups, popular methods like DiffMean, SAE, and SVM optimize for axes other than the critical steering metric, resulting in underperformance on that main task.
\end{customblockquote}
In synthetic settings (ImageNet-W), while most CAVs minimize collateral damage and possess high F1, only LAT (\Cref{tbl:results_watermark}), LR, PCA, and SAS (\Cref{tbl:results_imagenetw_church_clip})  achieve perfect steering, with PCA consistently yielding the lowest collateral damage on ImageNet-C.
LAT succeeds only under the counterfactual parity assumption and, interestingly, demonstrates that perfect steering does not require high separability (\Cref{tbl:results_imagenetw_church_clip,tbl:results_imagenetw_husky_clip}). We highlight that although standard optimization methods ($\bullet$) like SVM and LR yield peak performance and orthogonality (e.g., \Cref{fig:main}), both optimization and statistical methods ($\blacksquare$) struggle to achieve perfect steering. Instead, LAT and PCA lead in the benchmark for steering on ImageNet-W and ImageNet-C, respectively.

\begin{customblockquote}\label{finding4}
    Standard classification metrics, such as AUC and F1, are insufficient proxies for concept entanglement, robustness, collateral damage, or steering disparity.
\end{customblockquote}
Performance metrics such as AUC and F1 are inadequate for assessing the safety and fidelity of steering, a limitation highlighted in Findings~\ref{finding2} and \ref{finding3} and further corroborated by \Cref{fig:main}.
While improved disentanglement (lower MS) correlates with higher CCR, (Appendix~\ref{app:metric_analysis}), \Cref{fig:main} demonstrates that these metrics do not correlate with detection metrics. 
Furthermore, vector metrics fail to predict steering metrics, motivating the improved evaluation protocols introduced in \ourbenchmarkx{}.

\begin{table}[t]
    \centering
    \footnotesize
    \caption{
    \textbf{SigLIP representation steering evaluation on ImageNet-W watermark--class pairs (\emph{husky--cat} and \emph{church--goldfish}), and ImageNet-C \emph{Shot Noise}, and \emph{Glass Blur} corruptions.}
    CAVs are trained on $N=500$ counterfactual pairs sampled from all classes (All Classes Paired setup).
    % The best method (row) in each evaluation metric (column) is in \textbf{bold} and the second best is \underline{underlined}. 
    LAT achieves the best SD and CD on ImageNet-W but performs poorly on ImageNet-C, whereas AurA and DiffMedian perform well on both datasets. 
    Values report the mean $\pm$ 2SE across five seeds.
    }
    \label{tbl:results_watermark}
    \vspace{0.5em}
    \begin{tblr}{
      colspec = {lcccccccccc},
      column{1} = {leftsep=0pt},
      vline{2,5,8,10} = {dashed},
      hline{8,17} = {dashed},
      row{5,9,11,13,16} = {bg=black!10},
      rowsep=1.5pt,
      colsep = 2.6pt
    }
    & \SetCell[c=3]{c} \textbf{Husky on Cat} & & & \SetCell[c=3]{c} \textbf{Church on Goldfish} & & & \SetCell[c=2]{c} \textbf{Shot Noise} & & \SetCell[c=2]{c} \textbf{Glass Blur} & \\
    \textbf{Method} & F1 ($\uparrow$) & CD ($\downarrow$) & SD ($\downarrow$) & F1 ($\uparrow$) & CD ($\downarrow$) & SD ($\downarrow$) & F1 ($\uparrow$) & SD ($\downarrow$) & F1 ($\uparrow$) & SD ($\downarrow$) \\
    \hline
    Linear SVM & $.89_{\pm.05}$ & $\mathbf{.00}_{\pm.00}$ & $.39_{\pm.15}$ & $\mathbf{.97}_{\pm.01}$ & $\mathbf{.00}_{\pm.00}$ & $.69_{\pm.07}$ & $\mathbf{1.0}_{\pm.00}$ & $1.1_{\pm.02}$ & $\mathbf{1.0}_{\pm.00}$ & $1.0_{\pm.03}$ \\
    LR & $.88_{\pm.07}$ & $\mathbf{.00}_{\pm.00}$ & $.49_{\pm.27}$ & $\underline{.96}_{\pm.01}$ & $\mathbf{.00}_{\pm.00}$ & $.66_{\pm.00}$ & $\mathbf{1.0}_{\pm.00}$ & $.99_{\pm.02}$ & $\mathbf{1.0}_{\pm.00}$ & $.97_{\pm.04}$ \\
    \hspace{5pt}\includegraphics[width=6pt]{figures_new/subdirectory_arrow_right.png} SAE & $.71_{\pm.22}$ & $.04_{\pm.08}$ & $.34_{\pm.09}$ & $.86_{\pm.07}$ & $\mathbf{.00}_{\pm.00}$ & $.73_{\pm.08}$ & $\mathbf{1.0}_{\pm.00}$ & $1.1_{\pm.03}$ & $.98_{\pm.01}$ & $.96_{\pm.01}$ \\
    SAS & $.85_{\pm.16}$ & $.04_{\pm.08}$ & $.32_{\pm.28}$ & $.83_{\pm.03}$ & $\mathbf{.00}_{\pm.00}$ & $.49_{\pm.36}$ & $\mathbf{1.0}_{\pm.00}$ & $.99_{\pm.01}$ & $\mathbf{1.0}_{\pm.00}$ & $.93_{\pm.02}$ \\
    S\&PTopK & $\underline{.90}_{\pm.02}$ & $\mathbf{.00}_{\pm.00}$ & $.65_{\pm.20}$ & $\underline{.96}_{\pm.01}$ & $\mathbf{.00}_{\pm.00}$ & $.66_{\pm.00}$ & $\mathbf{1.0}_{\pm.00}$ & $.97_{\pm.02}$ & $.99_{\pm.00}$ & $.95_{\pm.00}$ \\
    DiffMean & $.88_{\pm.01}$ & $\mathbf{.00}_{\pm.00}$ & $.11_{\pm.00}$ & $.77_{\pm.04}$ & $.08_{\pm.10}$ & $.19_{\pm.07}$ & $\mathbf{1.0}_{\pm.00}$ & $\underline{.96}_{\pm.06}$ & $\mathbf{1.0}_{\pm.00}$ & $.91_{\pm.02}$ \\
    \hspace{5pt}\includegraphics[width=6pt]{figures_new/subdirectory_arrow_right.png} SAE & $.86_{\pm.01}$ & $.20_{\pm.00}$ & $.24_{\pm.00}$ & $.63_{\pm.02}$ & $.40_{\pm.00}$ & $.56_{\pm.08}$ & $\mathbf{1.0}_{\pm.00}$ & $.99_{\pm.05}$ & $\mathbf{1.0}_{\pm.00}$ & $.88_{\pm.02}$ \\
    DiffMedian & $.87_{\pm.02}$ & $\mathbf{.00}_{\pm.00}$ & $\underline{.09}_{\pm.05}$ & $.80_{\pm.01}$ & $.08_{\pm.10}$ & $\mathbf{.15}_{\pm.00}$ & $\mathbf{1.0}_{\pm.00}$ & $\underline{.96}_{\pm.05}$ & $\mathbf{1.0}_{\pm.00}$ & $.91_{\pm.00}$ \\
    \hspace{5pt}\includegraphics[width=6pt]{figures_new/subdirectory_arrow_right.png} SAE & $.82_{\pm.01}$ & $\mathbf{.00}_{\pm.00}$ & $.72_{\pm.19}$ & $.68_{\pm.03}$ & $\mathbf{.00}_{\pm.00}$ & $.53_{\pm.07}$ & $\mathbf{1.0}_{\pm.00}$ & $.96_{\pm.01}$ & $.99_{\pm.00}$ & $\underline{.87}_{\pm.02}$ \\
    FastCAV & $.88_{\pm.01}$ & $\mathbf{.00}_{\pm.00}$ & $.11_{\pm.00}$ & $.77_{\pm.04}$ & $.08_{\pm.10}$ & $.19_{\pm.07}$ & $\mathbf{1.0}_{\pm.00}$ & $\underline{.96}_{\pm.06}$ & $\mathbf{1.0}_{\pm.00}$ & $.91_{\pm.02}$ \\
    \hspace{5pt}\includegraphics[width=6pt]{figures_new/subdirectory_arrow_right.png} SAE & $.86_{\pm.01}$ & $.20_{\pm.00}$ & $.24_{\pm.00}$ & $.63_{\pm.02}$ & $.40_{\pm.00}$ & $.56_{\pm.08}$ & $\mathbf{1.0}_{\pm.00}$ & $.99_{\pm.05}$ & $\mathbf{1.0}_{\pm.00}$ & $.88_{\pm.02}$ \\
    PatCAV & $.88_{\pm.01}$ & $\mathbf{.00}_{\pm.00}$ & $.11_{\pm.00}$ & $.77_{\pm.04}$ & $.08_{\pm.10}$ & $.19_{\pm.07}$ & $\mathbf{1.0}_{\pm.00}$ & $\underline{.96}_{\pm.06}$ & $\mathbf{1.0}_{\pm.00}$ & $.91_{\pm.02}$ \\
    AurA & $\mathbf{.95}_{\pm.00}$ & $\mathbf{.00}_{\pm.00}$ & $.11_{\pm.00}$ & $.89_{\pm.02}$ & $\mathbf{.00}_{\pm.00}$ & $.49_{\pm.00}$ & $\mathbf{1.0}_{\pm.00}$ & $\underline{.96}_{\pm.03}$ & $\mathbf{1.0}_{\pm.00}$ & $\mathbf{.81}_{\pm.01}$ \\
    \hspace{5pt}\includegraphics[width=6pt]{figures_new/subdirectory_arrow_right.png} SAE & $.89_{\pm.01}$ & $\mathbf{.00}_{\pm.00}$ & $.24_{\pm.00}$ & $\underline{.96}_{\pm.00}$ & $\mathbf{.00}_{\pm.00}$ & $.66_{\pm.00}$ & $\mathbf{1.0}_{\pm.00}$ & $.98_{\pm.01}$ & $\mathbf{1.0}_{\pm.00}$ & $.96_{\pm.00}$ \\
    PCA & $.46_{\pm.04}$ & $8.8_{\pm1.9}$ & $3.6_{\pm1.0}$ & $.20_{\pm.01}$ & $9.1_{\pm4.5}$ & $2.6_{\pm.71}$ & $\mathbf{1.0}_{\pm.00}$ & $\mathbf{.93}_{\pm.04}$ & $\mathbf{1.0}_{\pm.00}$ & $.91_{\pm.04}$ \\
    LAT & $.51_{\pm.01}$ & $\mathbf{.00}_{\pm.00}$ & $\mathbf{.04}_{\pm.06}$ & $.54_{\pm.01}$ & $\mathbf{.00}_{\pm.00}$ & $\mathbf{.15}_{\pm.00}$ & $.75_{\pm.01}$ & $1.6_{\pm.07}$ & $.73_{\pm.00}$ & $1.5_{\pm.16}$ \\
    \hline
    \end{tblr}
\end{table}

\begin{customblockquote}\label{finding5}
    Filtering of noisy neurons is essential to recover utility from SAE representations.
    While methods like SAE-AurA and S\&PTopK are justifiable for isolating concepts in complex, entangled concept datasets, they are unnecessary in simpler regimes and add only training overhead.
\end{customblockquote}
It is known that directly substituting standard representations $h$ with SAE features typically degrades performance~\cite{kantamneni2025are}. 
However, when applying rigorous filtering strategies such as S\&PTopK, SAE methods consistently demonstrate strong separation and high concept disentanglement, often leading to lower CD (\Cref{fig:cdd_heatmap}). 
While SAE-AurA scores often worse on steering disparity (\Cref{tbl:results_watermark}), it reduces side effects on real data and synthetic ones (see Appendix~\ref{app:extended_results}).
Despite the SAE method's high computational overhead and non-trivial optimization, which often yield inconsistent results across backbones depending on SAE quality, methods like S\&PTopK and SAE-AurA achieve high concept detection with low collateral damage on complex datasets. On simpler datasets like ImageNet-W/C, baseline methods demonstrate consistent superiority across all metrics.

\textbf{Ablations on the sample-efficiency of CAVs.}
\Cref{fig:samples_ablations} shows that the optimal number of samples for training CAVs varies by method group. 
Statistical methods~($\blacksquare$) and LAT saturate rapidly, typically stabilizing around $N=300$, whereas optimization-based~($\bullet$) continue to improve both AUC and MS, possibly even over $N=500$, on complex datasets like CelebA. 
For simpler domains, saturation occurs much earlier (see Appendix~\ref{app:extended_ablation_samples}); on ImageNet-W, statistical~($\blacksquare$) methods converge as early as $N=50$--$100$, whereas optimization-based~($\bullet$) prove constant CAV optimization. 

\textbf{Ablations on different vision model backbones.}
We identify SigLIP as the superior backbone for concept detection and collateral damage, consistently matching or outperforming CLIP and DINOv2 on CelebA and ISIC, as confirmed by critical difference plots in \Cref{fig:cd_plots_celeba_extended,fig:cd_plots_isic}.
However, DINOv2 demonstrates distinct strengths: it exhibits robustness to watermarking (\Cref{tab:imagenet_confounders}) and achieves superior separation performance (\Cref{fig:cd_plots_isic}).
We hypothesize that these backbones excel in different regimes: SigLIP captures high-level semantic concepts, whereas DINOv2 excels at structural features.

%%%%%%%%%%%%%%%%%%%%%%%%%%%%%%%%%%%%%
\section{Conclusion}
\label{sec:conclusion}
As image representations become central to real-world applications, the ability to reliably steer their behavior is paramount. 
We introduce \ourbenchmark{} to unify and extend the fragmented landscape of steering and evaluation methods proposed in the literature. 
By rigorously evaluating concept activation vectors on visual tasks, our benchmark highlights the often-neglected second-order effects of steering, offering a critical assessment that existing metrics cannot.

\textbf{Limitations.}
\ourbenchmarkx{} is inherently limited by the limitations of datasets, some of which suffer from low data quality (e.g., Waterbirds), a lack of high-stakes tasks (e.g., CelebA), and sparse concept vocabulary with insufficient data (e.g., ISIC).
While we address this by aggregating our benchmark findings across a diverse set of tasks, future work should expand the evaluation to include synthetically generated images.
To focus our analysis of the 16 CAV methods across several dimensions and model architectures, we leave multi-concept and multi-layer/head steering tasks for future work.   

\textbf{Broader impact.} 
Ultimately, we envision \ourbenchmarkx{} as a living benchmark, continually expanding to include new datasets, metrics, and methods proposed by the community.
We further elaborate on the potential positive and negative societal impact of this work in Appendix~\ref{app:broader_impact}.

%%%%%%%%%%%%%%%%%%%%%%%%%%%%%%%%%%%%%

\begin{ack}
    Work on this project is financially supported by the Foundation for Polish Science (FNP) grant `Centre for Credible AI' No. FENG.02.01-IP.05-0058/24, and carried out with the support of the Laboratory of Bioinformatics and Computational Genomics and the High Performance Computing Center of the Faculty of Mathematics and Information Science, Warsaw University of Technology. Hubert Baniecki was supported by the Foundation for Polish Science (FNP) START scholarship.
\end{ack}

\clearpage
\bibliographystyle{plainnat}
\bibliography{main}

%%%%%%%%%%%%%%%%%%%%%%%%%%%%%%%%%%%%%%%%%%%%%%%%%%%%%%%%%%%%%%%%%%%%%%%%%%%%%%%
%%%%%%%%%%%%%%%%%%%%%%%%%%%%%%%%%%%%%%%%%%%%%%%%%%%%%%%%%%%%%%%%%%%%%%%%%%%%%%%
% APPENDIX
%%%%%%%%%%%%%%%%%%%%%%%%%%%%%%%%%%%%%%%%%%%%%%%%%%%%%%%%%%%%%%%%%%%%%%%%%%%%%%%
%%%%%%%%%%%%%%%%%%%%%%%%%%%%%%%%%%%%%%%%%%%%%%%%%%%%%%%%%%%%%%%%%%%%%%%%%%%%%%%
\newpage
\appendix
\section*{Appendix for ``SwordBench: Evaluating Orthogonality of Steering Image Representations''}

\startcontents[sections]
\printcontents[sections]{l}{1}{\setcounter{tocdepth}{2}}

%%%%
\clearpage
\section{Notation}
\label{app:notation}

\begin{table}[H]
\centering
\caption{Main notations used in the paper.}
\vspace{2pt}
\label{tab:notation}
\begin{tabular}{cl}
\toprule
\textbf{Notation} & \textbf{Meaning}\\
\hline
\multicolumn{2}{c}{\textit{Models \& representations}} \\
\hline
 $\phi$ & Frozen vision model backbone (e.g., CLIP, DINOv2)\\
 $h$ & Activation vector (representation) from the final layer of $\phi$, $h \in \mathbb{R}^d$\\
 $\psi$ & Downstream linear classifier\\
 $d$ & Dimensionality of the representation space\\
\hline
\multicolumn{2}{c}{\textit{Datasets \& concepts}} \\
\hline
 $\Aone, \Azero$ & Sets of samples where the concept is present ($y=1$) and absent ($y=0$)\\
 $N$ & Number of samples per balanced set \\
 $\mathbf{x}$ & A clean sample and its concept-infused variant (synthetic data)\\
 $y$ & Binary concept label\\
\hline
\multicolumn{2}{c}{\textit{Steering \& SAEs}} \\
\hline
 $\mathbf{v}_c$ & Concept activation vector (CAV)\\
 $\bar{h}$ & Steered representation (after removal/orthogonalization)\\
 $z$ & Sparse latent feature from Sparse Autoencoder (SAE)\\
 $W_{\text{enc}}, W_{\text{dec}}$ & SAE Encoder and Decoder weight matrices\\
 $\mathcal{S}$ & Set of selected Top-K feature indices (for S\&PTopK)\\
\hline
\multicolumn{2}{c}{\textit{Vector evaluation metrics}} \\
\hline
 $\text{AUC}$ & Detection Accuracy (Linear Separability)\\
  $\text{MAD}$ & Mean Activation Difference\\
 $\text{MS}$ & Maximum Similarity (Orthogonality/Disentanglement)\\
 $\text{CCR}$ & Cross-Concept Robustness (Functional Independence)\\
\hline
\multicolumn{2}{c}{\textit{Downstream steering metrics}} \\
\hline
 $\text{Acc}$ & Classification accuracy\\
 $\text{CD}$ & Collateral Damage (Safety metric on $\Azero$)\\
 $\text{SD}$ & Steering Disparity (Impact metric on counterfactual pairs $\Aone$)\\
 $\text{F1}$ & Harmonic mean of precision and recall\\
\hline
\end{tabular}
\end{table}
\vspace{10pt}

%%%%
\clearpage
\section{Extended Methods}
\subsection{Steering Intervention}
\label{app:steering_intervention}

After extracting a concept vector $\mathbf{v}_c$, steering interventions are typically performed by adding $\mathbf{v}_c$ to the original representation $h$ with a scaling parameter $\alpha$~\cite{rimsky2024steering,soo2025interpretable,wu2025axbench}:
\begin{equation}
    \bar{h} = h + \alpha\mathbf{v}_c.
\end{equation}
Here, $\alpha$ takes a negative value for concept removal and a positive one for addition. 
Since excessive steering can degrade model performance~\cite{o2024steering}, $\alpha$ must often be tuned for a specific concept task.
How to tune is an open research question~\citep{hedstrom2025tosteer}.
Moreover, \textbf{this optimization becomes non-trivial when a single vector is evaluated across multiple metrics}: one must either report results with $\alpha$ optimized separately for each metric or aggregate them through a principled heuristic. 
Neither approach has a clear canonical preference to apply for the fair comparison. 
While AxBench~\cite{wu2025axbench} aggregates metrics using a harmonic mean, we argue this may lead to biased optimization. 
Therefore, we focus on the concept erasure task via orthogonalization, eliminating the need to tune steering parameters.

In concept erasure, the objective is to remove specific information from the representation space. 
INLP~\cite{ravfogel2020null} observes that simple orthogonalization often fails to fully eliminate a concept, motivating more sophisticated approaches. 
They propose an iterative procedure that repeats orthogonalization until a linear classifier can no longer distinguish the concept from the activation space. 
RLACE~\cite{ravfogel2022linear} later framed this as a linear minimax game solved via convex relaxation. 
However, because CAVs are frequently correlated with other concepts (exhibiting non-zero MS), such iterative procedures may inflict substantial collateral damage to unrelated features while aggressively maximizing concept removal.
Notably, these procedures require optimizing the number of iterations for each separate CAV and task, which hinders fair benchmarking.
Given that simple orthogonalization is often sufficient on synthetic datasets, we do not use INLP in the \ourbenchmark{}. 
Furthermore, we omit LEACE~\cite{belrose2023leace}, which applies an affine transformation (centering and whitening) before projection, because the CAV methods are often evaluated on the raw rather than transformed activations. 
Further, estimating covariance requires a significant number of samples, which is not feasible in our benchmarking procedures, where $N\approx p$.
Applying such transformations might unfairly distort or compromise method performance.

We therefore use simple orthogonalization as our primary benchmark steering intervention. 
We emphasize that our goal is not to optimize for the best possible steering performance, but rather to compare existing methods on equal footing and present several findings, as presented in \Cref{sec:results}.

%%%%%
\subsection{Mean Activation Difference (MAD)}
\label{app:mad_metric}

To move beyond ranking metrics such as AUC, we adopt MAD from neuron interpretability literature~\cite{kopf2024cosy}. 
MAD quantifies the standardized mean projection score between positive ($\Aone$) and negative ($\Azero$) populations.
MAD is defined as:
\begin{equation}
    \text{MAD}(\Azero, \Aone) = \frac{\frac{1}{|\Aone|} \sum_{h \in \Aone} \mathbf{v}_c^T h - \frac{1}{|\Azero|} \sum_{h \in \Azero} \mathbf{v}_c^T h}{\sqrt{\frac{1}{|\Azero|-1} \sum_{h \in \Azero} (\mathbf{v}_c^T h - \mu_0)^2}},
\end{equation}
where $\mu_0 = \frac{1}{|\Azero|} \sum_{h \in \Azero} \mathbf{v}_c^T h$ is the mean activation of the negative group. 
Higher MAD values indicate a stronger signal-to-noise ratio for the concept direction.

However, in \Cref{fig:biplot} and further in \Cref{app:metric_analysis}, we demonstrate that MAD correlates highly with AUC, questioning the necessity of MAD given that AUC is a more established metric.

%%%%%
\subsection{The equivalence of FastCAV, PatCAV, and DiffMean}
\label{app:stat_cav_equivalence}

During preliminary results generation, we found that three CAV extraction methods — FastCAV, PatCAV, and DiffMean — produced similar results across all metrics. Notably, across all experiments, their CAVs have cosine similarities of over $0.99$. As we detail further in this section, their mathematical definitions, under assumptions fulfilled in our experimental setup, are equal, and only numerical errors may introduce any difference between them.

Let us first recall that the DiffMean method can be formulated as follows:
\begin{equation}
    \mathbf{v}_{\text{DiffMean}} = \frac{1}{N} \sum_{h \in \Aone} h - \frac{1}{N} \sum_{h \in \Azero} h
\end{equation}
A similar definition can be derived for FastCAV under the assumption that $|\Azero|=|\Aone|=N$, which holds in our experimental setup. We start the derivation with the initial FastCAV definition:
\begin{align}
    \mathbf{v}_{\text{FastCAV}} &= \frac{1}{N} \sum_{h \in \Aone} (h - \mu_{\text{global}}) = \frac{1}{N} \sum_{h \in \Aone} h - \frac{1}{2N} \sum_{h \in \Azero \cup \Aone} h \\
    &= \frac{1}{N}\sum_{h \in \Aone}h - \frac{1}{2N} \sum_{h \in \Azero} h - \frac{1}{2N} \sum_{h \in \Aone} h \\
    &= \frac{1}{2N}\sum_{h \in \Aone}h - \frac{1}{2N} \sum_{h \in \Azero} h \\
    &= \frac{1}{2} \left(\frac{1}{N}\sum_{h \in \Aone}h - \frac{1}{N} \sum_{h \in \Azero} h \right)
\end{align}
which is the same as the DiffMean up to the $\frac12$ scalar, which can be omitted since all CAVs are normalized.

A similar derivation for PatCAV is described in~\citep[][appendix B.3]{pahde2025navigating},
where the important assumption is that the concept labels are binary, which holds true throughout our experiments as well.

%%%%
\clearpage
\section{Experimental Setup and Training Details}
\label{app:experimental_setup_training}

Below, we detail the datasets, model configurations (CAV, SAE, and downstream classifiers), and the specific hyperparameters used in our evaluation.

\subsection{Details of the Datasets/Tasks}
\label{app:dataset_distribution}

In this section, we detail the class and concept distributions for all datasets used in \ourbenchmark{}. 
Throughout this section, we bold concepts in the train and validation splits (unless stated otherwise) when the sample count is fewer than $500$ in either set. Significantly low counts may indicate potential instability in CAV extraction.

\subsubsection{ISIC} 

The ISIC dataset is sourced from the benchmark repository~\cite{wu2023discover}, which derives from the original ISIC 2018 Task 1 \& 2 challenge~\cite{codella2019skin} with artifact annotations provided in~\cite{bissoto2020debiasing}. 
The authors manually annotated $8$ artifacts (e.g., rulers, markers) and constructed five overlapping ``trap'' sets intended to simulate varying bias intensities. The original ISIC 2018 dataset is under the CC-BY-NC 4.0 License and is openly available via the link: \url{https://drive.google.com/uc?id=1Os34EapIAJM34DrwZMw2rRRJij3HAUDV}

While the creators of the trap sets claim distinct bias factors for each version, our empirical analysis indicated that Traps 1--5 exhibit statistically similar bias distributions and concept overlaps.
Furthermore, upon investigating the splits, we observed significant overlap in the image data between the sets; consequently, we \textbf{evaluate only on Trap Set 1}. We report the distribution of the specific sample in \Cref{tab:isic_dataset}.
Following the protocol in~\cite{wu2023discover}, we strictly use the test split for all CAV evaluations to ensure validity, as the validation splits in this dataset are subsets of the training data.

We excluded concepts with fewer than $500$ training samples, labeled as \textit{Insufficient Data} in \Cref{tab:isic_dataset}. We aggregate results over the remaining four attributes: \textit{dark\_corner}, \textit{hair}, \textit{gel\_bubble}, and \textit{ruler}.

\begin{table}[H]
    \caption{
    \textbf{ISIC (trap 1) dataset distribution}. Total samples: train ($1,822$), val ($259$), test ($772$). 
    Counts below $500$ in train or test split are \textbf{bold}. The confounded class column identifies the class negatively impacted in the test set due to correlation reversal (e.g., \textit{ruler} is strongly correlated with \textit{malignant} in the train set, but acts as a confounder for \textit{benign} in the test set).
    The percent value denotes the fraction of samples in the given split that contain a concept. 
    Note that concepts may co-occur, so percentages in columns do not sum up to $100\%$.
    }
    \label{tab:isic_dataset}
    \centering
    \vspace{0.5em}
    \renewcommand{\arraystretch}{1.3} 
    \resizebox{\textwidth}{!}{%
    \begin{tabular}{m{1.5cm} l c cc cc cc}
    \toprule
    \multirow{2}{*}{\textbf{Image}} & 
    \multirow{2}{*}{\textbf{Concept}} & 
    \multirow{2}{*}{\textbf{\shortstack{Confounded\\class}}} & 
    \multicolumn{2}{c}{\textbf{Train data}} & 
    \multicolumn{2}{c}{\textbf{Val data}} & 
    \multicolumn{2}{c}{\textbf{Test data}} \\
    \cmidrule(lr){4-5} \cmidrule(lr){6-7} \cmidrule(lr){8-9}
     & & & benign & malignant & benign & malignant & benign & malignant \\
    \midrule
    % Row 1: dark_corner
    \includegraphics[width=0.08\textwidth]{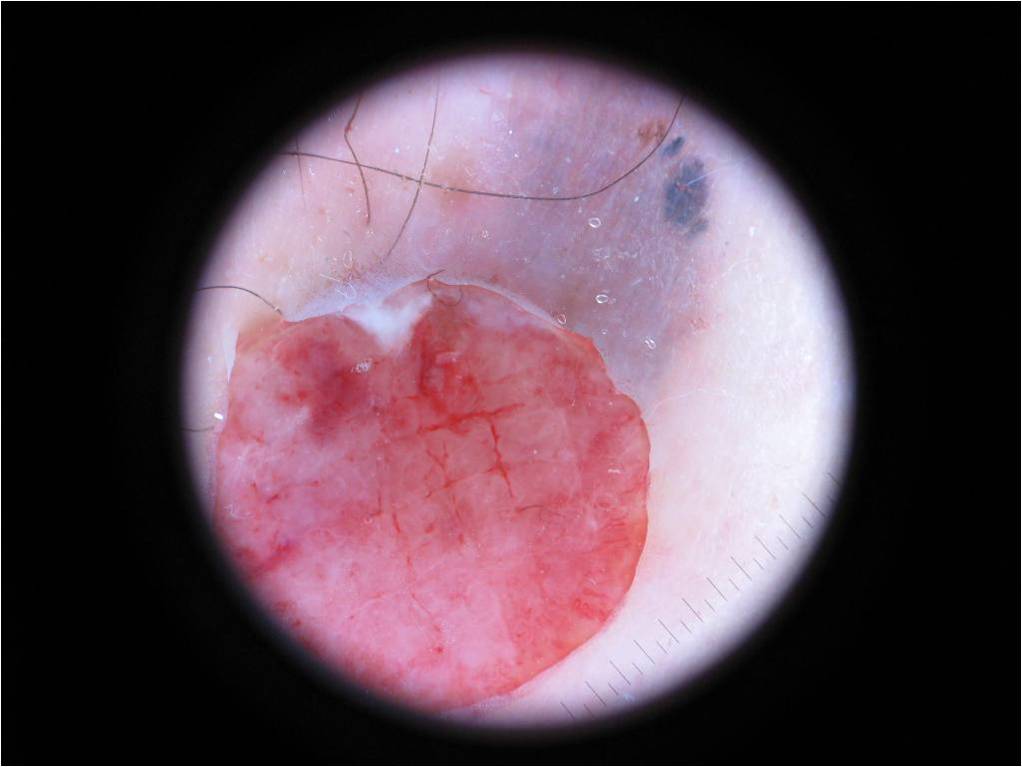} & 
    \textbf{dark\_corner} & benign & \textbf{386} ($27\%$) & \textbf{223} ($61\%$) & $51$ ($25\%$) & $29$ ($57\%$) & \textbf{340} ($55\%$) & \textbf{10} ($7\%$) \\
    % Row 2: hair
    \includegraphics[width=0.08\textwidth]{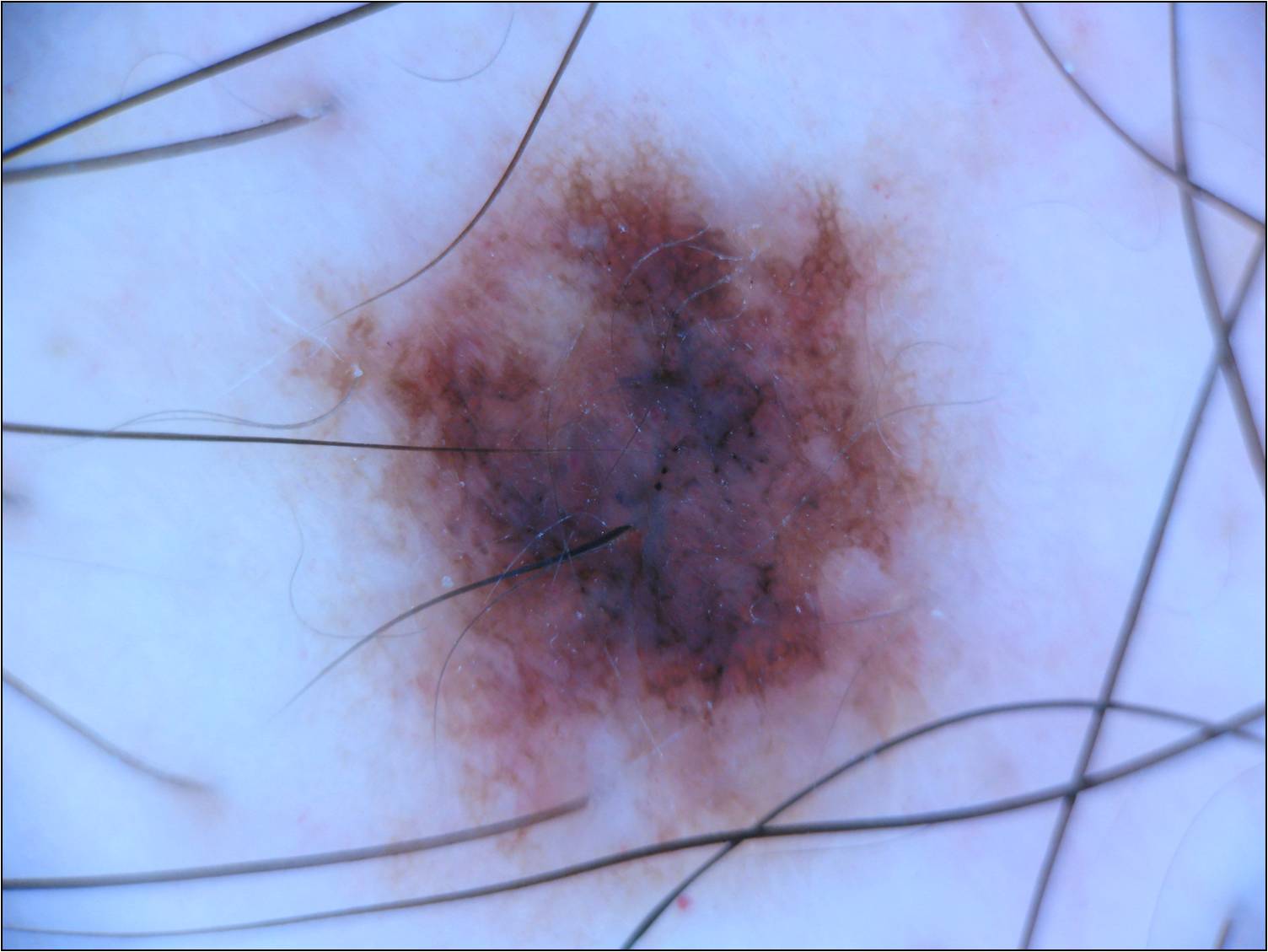} & 
    \textbf{hair} & malignant & $1,045$ ($72\%$) & \textbf{147} ($41\%$) & $153$ ($74\%$) & $17$ ($33\%$) & \textbf{209} ($34\%$) & \textbf{116} ($75\%$) \\
    % Row 3: gel_border
    \includegraphics[width=0.08\textwidth]{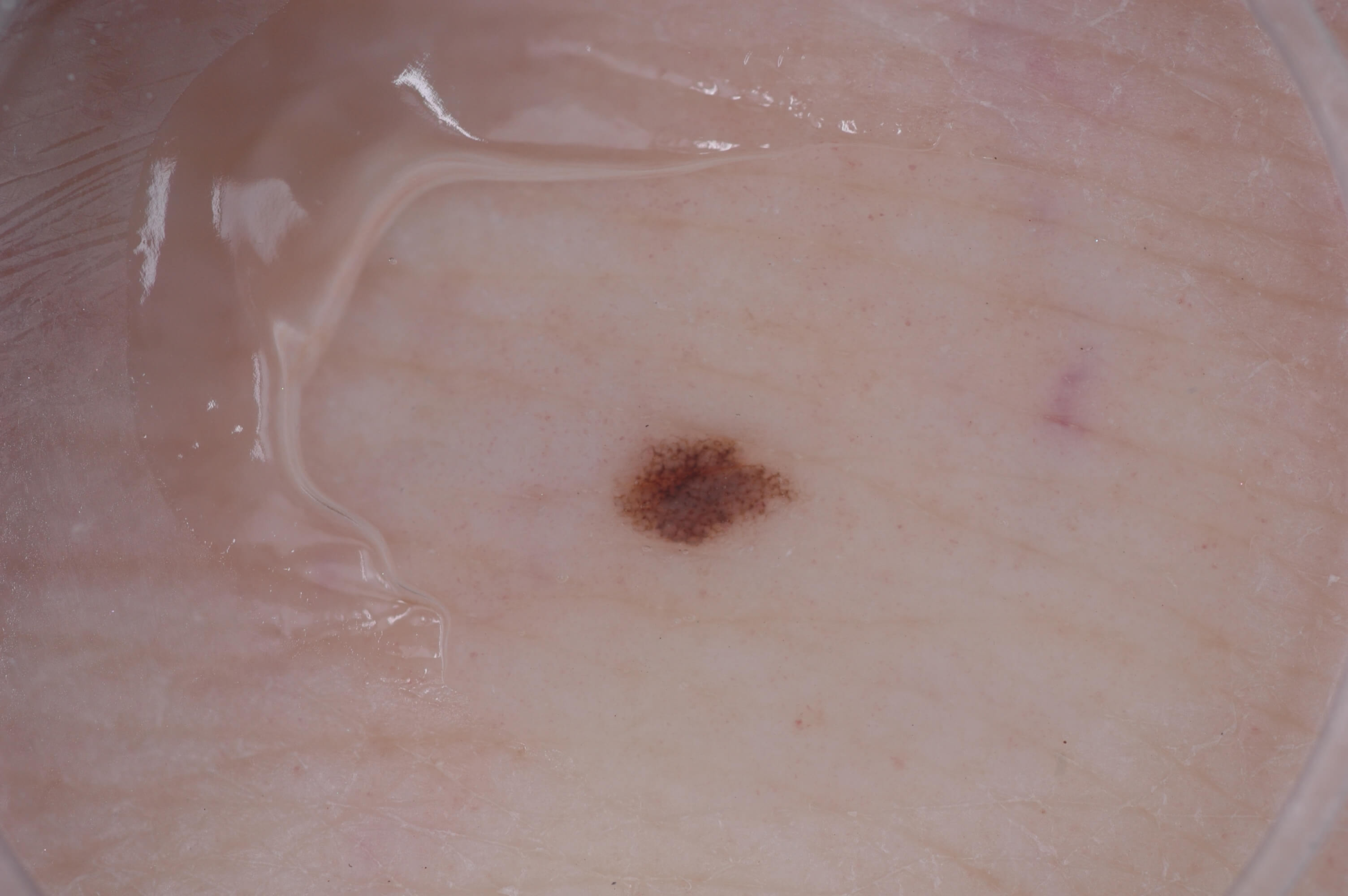} & 
    \textbf{gel\_border} & \footnotesize{\textit{Insufficient Data}} & \textbf{192} ($13\%$) & \textbf{89} ($24\%$) & $34$ ($16\%$) & $14$ ($28\%$) & \textbf{397} ($64\%$) & \textbf{0} ($0\%$) \\
    % Row 4: gel_bubble
    \includegraphics[width=0.08\textwidth]{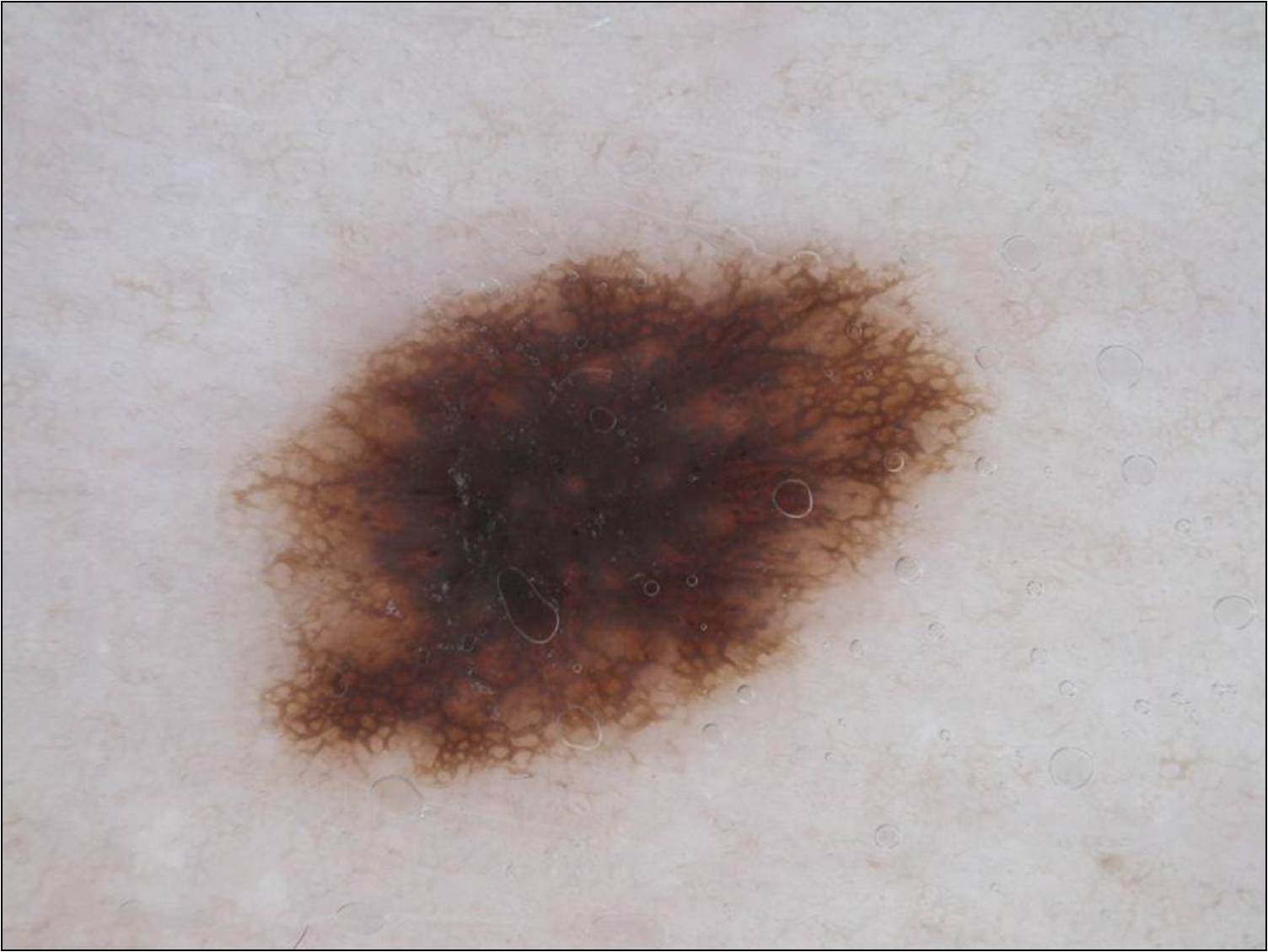} & 
    \textbf{gel\_bubble} & malignant & $847$ ($58\%$) & \textbf{128} ($35\%$) & $104$ ($50\%$) & $17$ ($33\%$) & \textbf{164} ($27\%$) & \textbf{129} ($84\%$) \\
    % Row 5: ruler
    \includegraphics[width=0.08\textwidth]{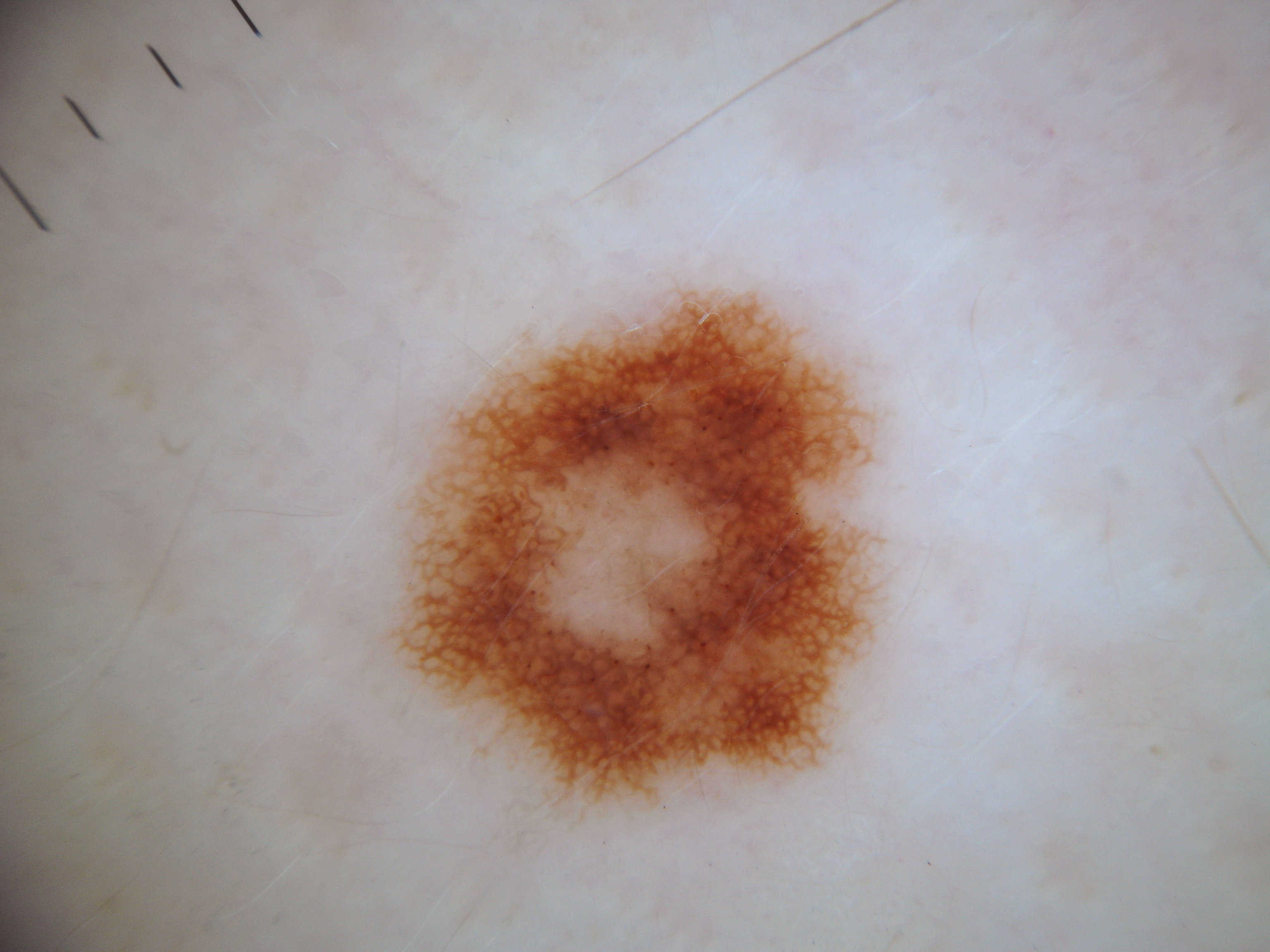} & 
    \textbf{ruler} & benign & \textbf{446} ($31\%$) & \textbf{295} ($81\%$) & $69$ ($33\%$) & $45$ ($88\%$) & $529$ ($86\%$) & \textbf{13} ($8\%$) \\
    % Row 6: ink
    \includegraphics[width=0.08\textwidth]{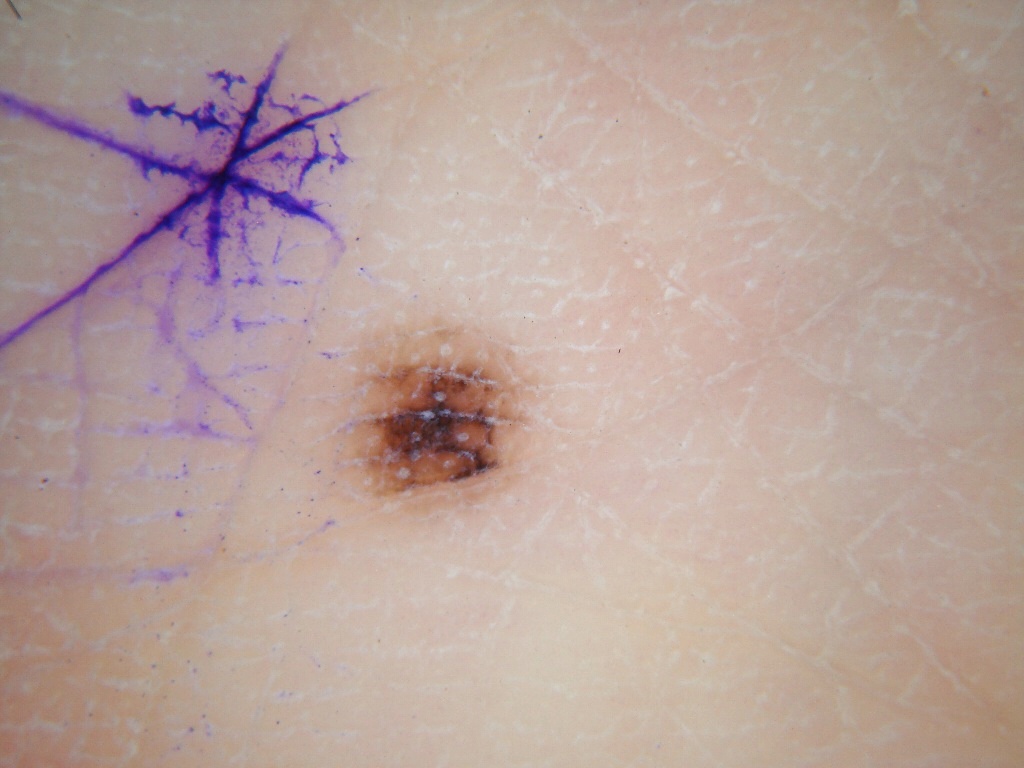} & 
    \textbf{ink} & \footnotesize{\textit{Insufficient Data}} & \textbf{57} ($4\%$) & \textbf{63} ($17\%$) & $7$ ($3\%$) & $8$ ($16\%$) & \textbf{322} ($52\%$) & \textbf{0} ($0\%$) \\
    % Row 7: patches
    \includegraphics[width=0.08\textwidth]{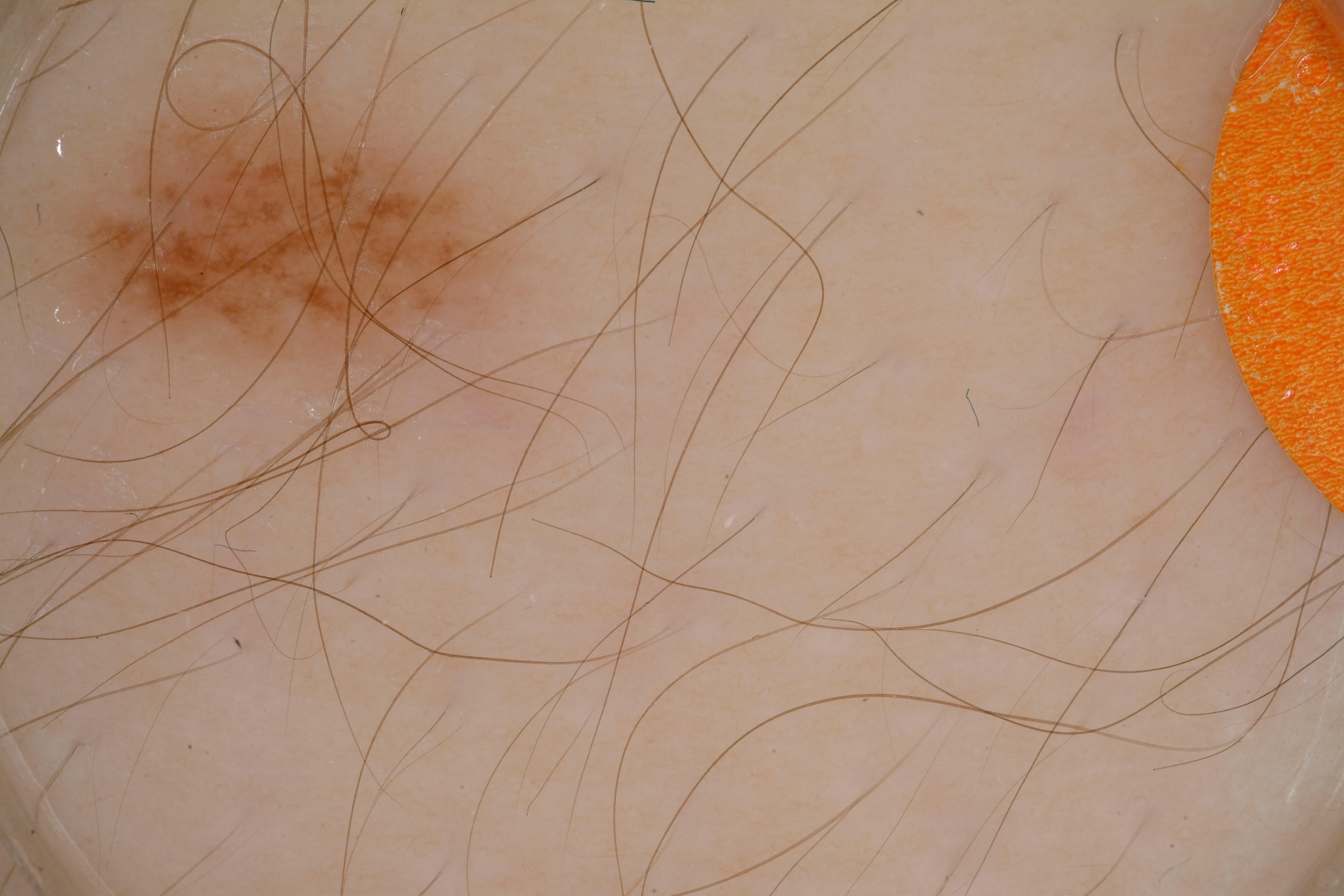} & 
    \textbf{patches} & \footnotesize{\textit{Insufficient Data}} & \textbf{111} ($8\%$) & \textbf{2} ($1\%$) & $21$ ($10\%$) & $0$ ($0\%$) & \textbf{74} ($12\%$) & \textbf{0} ($0\%$) \\
    % Row 8: vasc
    \includegraphics[width=0.08\textwidth]{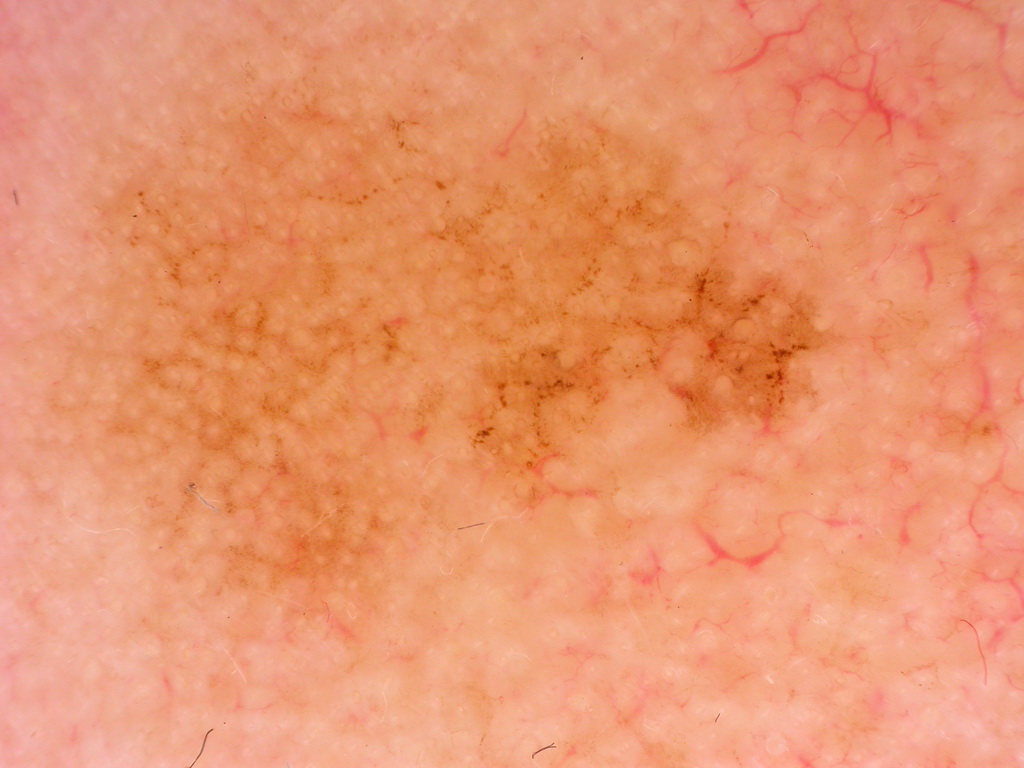} & 
    \textbf{vasc} & \footnotesize{\textit{Insufficient Data}} & \textbf{246} ($17\%$) & \textbf{109} ($30\%$) & $28$ ($14\%$) & $16$ ($31\%$) & \textbf{90} ($15\%$) & \textbf{49} ($32\%$) \\
    \bottomrule
    \end{tabular}%
    }
\end{table}

%%%%
% \clearpage
\subsubsection{CelebA}
The CelebA dataset contains facial images with attribute annotations but lacks a designated downstream target. This dataset is available under a Non-Commercial Research Only agreement and can be loaded via \texttt{torchvision.datasets.CelebA}. Given that gender is a prevalent confounding factor (\Cref{tab:celeba_dataset}) and is a popular attribute to evaluate~\cite{zaigrajew2025interpreting,sagawa2020distributionally}, we select \textit{Male} as our primary classification target. Hence, we evaluate and average all vector and steering results over the remaining 39 attributes. Detailed statistics of the dataset are provided in \Cref{tab:celeba_dataset}. 

\begin{table}[h!]
    \centering
    \footnotesize
    \caption{\textbf{CelebA dataset distribution}. Total samples: train ($162,770$), val ($19,867$), test ($19,962$). 
    Concepts with fewer than $500$ samples in train or val splits are \textbf{bolded}. 
    Overlap statistics report the conditional probability $P(\text{confounder} \mid \text{concept})$ in the training set. 
    High overlap (e.g., Mustache and Male) highlights risks where CAVs may learn confounders rather than targets.}
    \label{tab:celeba_dataset}
    \vspace{0.5em}
    \begin{tblr}{
        width = \textwidth,
        colspec = {X[1.4,l] X[r] X[r] X[r] X[r] X[l]},
        column{1} = {leftsep=0pt},
        hline{1,Z} = {0.08em}, 
        hline{2} = {0.05em},
        hline{41} = {dashed, 0.05em}, 
        row{1} = {font=\bfseries},
        rowsep = 1.1pt,
    }
    Concept & Train & Val & Test & $P(\text{conf} \mid \text{c})$ & Confounding concept \\
    Mustache & 6,642 (4.1\%) & 1,003 (5.0\%) & 772 (3.9\%) & 99.96\% & Male \\
    5\_o\_Clock\_Shadow & 18,177 (11.2\%) & 2,345 (11.8\%) & 1,994 (10.0\%) & 99.91\% & Male \\
    Wearing\_Lipstick & 76,437 (47.0\%) & 8,860 (44.6\%) & 10,418 (52.2\%) & 99.91\% & No\_Beard \\
    Heavy\_Makeup & 62,555 (38.4\%) & 7,751 (39.0\%) & 8,084 (40.5\%) & 99.91\% & No\_Beard \\
    Goatee & 10,337 (6.4\%) & 1,464 (7.4\%) & 915 (4.6\%) & 99.90\% & Male \\
    Sideburns & 9,156 (5.6\%) & 1,367 (6.9\%) & 926 (4.6\%) & 99.90\% & Male \\
    Wearing\_Necktie & 11,890 (7.3\%) & 1,443 (7.3\%) & 1,399 (7.0\%) & 99.76\% & Male \\
    Bald & 3,713 (2.3\%) & \textbf{411} (2.1\%) & 423 (2.1\%) & 99.63\% & Male \\
    Rosy\_Cheeks & 10,525 (6.5\%) & 1,358 (6.8\%) & 1,432 (7.2\%) & 99.26\% & No\_Beard \\
    Blond\_Hair & 24,267 (14.9\%) & 3,056 (15.4\%) & 2,660 (13.3\%) & 98.70\% & No\_Beard \\
    Wearing\_Earrings & 30,362 (18.7\%) & 3,789 (19.1\%) & 4,125 (20.7\%) & 97.81\% & No\_Beard \\
    Wearing\_Necklace & 19,764 (12.1\%) & 2,396 (12.1\%) & 2,753 (13.8\%) & 97.33\% & No\_Beard \\
    Arched\_Eyebrows & 43,278 (26.6\%) & 5,134 (25.8\%) & 5,678 (28.4\%) & 95.96\% & No\_Beard \\
    Bangs & 24,685 (15.2\%) & 2,915 (14.7\%) & 3,109 (15.6\%) & 95.13\% & No\_Beard \\
    Pale\_Skin & 7,005 (4.3\%) & 856 (4.3\%) & 840 (4.2\%) & 94.30\% & No\_Beard \\
    Attractive & 83,603 (51.4\%) & 10,332 (52.0\%) & 9,898 (49.6\%) & 93.19\% & Young \\
    Wavy\_Hair & 51,982 (31.9\%) & 5,495 (27.7\%) & 7,267 (36.4\%) & 92.18\% & No\_Beard \\
    High\_Cheekbones & 73,645 (45.2\%) & 8,926 (44.9\%) & 9,618 (48.2\%) & 91.19\% & No\_Beard \\
    Pointy\_Nose & 44,846 (27.6\%) & 5,660 (28.5\%) & 5,704 (28.6\%) & 89.29\% & No\_Beard \\
    Brown\_Hair & 33,192 (20.4\%) & 4,793 (24.1\%) & 3,587 (18.0\%) & 89.22\% & No\_Beard \\
    Double\_Chin & 7,571 (4.7\%) & 975 (4.9\%) & 913 (4.6\%) & 87.91\% & Male \\
    Smiling & 78,080 (48.0\%) & 9,602 (48.3\%) & 9,987 (50.0\%) & 87.83\% & No\_Beard \\
    Chubby & 9,389 (5.8\%) & 1,216 (6.1\%) & 1,058 (5.3\%) & 87.65\% & Male \\
    Oval\_Face & 46,101 (28.3\%) & 5,565 (28.0\%) & 5,901 (29.6\%) & 87.14\% & No\_Beard \\
    Mouth\_Slightly\_Open & 78,486 (48.2\%) & 9,573 (48.2\%) & 9,883 (49.5\%) & 86.78\% & No\_Beard \\
    Black\_Hair & 38,906 (23.9\%) & 4,144 (20.9\%) & 5,422 (27.2\%) & 86.42\% & Young \\
    Bushy\_Eyebrows & 23,386 (14.4\%) & 2,831 (14.3\%) & 2,586 (13.0\%) & 86.12\% & Young \\
    Young & 126,788 (77.9\%) & 14,832 (74.7\%) & 15,114 (75.7\%) & 85.88\% & No\_Beard \\
    Big\_Lips & 39,213 (24.1\%) & 3,044 (15.3\%) & 6,528 (32.7\%) & 85.37\% & Young \\
    Gray\_Hair & 6,896 (4.2\%) & 967 (4.9\%) & 636 (3.2\%) & 85.13\% & Male \\
    Straight\_Hair & 33,947 (20.9\%) & 4,085 (20.6\%) & 4,190 (21.0\%) & 85.09\% & No\_Beard \\
    Narrow\_Eyes & 18,869 (11.6\%) & 1,492 (7.5\%) & 2,968 (14.9\%) & 83.25\% & No\_Beard \\
    Blurry & 8,362 (5.1\%) & 940 (4.7\%) & 1,010 (5.1\%) & 82.82\% & No\_Beard \\
    No\_Beard & 135,779 (83.4\%) & 16,338 (82.2\%) & 17,041 (85.4\%) & 79.57\% & Young \\
    Eyeglasses & 10,521 (6.5\%) & 1,383 (7.0\%) & 1,289 (6.5\%) & 79.42\% & Male \\
    Receding\_Hairline & 13,040 (8.0\%) & 1,429 (7.2\%) & 1,694 (8.5\%) & 76.48\% & No\_Beard \\
    Big\_Nose & 38,341 (23.6\%) & 4,943 (24.9\%) & 4,232 (21.2\%) & 74.57\% & Male \\
    Bags\_Under\_Eyes & 33,280 (20.5\%) & 4,121 (20.7\%) & 4,045 (20.3\%) & 72.77\% & No\_Beard \\
    Wearing\_Hat & 8,039 (4.9\%) & 940 (4.7\%) & 839 (4.2\%) & 70.66\% & Young \\
    Male & 68,261 (41.9\%) & 8,458 (42.6\%) & 7,715 (38.7\%) & 63.30\% & Young \\
    \end{tblr}
\end{table}

%%%%
\clearpage
\subsubsection{Waterbirds}

The Waterbirds is a modified version of the Caltech-UCSD Birds-200-2011 (CUB) dataset~\cite{wah2011caltech}, where authors~\cite{sagawa2020distributionally} replaced image backgrounds to create spurious correlations between binary bird classes (\textit{landbird}, \textit{waterbird}) and backgrounds (\textit{land}, \textit{water}). This creates rare groups: \textit{landbird} on \textit{water} and \textit{waterbird} on \textit{land}. \Cref{tab:waterbirds_dataset} presents the distribution, showing that while training data is highly skewed, validation and test splits are balanced to rigorously test robustness. The dataset is available from the Stanford DRO repository under CC BY 4.0 license at \url{https://downloads.cs.stanford.edu/nlp/data/dro/waterbird_complete95_forest2water2.tar.gz}.

\textbf{Stratified sampling strategy.}
For Waterbirds, we use stratified sampling to ensure CAVs capture the background concept rather than the bird species. 
For the \textit{land} concept, positives are drawn from the majority class (\textit{landbird} on \textit{land}) and negatives from the same bird class but with the opposite background (\textit{landbird} on \textit{water}). 
Similarly, for \textit{water}, we contrast \textit{waterbird} on \textit{water} (positive) vs. \textit{waterbird} on \textit{land} (negative). 
Our preliminary experiments confirmed that naive random sampling produced CAVs aligned with bird features rather than the environment. To sum up, we investigate the CAV behavior when negative samples are drawn exclusively from the opposing bias group.

\begin{table}[h]
    \caption{\textbf{Waterbirds dataset distribution}. Total samples: train ($4,795$), val ($1,199$), test ($5,794$). 
    Concepts with fewer than $500$ samples in train or val splits are \textbf{bolded}. In contrast to the balanced validation and test sets, the training set exhibits extreme imbalance.}
    \label{tab:waterbirds_dataset}
    \vspace{0.5em}
    \centering
    \begin{tabular}{m{1.3cm} l c c c c}
    \toprule
    \textbf{Image} & 
    \textbf{Concept} & 
    \textbf{Confounded class} & 
    \textbf{Train data} & 
    \textbf{Val data} & 
    \textbf{Test data} \\
    \midrule
    % Row 1: Landbird on Water
    \includegraphics[width=0.08\textwidth]{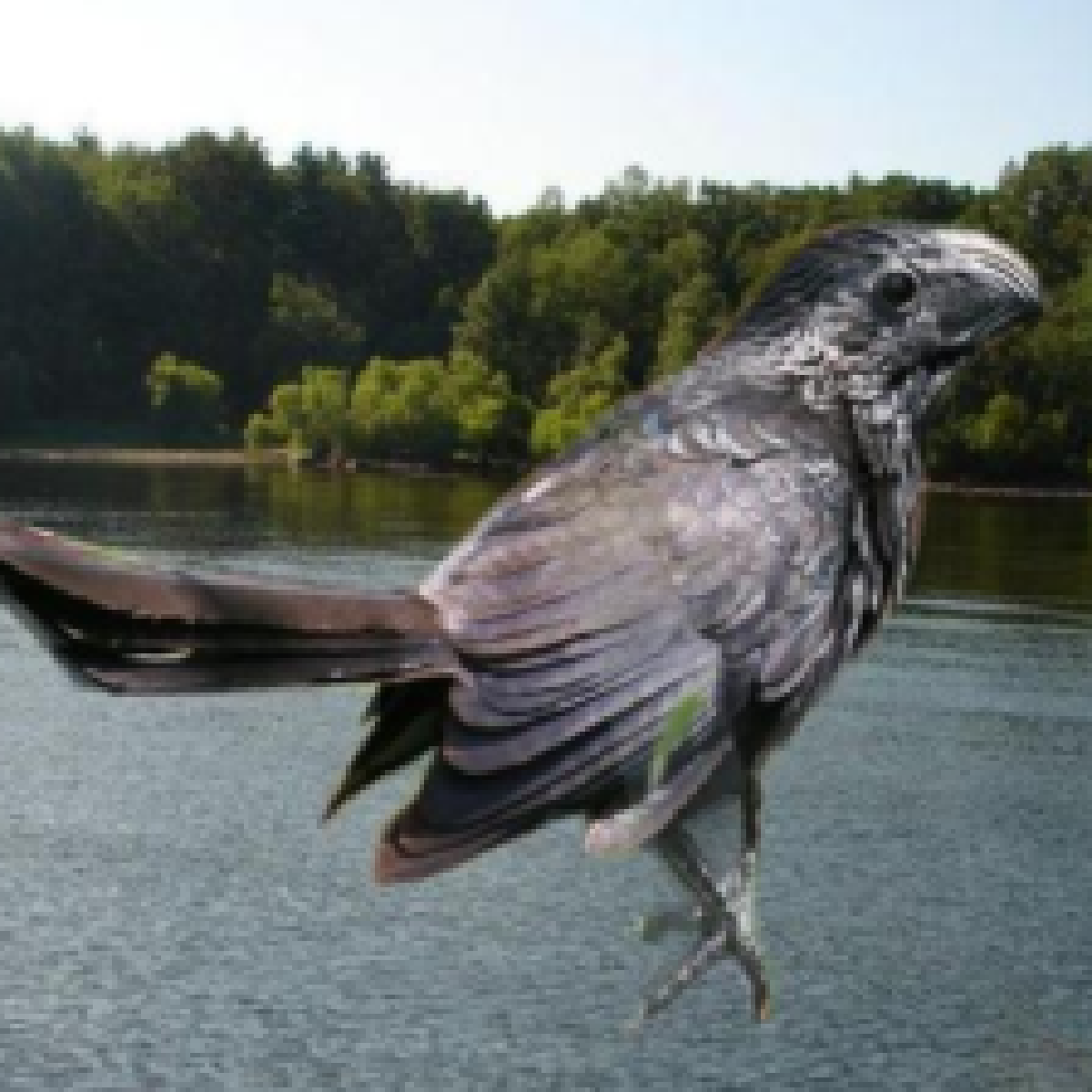} & 
    \shortstack[l]{\textbf{Water} \\ (with landbird)} & 
    landbird & 
    \textbf{184} ($5\%$) & \textbf{466} ($50\%$) & $2,255$ ($50\%$) \\
    % Row 2: Landbird on Land
    \includegraphics[width=0.08\textwidth]{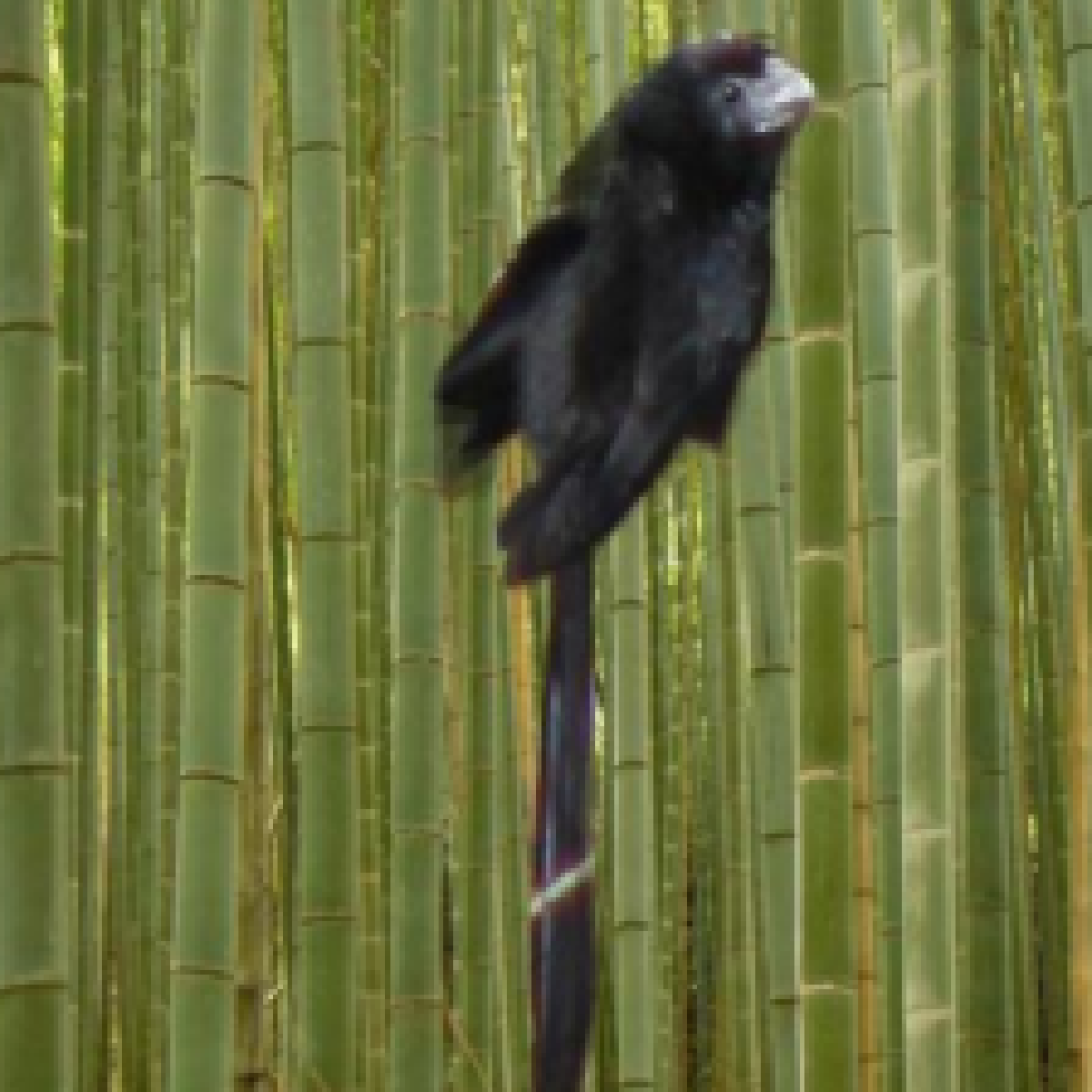} & 
    \shortstack[l]{\textbf{Land} \\ (with landbird)} & 
    --- & 
    $3,498$ ($95\%$) & \textbf{467} ($50\%$) & $2,255$ ($50\%$) \\
    % Row 3: Waterbird on Water
    \includegraphics[width=0.08\textwidth]{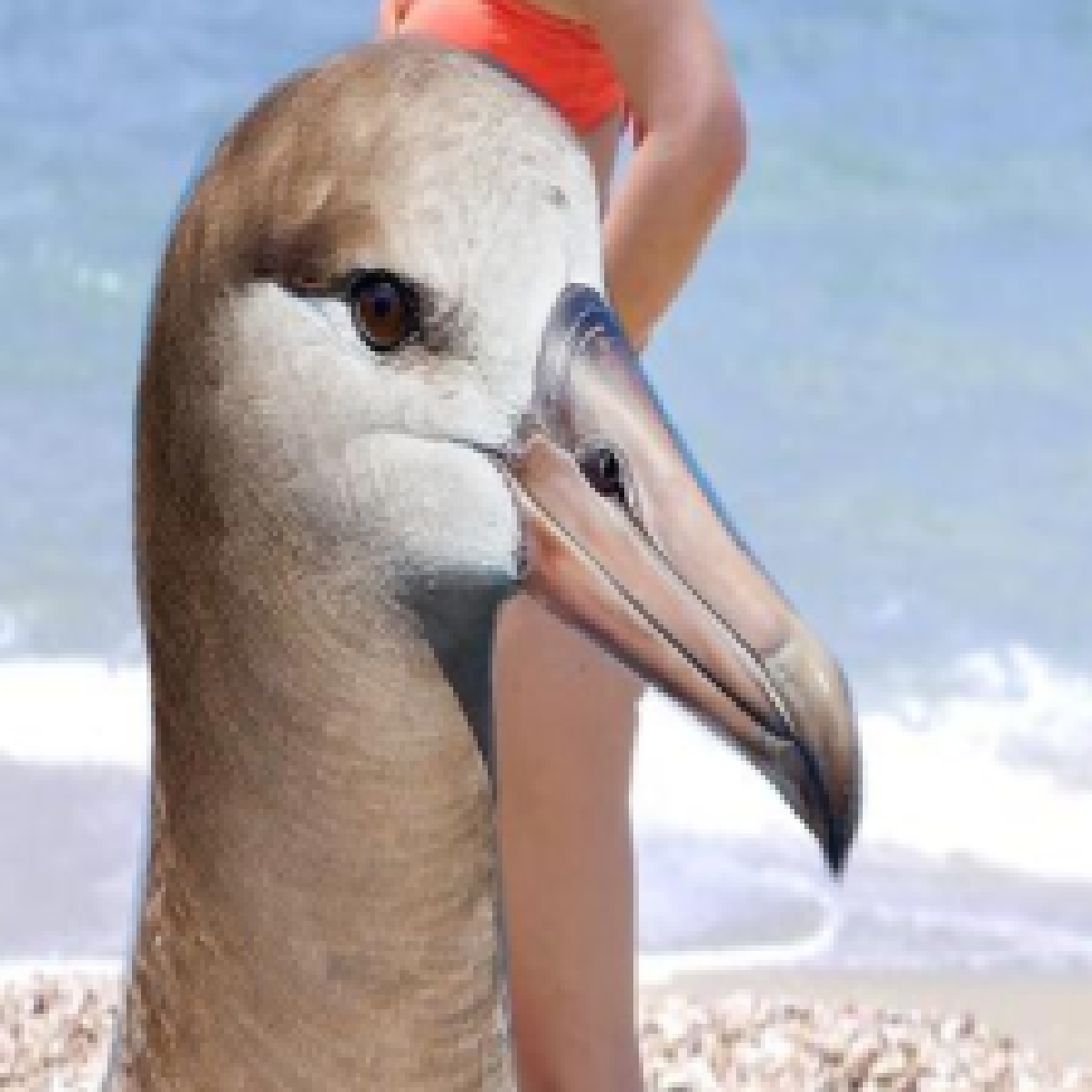} & 
    \shortstack[l]{\textbf{Water} \\ (with waterbird)} & 
    --- & 
    $1,057$ ($95\%$) & \textbf{133} ($50\%$) & $642$ ($50\%$) \\
    % Row 4: Waterbird on Land
    \includegraphics[width=0.08\textwidth]{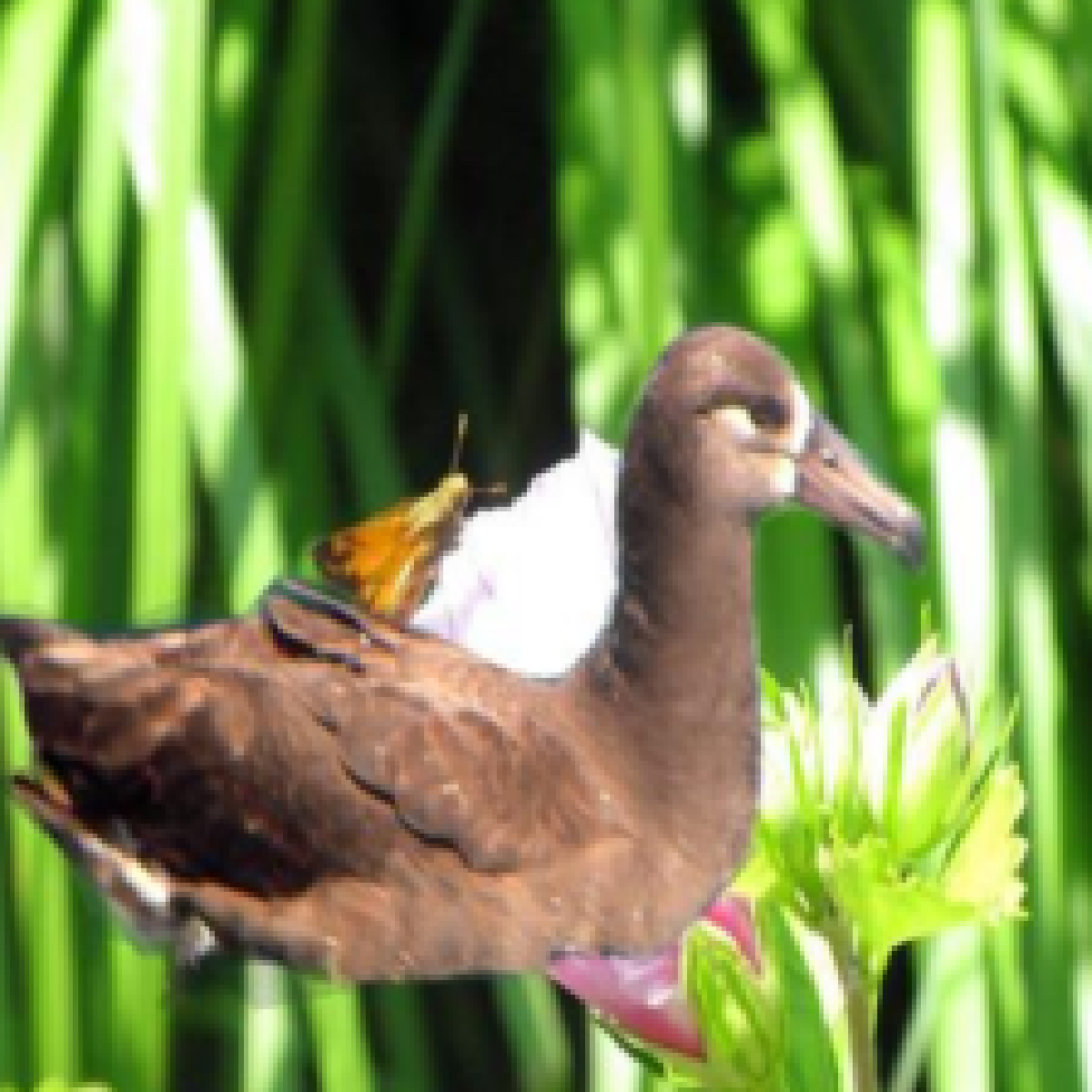} & 
    \shortstack[l]{\textbf{Land} \\ (with waterbird)} & 
    waterbird & 
    \textbf{56} ($5\%$) & \textbf{133} ($50\%$) & $642$ ($50\%$) \\
    \bottomrule
    \end{tabular}
\end{table}

\begin{figure}[b]
    \centering
    \includegraphics[width=0.9\textwidth]{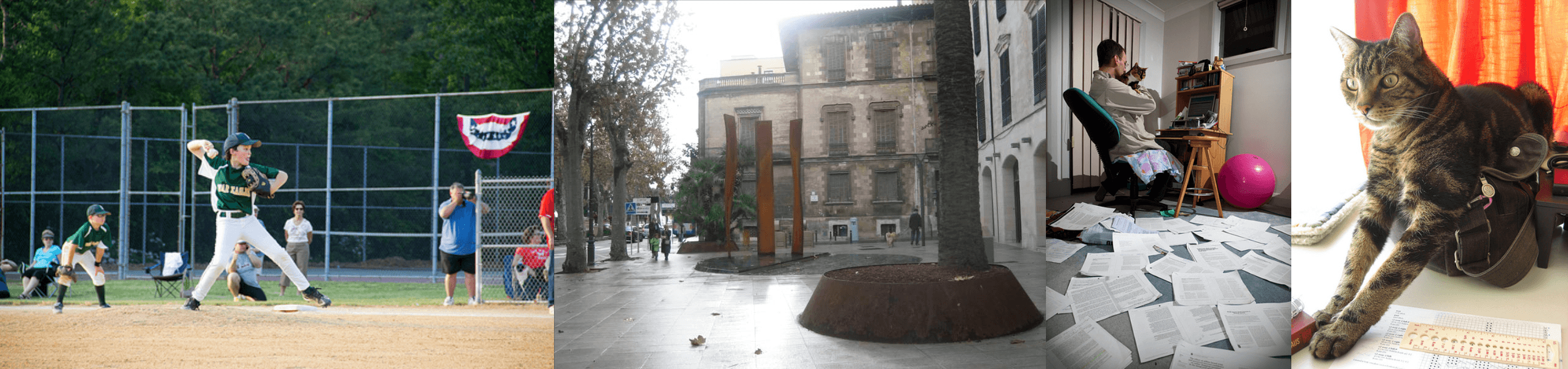}
    \caption{
    \textbf{Examples of images from the MetaShift dataset.} 
    Many of the samples are incorrectly labeled for both the class (\textit{dog} or \textit{car}) and the corresponding concept (\textit{shelf}, \textit{sofa}, \textit{bed}, and \textit{bike}). Here, the first two images are labeled as dogs with a concept \textit{shelf} and \textit{bike}, respectively. 
    Another two images are labeled as cats with a concept \textit{sofa} and \textit{bed}, respectively. 
    In addition to label mismatches, some images do not contain any of the concepts or classes claimed to be present.
    }
    \label{fig:metashift}
\end{figure}

\subsubsection{MetaShift}

This dataset was created with detecting data shift in mind~\citep{liang2022metashift}. It contains 410 classes (e.g., \textit{cat}, \textit{dog}, \textit{bus}, \textit{horse}) split into co-occurring objects in the image. For example, the \textit{cat} class is split into cats with \textit{keyboard}, \textit{sink}, or \textit{box}. We decided to exclude this data from our benchmark due to multiple instances of mislabeled data we found during manual exploration of the dataset. For example, images presented in \Cref{fig:metashift} show images that lack either the class (e.g., \textit{dog}) or the co-occurring concept (e.g., \textit{bike}). We acknowledge that we restricted our analysis to a subset of this dataset containing only cat and dog images. The dataset is available from \url{https://github.com/Weixin-Liang/MetaShift} under the MIT License.

%%%%%%
%\clearpage
\subsubsection{Counteranimal}
\label{app:counteranimal}

\begin{figure}[h]
    \centering
    \begin{minipage}{0.49\textwidth}
        \centering
        \includegraphics[height=4.5cm, keepaspectratio]{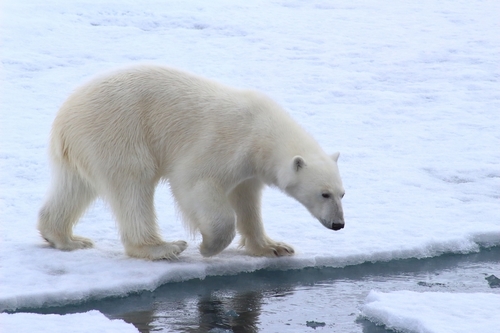}
    \end{minipage}
    \hfill
    \begin{minipage}{0.49\textwidth}
        \centering
        \includegraphics[height=4.5cm, keepaspectratio]{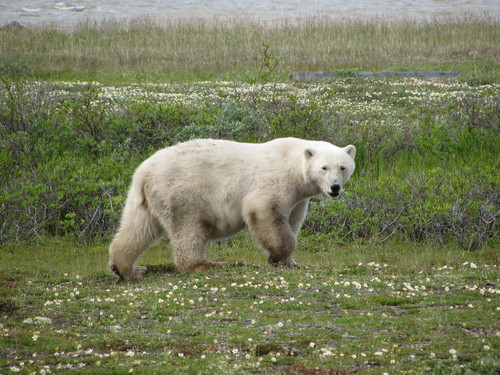}
    \end{minipage}
    \caption{\textbf{Example of Counteranimal dataset.} An animal is presented in its common environment (left) or with an atypical background (right).}
    \label{fig:counteranimal-example}
\end{figure}

The Counteranimal dataset~\citep{wang2024sober} was introduced as an additional validation for models from the CLIP family that highlights their biases and limitations not previously detected on datasets such as ImageNet~\citep{deng2009imagenet} or LAION~\citep{schuhmann2022laion}. It contains images of animals collected from the iNaturalist application, notably after manually filtering for low-quality or ambiguous-label images. The dataset contains 45 animals, whose labels align with ImageNet-1k classes, and is split into two parts: \textit{common} and \textit{uncommon}. The \textit{common} split contains images of animals in their typical habitat. Consequently, \textit{uncommon} split covers images of animals with an uncommon background, e.g., polar bears on the grass. \Cref{fig:counteranimal-example} shows an example. The authors verified that CLIP models perform decent classification in a zero-shot setup on \textit{common} split, but they also observed a significant drop in performance on \textit{uncommon} split, which highlights that CLIP bases its prediction partially on the context -- a background in the image -- rather than on the object in the image itself. The benchmark results on this dataset are in Appendix~\ref{app:counteranimal_results}. The dataset is available at \url{https://counteranimal.github.io/} under an unknown license. 

\textbf{Sampling strategy.} We use the \textit{common} split of Counteranimal dataset as a train and validation set in our experiments, with a random split between those in a ratio of 9:1. As a test set, we use all images from \textit{uncommon} split. This approach leverages the strengthening of the background bias in both the training and validation sets, which can later be removed using debiasing methods on the test set. The full statistics are provided in \Cref{tab:counteranimal-data}. Because the background labels are semantically similar (e.g., grass, green, forest), we decided to merge some labels as shown in \Cref{tab:counteranimal-concepts}. Further, we discarded classes that represent background types that cannot be explained by a more general group.

\begin{table}[h!]
    \caption{
    \textbf{Counteranimal dataset concept grouping.}
    }
    \label{tab:counteranimal-concepts}
    \vspace{0.5em}
    \centering
    \renewcommand{\arraystretch}{1.3}
    \begin{tabular}{l l}
    \toprule
    \textbf{grouped concept} & \textbf{original concepts} \\
    \midrule

    water & above water, underwater, water \\
    ground & earth, ground, rock, sand \\
    green & grass, green, shrub, tree, tree and grass \\
    snow & snow, white background \\
    sky & white or blue, sky \\
    not\_water & not in water, without water \\

    \bottomrule
    \end{tabular}
\end{table}

\begin{table}[h!]
    \caption{
    \textbf{Counteranimal dataset distribution}. 
    Counts for each class across splits, along with environment group annotations. Total samples: 10,774 (Train: 5,485; Validation: 610; Test: 4,679).
    }
    \label{tab:counteranimal-data}
    \vspace{0.5em}
    \centering
    \renewcommand{\arraystretch}{1.3}
    \begin{tabular}{l r l r l r l}
    \toprule
    \multirow{2}{*}{\textbf{Class}} &
    \multicolumn{2}{c}{\textbf{train data}} &
    \multicolumn{2}{c}{\textbf{validation data}} &
    \multicolumn{2}{c}{\textbf{test data}} \\
    \cmidrule(lr){2-3} \cmidrule(lr){4-5} \cmidrule(lr){6-7}
     & count & group & count & group & count & group \\
    \midrule

    African crocodile & 82 & ground & 9 & ground & 84 & green \\
    Arctic fox & 111 & snow & 12 & snow & 238 & green \\
    Vulture & 132 & sky & 15 & sky & 98 & green \\
    agama & 304 & ground & 34 & ground & 142 & green \\
    beaver & 93 & water & 10 & water & 100 & green \\
    bighorn & 85 & green & 9 & green & 95 & ground \\
    black stork & 73 & green & 8 & green & 149 & sky \\
    black swan & 184 & water & 20 & water & 106 & ground \\
    box turtle & 58 & green & 7 & green & 200 & ground \\
    brambling & 105 & green & 12 & green & 111 & sky \\
    bulbul & 110 & sky & 12 & sky & 185 & green \\
    bullfrog & 247 & water & 27 & water & 158 & not\_water \\
    centipede & 55 & snow & 6 & snow & 104 & green \\
    cicada & 130 & green & 14 & green & 88 & other \\
    common iguana & 45 & ground & 5 & ground & 120 & green \\
    dung beetle & 58 & ground & 7 & ground & 92 & other \\
    flamingo & 177 & water & 20 & water & 101 & sky \\
    garter snake & 70 & green & 8 & green & 249 & ground \\
    harvestman & 451 & green & 50 & green & 125 & ground \\
    hognose snake & 183 & ground & 20 & ground & 123 & green \\
    hyena & 311 & green & 35 & green & 100 & other \\
    ice bear & 112 & snow & 13 & snow & 109 & green \\
    jaguar & 59 & water & 6 & water & 226 & green \\
    king snake & 205 & ground & 23 & ground & 98 & green \\
    loggerhead & 198 & water & 22 & water & 91 & not\_water \\
    mink & 69 & green & 8 & green & 101 & water \\
    ostrich & 185 & ground & 21 & ground & 113 & water \\
    otter & 117 & water & 13 & water & 102 & green \\
    pelican & 209 & water & 23 & water & 98 & sky \\
    prairie chicken & 233 & green & 26 & green & 87 & snow \\
    ptarmigan & 51 & snow & 6 & snow & 107 & green \\
    red fox & 129 & green & 14 & green & 105 & other \\
    sea lion & 52 & ground & 6 & ground & 94 & water \\
    tarantula & 208 & ground & 23 & ground & 158 & green \\
    water ouzel & 234 & water & 26 & water & 159 & not\_water \\
    water snake & 136 & water & 15 & water & 163 & ground \\
    whiptail & 224 & ground & 25 & ground & 100 & other \\

    \bottomrule
    \end{tabular}
\end{table}

%%%%%%
\clearpage
\subsubsection{ImageNet-W (Watermark)}

\begin{figure}[h!]
    \centering
    \includegraphics[width=0.6\textwidth]{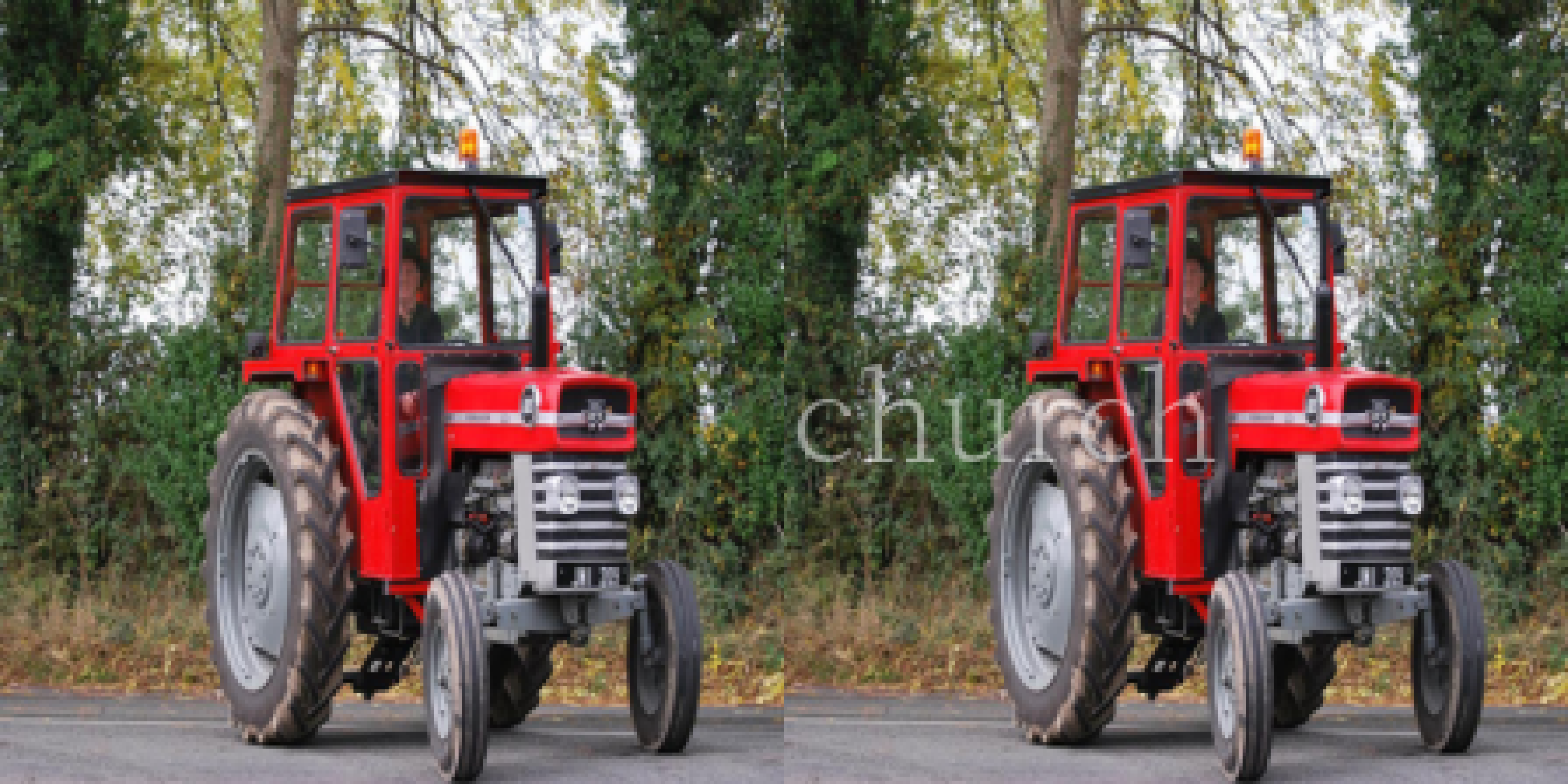}
    \caption{\textbf{Example of watermarking in ImageNet-W}. 
    \emph{Left}: A clean image of a \textit{Tractor}. 
    \emph{Right}: The same image infused with the watermark text \texttt{church}. The watermark overlays transparent text, which is clearly visible but does not obscure the underlying structural details of the object.}
    \label{fig:watermark}
\end{figure}

We use the ImageNet-1k dataset~\cite{deng2009imagenet} from Hugging Face \texttt{ILSVRC/imagenet-1k} (ImageNet Agreement), sampling $10$ diverse classes following \cite{baniecki2025explaining} to optimize computational efficiency without sacrificing evaluation depth. 
Our ImageNet-W dataset features a balanced distribution across 10 distinct classes: tractor, goldfish, cat, husky, banana, pizza, plane, ball, church, and iPod. 
It contains 11700 training samples, 1300 validation samples, and 500 test samples. 
Each individual class is uniformly represented by 1,170 training images, 130 validation images, and 50 test images, with these per-class quantities corresponding exactly to 10\% of the official dataset split.
To ensure robust evaluation, we repartition the official splits as follows:
(i) Test set: The official ImageNet Validation split ($50$ images per class).
(ii) The official training split is partitioned into new training ($90\%$) and validation ($10\%$) sets per class.
During downstream training, we augment each sample with its corresponding class watermark to increase the model's sensitivity to watermarks.
The watermark augmentation pipeline, adapted from \cite{li2023whac}, generates counterfactual pairs on the fly. The visual effect of this augmentation is presented in \Cref{fig:watermark}.

\textbf{Confounding watermarks.}
For downstream steering, we restrict our evaluation to high-degradation watermark-class pairs (\Cref{tab:imagenet_confounders}), defined as pairs in which the watermark text causes a significant drop in classification accuracy, serving as a strong confounder. 
Notably, we find that text watermarks do not degrade DINOv2 performance ($0\%$ accuracy drop). 
We hypothesize this is due to DINOv2's vision-only self-supervised objective, which, unlike text-aligned models such as CLIP and SigLIP, does not learn to attend to text hints. 
Therefore, we report steering results primarily for the text-aligned models where the artifact poses a genuine safety risk.

\begin{table}[h!]
    \caption{\textbf{ImageNet-W confounding pairs}. 
    The confounding watermark column indicates the specific text overlay used to induce spurious correlation. 
    The \textit{Degradation Accuracy Drop} reports the performance gap between clean and watermarked samples. DINOv2 is robust to watermarks.
    }
    \label{tab:imagenet_confounders}
    \vspace{0.5em}
    \centering
    \begin{tabular}{ll ccc}
    \toprule
    \multirow{2}{*}{\textbf{Visual class}} & \multirow{2}{*}{\textbf{\shortstack[l]{Confounding\\watermark (text)}}} & \multicolumn{3}{c}{\textbf{Degradation accuracy drop}} \\
    \cmidrule(lr){3-5}
     & & \textbf{CLIP} & \textbf{DINOv2} & \textbf{SigLIP} \\
    \midrule
    \emph{cat} & \emph{husky} & $42\%$ & $0\%$ & $16\%$ \\
    \emph{goldfish} & \emph{church} & $24\%$ & $0\%$ & $12\%$ \\
    \bottomrule
    \end{tabular}
\end{table}

%%%%%%%
% \clearpage
\subsubsection{ImageNet-C (Corruptions)}
\label{app:imagenetc}

\begin{figure}[ht]
    \centering
    \includegraphics[width=0.7\textwidth]{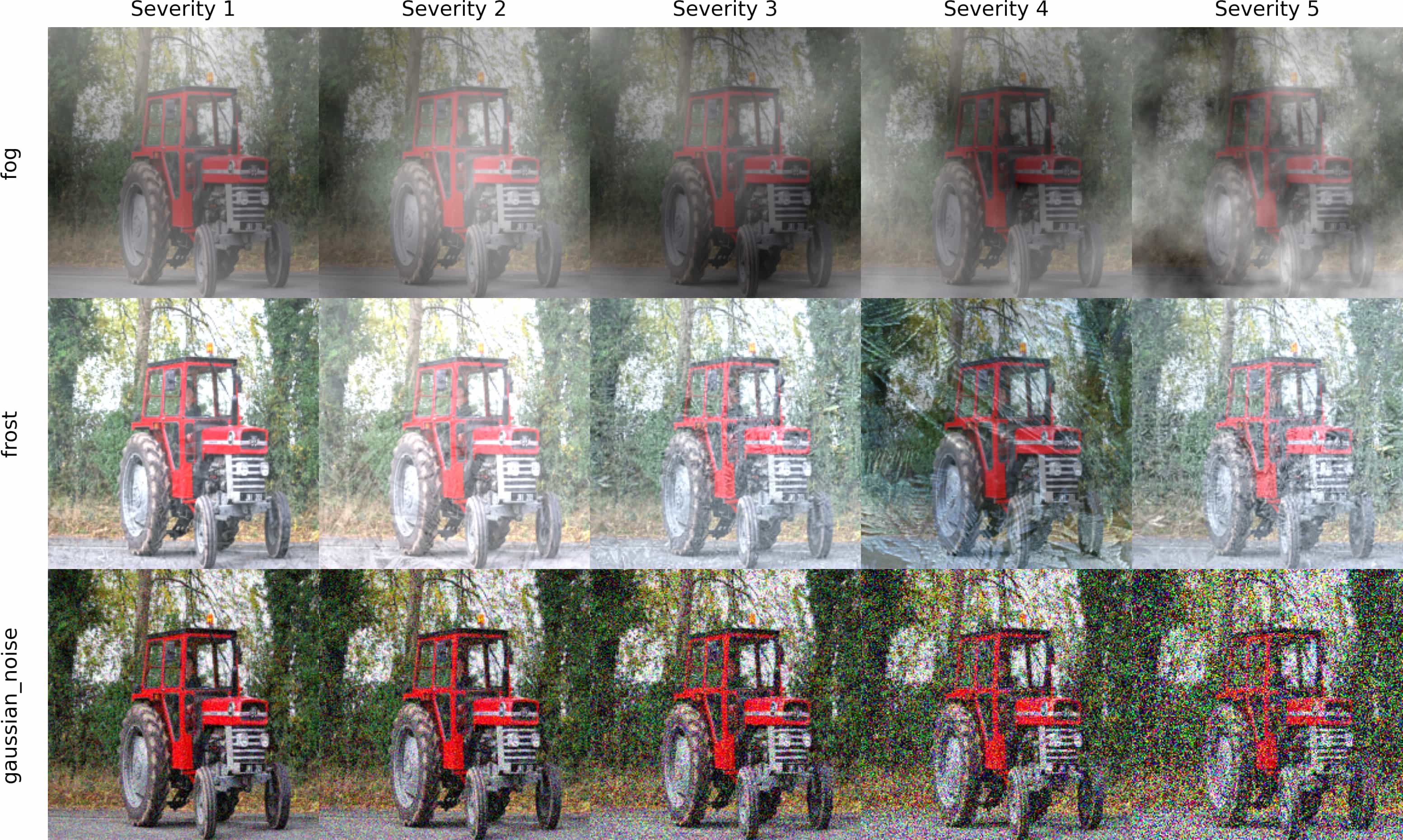}
    \caption{\textbf{Example of corruptions in ImageNet-C.} 
    We visualize three sample corruptions: \textit{Fog}, \textit{Frost}, and \textit{Gaussian Noise}. Each row displays the corruption across increasing severity levels.}
    \label{fig:corruption}
\end{figure}

\begin{figure}[ht]
    \centering
    \includegraphics[width=0.7\textwidth]{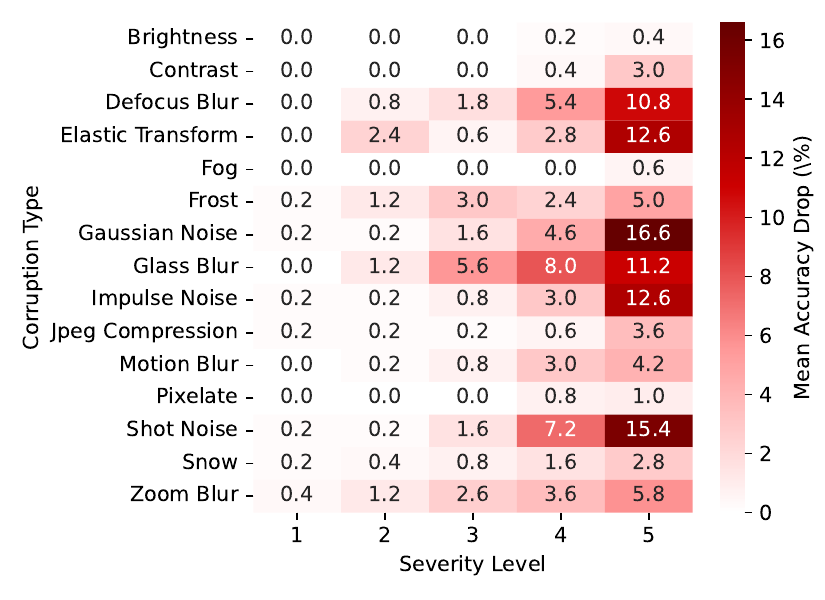}
    \caption{\textbf{Heatmap of corruptions $\times$ severity impact on accuracy drop for CLIP.} \textit{Gaussian Noise} visibly degrades performance from severity 3, whereas \textit{Shot Noise} degrades it from severity 4. The largest drop in accuracy is observed at severity level 5, but, as shown in \Cref{fig:corruption}, we exclude this level from our experiments because it alters images too drastically compared to lower severities.}
    \label{fig:corruption_impact}
\end{figure}

ImageNet-C shares a similar structure with ImageNet-W; however, rather than adding watermarks to the original ImageNet dataset, it applies one of 19 corruptions defined in~\cite{hendrycks2019benchmarking} and presented in \Cref{fig:corruption_impact}. Each corruption can be applied at five different severity levels, as visualized for three representative concepts in \Cref{fig:corruption}. Given the consistent performance drop under \textit{Shot Noise} and \textit{Gaussian Noise} at severity 4 for both concepts (\Cref{fig:corruption_impact}), we evaluate downstream tasks across five seeds for both corruptions and present the detailed results in \Cref{app:extended_results}. These accuracy drops remain consistent across various backbones for only those two corruptions. We exclude severity level 5 from our evaluation, as the severe degradation of the original image often obscures the target object, rendering classification impractical.

%%%%%%%
\clearpage
\subsection{Downstream Model Training}
\label{app:downstream_model_training}

We train lightweight classifiers $\psi$ on top of the frozen backbone representations $h$ to evaluate steering performance. We use the following openly available backbone models with default hyperparameters:
\begin{itemize}
    \item CLIP ViT-L/14~\cite{radford2021learning}: loaded via the clip Python library \\ \url{https://github.com/openai/CLIP} \texttt{ViT-L/14} (MIT License)
    \item DINOv2 ViT-B/14~\cite{oquab2024dinov2}: \texttt{facebookresearch/dinov2 — dinov2\_vitb14}, loaded via PyTorch Hub (Apache License 2.0, CC-BY-4.0 License)
    \item SigLIP-2 ViT-B/16~\cite{zhai2023sigmoid}: \texttt{google/siglip2-base-patch16-224} loaded from Hugging Face (Apache License 2.0)
\end{itemize}
All downstream models are trained using the AdamW optimizer with a cosine decay learning rate schedule and a linear warmup of $2$ epochs (initial $lr=10^{-6}$).
For every dataset, we train and evaluate two distinct architectures, selecting the one that maximizes validation performance:
\begin{itemize}
    \item linear probe (LP): A single linear layer mapping representations directly to class logits.
    \item multi-layer perceptron (MLP): A two-layer network with ReLU activation and Dropout, designed to capture nonlinear decision boundaries. The hidden dimension $d_{\text{hidden}}$ is set to the largest power of two less than the input dimension $d$ (e.g., for a ViT-L/14 with $d=768$, the hidden layer is set to $512$).
\end{itemize}
We perform a grid search to select the optimal configuration for each dataset, choosing the model with the highest F1 score on the validation set. 
The number of epochs is tuned per dataset to ensure convergence: 4 for ImageNet-W, 8 for Waterbirds, and 25 for ISIC. 
The search space and fixed training parameters are detailed in \Cref{tab:training_config}.
\begin{table}[H]
\centering
\caption{
    \textbf{Downstream training configuration.} 
    We performed a full sweep over learning rates and weight decay for both architectures to ensure robust baselines. 
}
\label{tab:training_config}
\vspace{0.5em}
\begin{tabular}{ll|lcc}
\toprule
\multicolumn{2}{c|}{\textbf{Fixed parameters}} & \multicolumn{3}{c}{\textbf{Hyperparameter grid}} \\
 & Value & & LP &  MLP \\
\midrule
optimizer & AdamW & learning rate & \{1e-2, 1e-3, 5e-4\} & \{1e-2, 1e-3, 5e-4\} \\
batch size & 256 & dropout & N/A &  \{0.0, 0.5\} \\
scheduler & cosine + warmup & weight decay & \{0.0, 1e-4\} & \{0.0, 1e-4\} \\
% min. learning rate & $1 \times 10^{-6}$ & dropout & N/A &  \{0.0, 0.5\} \\
\bottomrule
\end{tabular}
\end{table}

%%%%%%%%%%%%%%%%%%%%%%%%%%
\clearpage
\subsection{CAV Training Details}
\label{app:cav_training_details}

In this section, we detail the hyperparameter optimization protocols for the CAV extraction methods. 
Importantly, all performance metrics reported in the main paper are computed on strictly held-out validation sets.

We selected specific backbone versions for DINOv2 (ViT-B/14), CLIP (ViT-L/14), and SigLIP (ViT-B/16) for shared output dimensionality of $d=768$. This consistency facilitates fair CAV evaluations and permits a fixed SAE expansion rate across all models, simplifying the experimental setup.

All downstream models and SAE models were trained on NVIDIA A100. Concept Activation Vectors were trained on CPU, and vector metrics were similarly extracted using CPU resources. Steering metrics were extracted on NVIDIA A100 GPUs to accommodate the significant computational demands of the backbone models and downstream classifiers.

% \textbf{Reproducibility Statement}
% We provide the complete codebase for all results presented in this paper in the zip file. The \texttt{src/scripts} directory contains the code required to download datasets and compute metrics. Precalculated data is available as CSV files within the \texttt{data} directory, and the exact tables featured in the paper can be generated using \texttt{experiments/paper\_results/tables.ipynb}. The entire codebase is comprehensively documented.

\subsubsection{Optimization-Based Linear CAVs}
For methods requiring iterative optimization, specifically logistic regression (LR) and linear SVM, we adopt the robust training pipeline from \citep{kantamneni2025are}.\footnote{\url{https://github.com/JoshEngels/SAE-Probes}}
To accommodate varying concept frequencies, we employ an adaptive cross-validation strategy:
\begin{itemize}
    \item small datasets ($< 128$ samples): we use stratified $K$-fold CV (default $K=5$).
    \item large datasets ($\geq 128$ samples): we use a stratified shuffle split with a single validation fold, sized at $\min(0.2 \times N, 100)$.
\end{itemize}

\textbf{Grid search.} 
For all SVM- and LR-based CAV extraction methods, we optimize the regularization strength parameter $C$ over 20 logarithmically spaced values in the range $[10^{-3}, 10^{3}]$. 
All linear models are trained with:
\begin{itemize}
    \item class balancing: weights are adjusted inversely proportional to class frequencies to handle imbalances,
    \item intercept: fitting is enabled,
    \item convergence: a maximum of 10,000 iterations.
\end{itemize}
The optimal $C$ is selected to maximize AUC on validation splits from the training set. 
The final CAV is obtained by retraining on the full training set using the best $C$.
Additionally, S\&PTopK was tested with LR- (including $C$ optimization) and Stylist-based neuron selection. In \Cref{fig:main}, we refer to S\&PTopK LR as S\&PTopK\_probe to distinguish it from standard LR.
For SAE-AurA, we use $W_{\text{dec}}$ and $W_{\text{enc}}$ as the feature directions.
The whole grid was evaluated for each CAV independently. 
\subsubsection{Nonlinear Probes}

\textbf{SVC with RBF kernel (SVM-RBF).}
We standardize features and perform a grid search using 5-fold cross-validation over 105 parameter combinations:
\begin{itemize}
    \item $C$: 15 logarithmically spaced values in $[10^{-3}, 10^{3}]$.
    \item $\gamma$: A discrete set $\{10^{-4}, 5 \times 10^{-3}, 10^{-3}, 10^{-2}, 10^{-1},\texttt{scale}, \texttt{auto}\}$.
\end{itemize}

\textbf{TabPFN.}
We use the transformer-based TabPFN~\cite{hollmann2023tabpfn} version 2.5, which performs well on tabular datasets. 
As a pre-trained network, it requires no hyperparameter tuning, providing a deterministic, robust nonlinear baseline out of the box.

\subsubsection{SAE Density Search}
For density-based sparse autoencoder methods--SAE-Density and SAE-DensityCommon--the optimal activation density threshold $\tau$ varies significantly across concepts. We employ a 5-fold CV sweep to adaptively select this threshold:

\begin{enumerate}
    \item search space: we define a range of candidate density thresholds $d$ (from 0.3 to 1.0, i.e., features firing in more than $d \times 100\%$ of samples), with step size 0.01,
    \item evaluation: for each candidate $d$:
    \begin{enumerate}
        \item we compute the steering vector on the training fold using features filtered by $d$,
        \item we evaluate the vector's separation ability (AUC) on the validation fold,
    \end{enumerate}
    \item selection: we select the density $d^*$ that yields the highest mean validation AUC,
    \item refinement: the final vector is re-computed on the full dataset using the optimal density threshold $d^*$.
\end{enumerate}

%%%%
\clearpage
\subsection{SAE Implementation Details}
\label{app:sae_training}
We train Sparse Autoencoders (SAEs) with Top-K sparsity~\citep{gao2024scaling} and hyperparameter settings based on \citep{costa2025from}. Our implementation utilizes the \texttt{overcomplete} Python package\footnote{\url{https://github.com/jkminder/overcomplete}}, which provides efficient kernels for SAE training. 

\textbf{Architecture.}
We employ the Top-K SAE architecture, which enforces sparsity by strictly retaining only the $k$ largest activations in the latent code. Given an input $h \in \mathbb{R}^d$, the forward pass is defined as:
\begin{align}
    \mathbf{z}_{\text{pre}} &= \text{ReLU}(W_{\text{enc}}(h - \mathbf{b}_d)) + b_{\text{enc}}) \\
    \mathbf{z} &= \text{TopK}(\mathbf{z}_{\text{pre}}, k) \\
    \hat{h} &= W_{\text{dec}} \mathbf{z} + \mathbf{b}_d
\end{align}
where $W_{\text{enc}} \in \mathbb{R}^{m \times d}$ is the encoder, $W_{dec} \in \mathbb{R}^{d \times m}$ is the decoder (dictionary), $m = r \cdot d$ is the number of latent features determined by the expansion rate $r$, and $\hat{h}$ is a reconstructed input. 
The \texttt{overcomplete} implementation omits the explicit decoder bias parameter $\mathbf{b}_d$ in the final reconstruction step, which we acknowledge may influence the model's ability to recover the mean shift.

\textbf{Loss Function.} 
The training objective minimizes the reconstruction error combined with an auxiliary reanimation regularizer to mitigate dead features (latents that are rarely active):
\begin{equation}
    \mathcal{L} = \mathcal{L}_{\text{recon}} + \lambda \mathcal{L}_{\text{aux}}
\end{equation}
where $\mathcal{L}_{\text{recon}} = \frac{1}{n}\sum_{i=1}^{n} \|\mathbf{x}_i - \hat{\mathbf{x}}_i\|_2^2$ is the Mean Squared Error, and $\mathcal{L}_{\text{aux}}$ is the reanimation loss implemented in \texttt{overcomplete}. We set the auxiliary weight $\lambda = 10^{-7}$.

\textbf{Training data \& preprocessing.}
SAEs are trained on image embeddings extracted from the CC3M dataset~\citep{sharma2018conceptual}, a standard choice for pre-training vision models. CC3M is openly available via Hugging Face \texttt{pixparse/cc3m-wds} CC BY 4.0 license.
Validation is performed on ImageNet-1k embeddings from the vision model. 
Following standard practice,\footnote{\url{https://transformer-circuits.pub/2024/april-update/index.html}} we normalize embeddings before SAE training and rescale SAE outputs to the original magnitude during evaluation. 
Specifically, we normalize by the RMS of norms, scaling so that $\sqrt{\mathbb{E}[|x|^2]} = \sqrt{m}$, rather than the mean norm $\mathbb{E}[|x|] = \sqrt{m}$ described in the reference. 
Our primary motivation for this normalization is to keep the sparsity penalty $\lambda$ comparable across representation sizes $d$. 
Since we use a TopK SAE with no $\lambda$ parameter, the precise normalization criterion has little impact. 
We have empirically verified that training with such normalization yields marginally better reconstruction quality than without normalization, and therefore retain it. 

\textbf{Training configuration.}
We perform a grid search over learning rates and Top-K values (54 total combinations). Optimization uses AdamW (no weight decay) with gradient clipping ($1.0$). We employ the same cosine scheduler with linear warmup described in \Cref{app:downstream_model_training}. Detailed settings are provided in \Cref{tab:method_training_config}.

\begin{table}[h]
\centering
\caption{\textbf{SAE training configuration.} We utilize a fixed expansion rate while sweeping over learning rates and Top-K values. A large batch size and consistent seed ensure stable, reproducible training across 50 epochs.}
\label{tab:method_training_config}
\vspace{0.5em}
\begin{tabular}{ll|lc}
\toprule
\multicolumn{2}{c|}{\textbf{Fixed parameters}} & \multicolumn{2}{c}{\textbf{Hyperparameter grid}} \\
Setting & Value & Hyperparameter & Values \\
\midrule
optimizer & AdamW ($\epsilon=6.25\text{e-}10$) & learning rate & \{1e-3, 5e-4, 1e-4\} \\
batch size & 4096 & Top-K & \{32, 64, 128\} \\
epochs & 50 & & \\
scheduler & Cosine + 10-Epoch Warmup & & \\
grad clip & 1.0 & & \\
weight decay & 0 & & \\
expansion rate & 32 & & \\
seed & 42 & & \\
\bottomrule
\end{tabular}
\end{table}

\subsubsection{SAE Performance Metrics}

\Cref{tab:sae_performance} summarizes the performance of the selected SAEs across all datasets.
For model selection, we prioritized the trade-off between reconstruction fidelity ($R^2$) and sparsity ($L_0$), ultimately selecting Top-K = 64 for all reported experiments. 
We report:
\begin{itemize}
    \item $R^2$ score: the coefficient of determination for reconstruction quality.
    \item $L_0$ sparsity: the average number of non-zero active features per sample.
    \item dead feature ratio: The fraction of features that never activate across the entire training set.
\end{itemize}

\begin{table}[H]
    \caption{
    \textbf{SAE Performance across architectures and datasets.} 
    We report reconstruction quality ($R^2$) separately for concept-present ($\Aone$) and concept-absent ($\Azero$) splits to assess potential bias. For ImageNet-1k validation, we report only ($\Aone$), as it represents the entire validation set. For downstream tasks, we report results based on the validation set used in the CAV evaluation; however, we do not report the dead-features. This metric was recorded only during training.}
    \label{tab:sae_performance}
    \vspace{0.5em}
    \centering
    \begin{tabular}{l l c c c c}
    \toprule
    \multirow{2}{*}{\textbf{Dataset}} & \multirow{2}{*}{\textbf{Model}} & \multirow{2}{*}{$L_0$} & \multicolumn{2}{c}{$R^2$} & \multirow{2}{*}{\textbf{Dead Feature Ratio}} \\
    \cmidrule(lr){4-5}
     & & & $\Aone$ & $\Azero$ & \\
    \midrule
    \multirow{3}{*}{\textbf{ImageNet-1k}} 
      & CLIP & 64 & .91 & - & 5\% \\
      & DINOv2 & 64 & .80 & - & 0\% \\
      & SigLIP & 64 & .92 & - & 3\% \\
    \midrule
    \multirow{3}{*}{\textbf{CelebA}} 
      & CLIP & 64 & .89 & .89 & - \\
      & DINOv2 & 64 & .85 & .85 & - \\
      & SigLIP & 64 & .91 & .91 & - \\
    \midrule
    \multirow{3}{*}{\textbf{Waterbirds}} 
      & CLIP & 64 & .89 & .89 & - \\
      & DINOv2 & 64 & .71 & .70 & - \\
      & SigLIP & 64 & .91 & .90 & - \\
    \midrule
    \multirow{3}{*}{\textbf{ISIC}} 
      & CLIP & 64 & .86 & .86 & - \\
      & DINOv2 & 64 & .69 & .70 & - \\
      & SigLIP & 64 & .89 & .89 & - \\
    \midrule
    \multirow{3}{*}{\textbf{Counteranimal}} 
      & CLIP & 64 & .92 & .92 & - \\
      & DINOv2 & 64 & .77 & .77 & - \\
      & SigLIP & 64 & .94 & .93 & - \\
    \midrule
    \multirow{3}{*}{\textbf{ImageNet-C}} 
      & CLIP & 64 & .87 & .92 & - \\
      & DINOv2 & 64 & .76 & .84 & - \\
      & SigLIP & 64 & .91 & .93 & - \\
    \midrule
    \multirow{2}{*}{\textbf{ImageNet-W}} 
      & CLIP & 64 & .87 & .92 & - \\
      & SigLIP & 64 & .91 & .93 & - \\
      
    \bottomrule
    \end{tabular}
\end{table}

\subsection{Compute Resources}
\label{app:compute_resources}

We estimate the compute resources needed to reproduce all results in this paper for ca. 5K A100 GPU hours plus 10K CPU hours on a high-RAM machine. 
We effectively used $2{-}3\times$ more compute for preliminary and failed experiments.
Note that CAV methods use cached representations from the backbones and thus are highly efficient to compute on a CPU; see specific compute times in \Cref{tbl:results_waterbirds_clip,tbl:results_waterbirds_dinov2,tbl:results_waterbirds_siglip}. 
In contrast, training downstream models and SAEs, as well as computing the metrics, was performed on a~GPU.

%%%%%%%%%%%%%%%%
\clearpage
\section{Extended Metric Results}
\label{app:extended_results}

Tables~\ref{tbl:results_celeba}--\ref{tbl:results_imagenet_c_siglip} extend the evaluations from \Cref{sec:results} by reporting metrics for all three vision backbones across the datasets introduced in \Cref{sec:datasets}. 
Metrics are aggregated across the concepts specified for each dataset in Appendix~\ref{app:dataset_distribution}. For ImageNet-W and ImageNet-C, results are further aggregated over five random seeds. We report the mean $\pm$ two standard errors (2SE) across these concepts or seeds for ImageNet-W/C.
The counteranimals results are presented in \Cref{app:counteranimal_results}.
In addition, we provide a critical difference ranking in \Cref{app:cdd} to determine which CelebA-based method results in the minimum collateral damage.
Consistent with Tables~\ref{tbl:results_dinov2}--\ref{tbl:results_watermark}, we report only the best-performing variant for each method—SAE-AurA (Encoder/Decoder), S\&PTopK (LR/Stylist), and PCA/PosPCA—selected based on overall performance. 

\begin{table}[h]
    \centering
    \footnotesize
    \caption{\textbf{Evaluating methods for steering image representations across two backbones CLIP nd SigLIP on the CelebA dataset.}
    Expanding on \Cref{tbl:results_dinov2}, we compare performance across two additional backbones: CLIP and DINOv2 backbone models.
    Columns report Concept Detection (AUC), Cross-Concept Robustness (CCR), Maximum Similarity (MS), and Collateral Damage (CD). Values represent the mean $\pm$ two standard errors across concepts, and we indicate the best method in each evaluation metric in \textbf{Bold} while the second best are \underline{underlined}. 
    Results on the other two backbones overlap with previous results, indicating that SVM and LR have comparable performance to nonlinear models and also provide better disentangled representations than statistical methods or even SAE-based version of the methods.
    }
    \label{tbl:results_celeba}
    \vspace{0.5em}
    \begin{tblr}{
      colspec = {lcccccccccc},
      column{1} = {leftsep=0pt},
      vline{2,7} = {dashed},
      hline{8,17,19} = {dashed},
      row{5,9,11,13,16} = {bg=black!10},
      rowsep=1.5pt,
      colsep = 2.6pt
    }
    & \SetCell[c=5]{c} \textbf{CLIP} & & & & & \SetCell[c=5]{c} \textbf{SigLIP} & & & & \\
    \textbf{Method} & AUC ($\uparrow$) & CCR ($\uparrow$) & MS ($\downarrow$) & F1 ($\uparrow$) & CD ($\downarrow$) & AUC ($\uparrow$) & CCR ($\uparrow$) & MS ($\downarrow$) & F1 ($\uparrow$) & CD ($\downarrow$) \\
    \hline
    Linear SVM & $\mathbf{.90}_{\pm.03}$ & $\mathbf{.94}_{\pm.01}$ & $\underline{.35}_{\pm.05}$ & $\mathbf{.59}_{\pm.06}$ & $0.20_{\pm0.21}$ & $\mathbf{.91}_{\pm.03}$ & $\mathbf{.94}_{\pm.02}$ & $\underline{.32}_{\pm.06}$ & $\mathbf{.62}_{\pm.06}$ & $0.32_{\pm0.37}$ \\
    LR & $\mathbf{.90}_{\pm.03}$ & $.89_{\pm.03}$ & $.40_{\pm.06}$ & $\mathbf{.59}_{\pm.06}$ & $1.23_{\pm1.30}$ & $\mathbf{.91}_{\pm.02}$ & $.88_{\pm.04}$ & $.34_{\pm.06}$ & $\underline{.61}_{\pm.06}$ & $1.27_{\pm1.82}$ \\
    \hspace{5pt}\includegraphics[width=6pt]{figures_new/subdirectory_arrow_right.png} SAE & $\underline{.87}_{\pm.03}$ & $.90_{\pm.02}$ & $.42_{\pm.07}$ & $\underline{.53}_{\pm.07}$ & $0.21_{\pm0.16}$ & $\underline{.88}_{\pm.03}$ & $\underline{.89}_{\pm.03}$ & $.35_{\pm.06}$ & $.56_{\pm.07}$ & $0.47_{\pm0.62}$ \\
    SAS & $.80_{\pm.04}$ & $.86_{\pm.03}$ & $.61_{\pm.08}$ & $.47_{\pm.06}$ & $5.69_{\pm3.14}$ & $.85_{\pm.04}$ & $.85_{\pm.03}$ & $.64_{\pm.07}$ & $.53_{\pm.07}$ & $5.70_{\pm4.55}$ \\
    S\&PTopK & $.80_{\pm.04}$ & $\underline{.92}_{\pm.02}$ & $\mathbf{.24}_{\pm.04}$ & $.48_{\pm.07}$ & $\mathbf{0.00}_{\pm0.01}$ & $.84_{\pm.04}$ & $\mathbf{.94}_{\pm.02}$ & $\mathbf{.26}_{\pm.05}$ & $.53_{\pm.07}$ & $\mathbf{0.03}_{\pm0.02}$ \\
    DiffMean & $.85_{\pm.03}$ & $.87_{\pm.03}$ & $.77_{\pm.06}$ & $.51_{\pm.06}$ & $17.6_{\pm5.5}$ & $.86_{\pm.03}$ & $.86_{\pm.03}$ & $.79_{\pm.06}$ & $.53_{\pm.06}$ & $11.3_{\pm3.8}$ \\
    \hspace{5pt}\includegraphics[width=6pt]{figures_new/subdirectory_arrow_right.png} SAE & $.85_{\pm.03}$ & $.86_{\pm.02}$ & $.78_{\pm.06}$ & $.51_{\pm.06}$ & $17.7_{\pm5.4}$ & $.86_{\pm.03}$ & $.85_{\pm.03}$ & $.80_{\pm.06}$ & $.53_{\pm.06}$ & $10.4_{\pm3.5}$ \\
    DiffMedian & $.84_{\pm.03}$ & $.88_{\pm.03}$ & $.78_{\pm.06}$ & $.50_{\pm.06}$ & $18.3_{\pm5.6}$ & $.86_{\pm.03}$ & $.87_{\pm.03}$ & $.80_{\pm.06}$ & $.53_{\pm.06}$ & $11.3_{\pm3.7}$ \\
    \hspace{5pt}\includegraphics[width=6pt]{figures_new/subdirectory_arrow_right.png} SAE & $.77_{\pm.04}$ & $.79_{\pm.04}$ & $.72_{\pm.06}$ & $.44_{\pm.06}$ & $2.74_{\pm2.19}$ & $.81_{\pm.03}$ & $.77_{\pm.04}$ & $.83_{\pm.06}$ & $.48_{\pm.06}$ & $12.1_{\pm3.4}$ \\
    FastCAV & $.85_{\pm.03}$ & $.87_{\pm.03}$ & $.77_{\pm.06}$ & $.51_{\pm.06}$ & $17.6_{\pm5.5}$ & $.86_{\pm.03}$ & $.86_{\pm.03}$ & $.79_{\pm.06}$ & $.53_{\pm.06}$ & $11.3_{\pm3.8}$ \\
    \hspace{5pt}\includegraphics[width=6pt]{figures_new/subdirectory_arrow_right.png} SAE & $.85_{\pm.03}$ & $.86_{\pm.02}$ & $.78_{\pm.06}$ & $.51_{\pm.06}$ & $17.7_{\pm5.4}$ & $.86_{\pm.03}$ & $.85_{\pm.03}$ & $.80_{\pm.06}$ & $.53_{\pm.06}$ & $10.4_{\pm3.5}$ \\
    PatCAV & $.85_{\pm.03}$ & $.87_{\pm.03}$ & $.77_{\pm.06}$ & $.51_{\pm.06}$ & $17.6_{\pm5.5}$ & $.86_{\pm.03}$ & $.86_{\pm.03}$ & $.79_{\pm.06}$ & $.53_{\pm.06}$ & $11.3_{\pm3.8}$ \\
    AurA & $.86_{\pm.03}$ & $.86_{\pm.03}$ & $.78_{\pm.05}$ & $.52_{\pm.06}$ & $0.58_{\pm0.32}$ & $.87_{\pm.03}$ & $.85_{\pm.03}$ & $.80_{\pm.05}$ & $.55_{\pm.06}$ & $0.24_{\pm0.11}$ \\
    \hspace{5pt}\includegraphics[width=6pt]{figures_new/subdirectory_arrow_right.png} SAE & $.83_{\pm.03}$ & $.73_{\pm.03}$ & $.94_{\pm.01}$ & $.48_{\pm.06}$ & $\underline{0.12}_{\pm0.05}$ & $.86_{\pm.03}$ & $.73_{\pm.04}$ & $.92_{\pm.02}$ & $.54_{\pm.06}$ & $\underline{0.04}_{\pm0.02}$ \\
    PCA & $.75_{\pm.04}$ & $.74_{\pm.03}$ & $.97_{\pm.01}$ & $.42_{\pm.07}$ & $49.5_{\pm4.8}$ & $.75_{\pm.04}$ & $.73_{\pm.04}$ & $.98_{\pm.01}$ & $.43_{\pm.06}$ & $32.9_{\pm3.1}$ \\
    LAT & $.63_{\pm.02}$ & $.61_{\pm.05}$ & $.91_{\pm.02}$ & $.35_{\pm.06}$ & $33.9_{\pm7.6}$ & $.66_{\pm.03}$ & $.50_{\pm.05}$ & $.94_{\pm.02}$ & $.38_{\pm.06}$ & $25.1_{\pm5.3}$ \\
    {\color{gray} SVM-RBF} & {\color{gray} $\mathbf{.90}_{\pm.03}$} & -- & -- & -- & -- & {\color{gray} $\mathbf{.91}_{\pm.02}$} & -- & -- & -- & -- \\
    {\color{gray} TabPFN} & {\color{gray} $\mathbf{.90}_{\pm.03}$} & -- & -- & -- & -- & {\color{gray} $\mathbf{.91}_{\pm.03}$} & -- & -- & -- & -- \\
    \hline
    \end{tblr}
\end{table}

\begin{table}[h]
    \centering
    \footnotesize
    \caption{\textbf{Evaluating methods for steering image representations of CLIP and SigLIP models on the ISIC dataset.} 
    Expanding on \Cref{tbl:results_dinov2}, we compare performance across two additional backbones: CLIP and DINOv2 backbone models. Columns report Concept Detection (AUC), Cross-Concept Robustness (CCR), Maximum Cosine Similarity (MS), Downstream Concept Detection (F1), and Collateral Damage (CD). We indicate the best method for each evaluation metric in \textbf{Bold}, while the second-best methods are \underline{underlined}.
    }
    \label{tbl:results_isic}
    \vspace{0.5em}
    \begin{tblr}{
      colspec = {lcccccccccc},
      column{1} = {leftsep=0pt},
      vline{2,7} = {dashed},
      hline{8,17} = {dashed},
      row{5,9,11,13,16} = {bg=black!10},
      rowsep=1.5pt,
      colsep = 2.5pt
    }
    & \SetCell[c=5]{c} \textbf{CLIP} & & & & & \SetCell[c=5]{c} \textbf{SigLIP} & & & & \\
    \textbf{Method} & AUC ($\uparrow$) & CCR ($\uparrow$) & MS ($\downarrow$) & F1 ($\uparrow$) & CD ($\downarrow$) & AUC ($\uparrow$) & CCR ($\uparrow$) & MS ($\downarrow$) & F1 ($\uparrow$) & CD ($\downarrow$) \\
    \hline
    Linear SVM & $\mathbf{.90}_{\pm.11}$ & $\underline{.97}_{\pm.03}$ & $\underline{.10}_{\pm.05}$ & $\mathbf{.80}_{\pm.14}$ & $4.14_{\pm3.63}$ & $\mathbf{.89}_{\pm.11}$ & $.95_{\pm.05}$ & $\underline{.11}_{\pm.08}$ & $\underline{.78}_{\pm.15}$ & $1.67_{\pm2.31}$ \\
    LR & $\underline{.89}_{\pm.11}$ & $\underline{.97}_{\pm.03}$ & $.11_{\pm.04}$ & $\underline{.79}_{\pm.13}$ & $4.65_{\pm3.98}$ & $\mathbf{.89}_{\pm.11}$ & $.94_{\pm.06}$ & $\mathbf{.10}_{\pm.06}$ & $\mathbf{.79}_{\pm.16}$ & $1.76_{\pm2.63}$ \\
    \hspace{5pt}\includegraphics[width=6pt]{figures_new/subdirectory_arrow_right.png} SAE & $.83_{\pm.14}$ & $\mathbf{.99}_{\pm.01}$ & $.17_{\pm.08}$ & $.73_{\pm.16}$ & $\underline{1.55}_{\pm2.47}$ & $\underline{.83}_{\pm.14}$ & $\mathbf{.98}_{\pm.02}$ & $.17_{\pm.06}$ & $.74_{\pm.16}$ & $5.12_{\pm3.26}$ \\
    SAS & $.71_{\pm.14}$ & $\underline{.97}_{\pm.04}$ & $.30_{\pm.11}$ & $.66_{\pm.15}$ & $\mathbf{1.25}_{\pm8.64}$ & $.73_{\pm.07}$ & $\underline{.97}_{\pm.03}$ & $.34_{\pm.15}$ & $.68_{\pm.13}$ & $2.03_{\pm4.86}$ \\
    S\&PTopK & $.74_{\pm.12}$ & $\mathbf{.99}_{\pm.01}$ & $\mathbf{.05}_{\pm.02}$ & $.64_{\pm.18}$ & $1.95_{\pm1.56}$ & $.79_{\pm.12}$ & $\underline{.97}_{\pm.04}$ & $.12_{\pm.07}$ & $.70_{\pm.17}$ & $\underline{0.62}_{\pm1.34}$ \\
    DiffMean & $.81_{\pm.13}$ & $.94_{\pm.05}$ & $.51_{\pm.29}$ & $.70_{\pm.14}$ & $4.36_{\pm8.19}$ & $.82_{\pm.10}$ & $.94_{\pm.05}$ & $.53_{\pm.25}$ & $.71_{\pm.14}$ & $7.70_{\pm10.6}$ \\
    \hspace{5pt}\includegraphics[width=6pt]{figures_new/subdirectory_arrow_right.png} SAE & $.78_{\pm.13}$ & $.94_{\pm.07}$ & $.56_{\pm.26}$ & $.68_{\pm.14}$ & $2.75_{\pm4.17}$ & $.80_{\pm.09}$ & $.94_{\pm.05}$ & $.51_{\pm.23}$ & $.70_{\pm.14}$ & $3.16_{\pm5.55}$ \\
    DiffMedian & $.80_{\pm.13}$ & $.95_{\pm.04}$ & $.52_{\pm.29}$ & $.70_{\pm.14}$ & $2.78_{\pm7.52}$ & $.81_{\pm.10}$ & $.94_{\pm.05}$ & $.52_{\pm.27}$ & $.71_{\pm.14}$ & $8.44_{\pm10.5}$ \\
    \hspace{5pt}\includegraphics[width=6pt]{figures_new/subdirectory_arrow_right.png} SAE & $.64_{\pm.08}$ & $.93_{\pm.07}$ & $.70_{\pm.04}$ & $.64_{\pm.12}$ & $2.83_{\pm9.07}$ & $.77_{\pm.12}$ & $.93_{\pm.04}$ & $.41_{\pm.23}$ & $.59_{\pm.31}$ & $3.58_{\pm3.78}$ \\
    FastCAV & $.81_{\pm.13}$ & $.94_{\pm.05}$ & $.51_{\pm.29}$ & $.70_{\pm.14}$ & $4.36_{\pm8.19}$ & $.82_{\pm.10}$ & $.94_{\pm.05}$ & $.53_{\pm.25}$ & $.71_{\pm.14}$ & $7.70_{\pm10.6}$ \\
    \hspace{5pt}\includegraphics[width=6pt]{figures_new/subdirectory_arrow_right.png} SAE & $.78_{\pm.13}$ & $.94_{\pm.07}$ & $.56_{\pm.26}$ & $.68_{\pm.14}$ & $2.75_{\pm4.17}$ & $.80_{\pm.09}$ & $.94_{\pm.05}$ & $.51_{\pm.23}$ & $.70_{\pm.14}$ & $3.16_{\pm5.55}$ \\
    PatCAV & $.81_{\pm.13}$ & $.94_{\pm.05}$ & $.51_{\pm.29}$ & $.70_{\pm.14}$ & $4.36_{\pm8.19}$ & $.82_{\pm.10}$ & $.94_{\pm.05}$ & $.53_{\pm.25}$ & $.71_{\pm.14}$ & $7.70_{\pm10.6}$ \\
    AurA & $.80_{\pm.12}$ & $.93_{\pm.05}$ & $.61_{\pm.18}$ & $.70_{\pm.14}$ & $2.76_{\pm4.58}$ & $\underline{.83}_{\pm.10}$ & $.91_{\pm.08}$ & $.64_{\pm.16}$ & $.71_{\pm.14}$ & $3.35_{\pm5.45}$ \\
    \hspace{5pt}\includegraphics[width=6pt]{figures_new/subdirectory_arrow_right.png} SAE & $.76_{\pm.11}$ & $.86_{\pm.12}$ & $.88_{\pm.09}$ & $.69_{\pm.12}$ & $3.73_{\pm4.71}$ & $.77_{\pm.08}$ & $.87_{\pm.09}$ & $.83_{\pm.10}$ & $.69_{\pm.14}$ & $\mathbf{0.43}_{\pm4.28}$ \\
    PCA & $.54_{\pm.05}$ & $.86_{\pm.22}$ & $1.00_{\pm.00}$ & $.45_{\pm.13}$ & $2.50_{\pm1.80}$ & $.49_{\pm.02}$ & $.67_{\pm.17}$ & $.99_{\pm.00}$ & $.57_{\pm.10}$ & $1.91_{\pm3.62}$ \\
    LAT & $.53_{\pm.05}$ & $.84_{\pm.05}$ & $.99_{\pm.00}$ & $.45_{\pm.14}$ & $2.70_{\pm1.62}$ & $.49_{\pm.02}$ & $.65_{\pm.16}$ & $.99_{\pm.00}$ & $.56_{\pm.11}$ & $1.27_{\pm3.07}$ \\
    \hline
    \end{tblr}
\end{table}

\begin{table}[h]
    \centering
    \footnotesize
    \caption{\textbf{Evaluating methods for steering image representations of CLIP model on the Waterbirds dataset.} 
    Values show the results for both concepts (``land'' on the left and ``water'' on the right). We report computation time of each method (Time) in seconds. For MS, we report absolute values because most methods yield negative scores, indicating that the CAVs for ``land'' and ``water'' are nearly antipodal, with the exception of AurA, which has non-negative values. We attribute AurA's performance to its targeted use of negative samples during training. It utilizes them strictly for neuron filtering, steering only the isolated neurons to increase concept presence, rather than actively pushing the representation away from the negative concept.
    For AurA SAE, CCR results exceed 1.0, demonstrating that orthogonalization improved performance beyond the original score.
    }
    \label{tbl:results_waterbirds_clip}
    \vspace{0.5em}
    \begin{tblr}{
      colspec = {lcccccc},
      column{1} = {leftsep=0pt},
      vline{2} = {dashed},
      hline{7,16} = {dashed},
      row{4,8,10,12,15} = {bg=black!10},
      rowsep=1.5pt
    }
    \textbf{Method} & AUC ($\uparrow$) & CCR ($\uparrow$) & MS ($\downarrow$) & F1 ($\uparrow$) & CD ($\downarrow$) & Time (s) \\
    \hline
    Linear SVM & $\mathbf{.98}/\mathbf{.98}$ & $.96/.86$ & $-.70/-.70$ & $\mathbf{.92}/\mathbf{.92}$ & $5.63/4.73$ & $0.49/1.18$ \\
    LR & $\mathbf{.98}/\mathbf{.98}$ & $.95/.81$ & $-.71/-.71$ & $\mathbf{.92}/\mathbf{.92}$ & $5.21/4.80$ & $0.53/0.83$ \\
    \hspace{5pt}\includegraphics[width=6pt]{figures_new/subdirectory_arrow_right.png} SAE & $\mathbf{.98}/\underline{.97}$ & $.80/.72$ & $-.83/-.83$ & $.89/\underline{.89}$ & $5.01/3.56$ & $1.08/1.92$ \\
    SAS & $.91/.91$ & $.87/.82$ & $-.57/-.57$ & $.82/.84$ & $18.0/2.76$ & $0.86/1.60$ \\
    S\&PTopK & $\mathbf{.98}/\underline{.97}$ & $.97/.95$ & $\underline{-.50}/\underline{-.50}$ & $\underline{.90}/\underline{.89}$ & $4.14/2.11$ & $0.23/0.33$ \\
    DiffMean & $.96/.94$ & $.93/.83$ & $-.76/-.76$ & $.89/\underline{.89}$ & $3.38/1.93$ & $\mathbf{0.00}/\mathbf{0.00}$ \\
    \hspace{5pt}\includegraphics[width=6pt]{figures_new/subdirectory_arrow_right.png} SAE & $.96/.93$ & $.90/.81$ & $-.77/-.77$ & $.88/.88$ & $\mathbf{0.28}/0.69$ & $0.13/0.05$ \\
    DiffMedian & $.96/.94$ & $.95/.84$ & $-.72/-.72$ & $\underline{.90}/.88$ & $3.45/3.38$ & $\mathbf{0.00}/\mathbf{0.00}$ \\
    \hspace{5pt}\includegraphics[width=6pt]{figures_new/subdirectory_arrow_right.png} SAE & $.93/.94$ & $.82/.80$ & $-.71/-.71$ & $.83/.86$ & $\underline{0.72}/2.42$ & $0.12/0.05$ \\
    FastCAV & $.96/.94$ & $.93/.83$ & $-.76/-.76$ & $.89/\underline{.89}$ & $3.38/1.93$ & $\mathbf{0.00}/\mathbf{0.00}$ \\
    \hspace{5pt}\includegraphics[width=6pt]{figures_new/subdirectory_arrow_right.png} SAE & $.96/.93$ & $.90/.81$ & $-.77/-.77$ & $.88/.88$ & $\mathbf{0.28}/0.69$ & $0.13/0.04$ \\
    PatCAV & $.96/.94$ & $.93/.83$ & $-.76/-.76$ & $.89/\underline{.89}$ & $3.38/1.93$ & $\mathbf{0.00}/\mathbf{0.00}$ \\
    AurA & $\underline{.97}/.95$ & $\underline{1.0}/\underline{1.0}$ & $\mathbf{+.04}/\mathbf{+.04}$ & $.89/\underline{.89}$ & $4.14/2.00$ & $0.81/0.75$ \\
    \hspace{5pt}\includegraphics[width=6pt]{figures_new/subdirectory_arrow_right.png} SAE & $.90/.89$ & $\mathbf{1.1}/\mathbf{1.1}$ & $+.69/+.69$ & $.81/.83$ & $2.45/\underline{0.45}$ & $25.9/24.7$ \\
    PCA & $.84/.75$ & $\underline{1.0}/.78$ & $-.75/-.75$ & $.73/.69$ & $3.49/3.97$ & $\underline{0.02}/0.05$ \\
    LAT & $.66/.66$ & $.95/.97$ & $-.65/-.65$ & $.59/.52$ & $0.79/\mathbf{0.03}$ & $0.06/\underline{0.03}$ \\
    \hline
    \end{tblr}
\end{table}

\begin{table}[h]
    \centering
    \footnotesize
    \caption{\textbf{DINOv2 representations steering evaluation on Waterbirds.} In addition to AurA, PCA also exhibits the non-negative values for MS. 
    }
    \label{tbl:results_waterbirds_dinov2}
    \vspace{0.5em}
    \begin{tblr}{
      colspec = {lcccccc},
      column{1} = {leftsep=0pt},
      vline{2} = {dashed},
      hline{7,16} = {dashed},
      row{4,8,10,12,15} = {bg=black!10},
      rowsep=1.5pt
    }
    \textbf{Method} & AUC ($\uparrow$) & CCR ($\uparrow$) & MS ($\downarrow$) & F1 ($\uparrow$) & CD ($\downarrow$) & Time (s) \\
    \hline
    Linear SVM & $\mathbf{.97}/\mathbf{.94}$ & $0.98/0.94$ & $-.33/-.33$ & $\mathbf{.91}/\underline{.83}$ & $2.55/1.00$ & $0.55/0.86$ \\
    LR & $\mathbf{.97}/\mathbf{.94}$ & $0.98/0.87$ & $-.39/-.39$ & $\underline{.90}/\mathbf{.84}$ & $4.00/0.90$ & $0.80/1.34$ \\
    \hspace{5pt}\includegraphics[width=6pt]{figures_new/subdirectory_arrow_right.png} SAE & $\underline{.92}/.83$ & $0.95/0.89$ & $-.28/-.28$ & $.82/.76$ & $2.49/0.66$ & $1.01/1.83$ \\
    SAS & $.63/.66$ & $1.00/1.00$ & $\mathbf{-.02}/\mathbf{-.02}$ & $.60/.59$ & $0.38/0.83$ & $1.09/2.20$ \\
    S\&PTopK & $.90/\underline{.88}$ & $0.87/0.87$ & $-.64/-.64$ & $.80/.75$ & $0.93/0.59$ & $0.36/0.38$ \\
    DiffMean & $\underline{.92}/.84$ & $0.93/0.88$ & $-.50/-.50$ & $.83/.80$ & $4.52/1.69$ & $\underline{0.01}/\mathbf{0.00}$ \\
    \hspace{5pt}\includegraphics[width=6pt]{figures_new/subdirectory_arrow_right.png} SAE & $.86/.76$ & $0.93/0.91$ & $-.38/-.38$ & $.77/.71$ & $4.59/0.90$ & $0.13/0.04$ \\
    DiffMedian & $.91/.84$ & $0.94/0.89$ & $-.44/-.44$ & $.83/.78$ & $4.38/1.52$ & $\mathbf{0.00}/\mathbf{0.00}$ \\
    \hspace{5pt}\includegraphics[width=6pt]{figures_new/subdirectory_arrow_right.png} SAE & $.52/.46$ & $1.00/1.00$ & $\mathbf{-.02}/\mathbf{-.02}$ & $.00/.00$ & $2.28/1.66$ & $0.16/0.06$ \\
    FastCAV & $\underline{.92}/.84$ & $0.93/0.88$ & $-.50/-.50$ & $.83/.80$ & $4.52/1.69$ & $0.04/0.04$ \\
    \hspace{5pt}\includegraphics[width=6pt]{figures_new/subdirectory_arrow_right.png} SAE & $.86/.76$ & $0.93/0.91$ & $-.38/-.38$ & $.77/.71$ & $4.59/0.90$ & $0.42/0.14$ \\
    PatCAV & $\underline{.92}/.84$ & $0.93/0.88$ & $-.50/-.50$ & $.83/.80$ & $4.52/1.69$ & $\mathbf{0.00}/\mathbf{0.00}$ \\
    AurA & $\underline{.92}/.87$ & $\underline{1.01}/\underline{1.01}$ & $\underline{+.11}/\underline{+.11}$ & $.83/.82$ & $3.24/1.14$ & $1.15/0.76$ \\
    \hspace{5pt}\includegraphics[width=6pt]{figures_new/subdirectory_arrow_right.png} SAE & $.71/.73$ & $\mathbf{1.17}/\mathbf{1.11}$ & $+.98/+.98$ & $.66/.54$ & $\underline{0.24}/\underline{0.10}$ & $26.1/24.8$ \\
    PCA & $.51/.53$ & $1.00/1.00$ & $\mathbf{+.02}/\mathbf{+.02}$ & $.49/.30$ & $\mathbf{0.07}/\underline{0.10}$ & $0.04/\underline{0.02}$ \\
    LAT & $.47/.54$ & $1.00/1.00$ & $\mathbf{-.02}/\mathbf{-.02}$ & $.24/.24$ & $0.28/\mathbf{0.07}$ & $0.03/0.04$ \\
    \hline
    \end{tblr}
\end{table}

\begin{table}[h]
    \centering
    \footnotesize
    \caption{\textbf{SigLIP representations steering evaluation on Waterbirds.} 
    }
    \label{tbl:results_waterbirds_siglip}
    \vspace{0.5em}
    \begin{tblr}{
      colspec = {lcccccc},
      column{1} = {leftsep=0pt},
      vline{2} = {dashed},
      hline{7,16} = {dashed},
      row{4,8,10,12,15} = {bg=black!10},
      rowsep=1.5pt
    }
    \textbf{Method} & AUC ($\uparrow$) & CCR ($\uparrow$) & MS ($\downarrow$) & F1 ($\uparrow$) & CD ($\downarrow$) & Time (s) \\
    \hline
    Linear SVM & $\mathbf{.99}/\underline{.98}$ & $0.95/0.80$ & $-.80/-.80$ & $\mathbf{.94}/\underline{.92}$ & $11.6/2.52$ & $0.78/1.11$ \\
    LR & $\mathbf{.99}/\underline{.98}$ & $0.95/0.79$ & $-.80/-.80$ & $\mathbf{.94}/\underline{.92}$ & $11.6/2.49$ & $0.61/0.89$ \\
    \hspace{5pt}\includegraphics[width=6pt]{figures_new/subdirectory_arrow_right.png} SAE & $\mathbf{.99}/\mathbf{.99}$ & $0.94/0.79$ & $-.64/-.64$ & $.90/.91$ & $11.7/2.35$ & $1.04/1.79$ \\
    SAS & $.96/\mathbf{.99}$ & $0.78/0.84$ & $-.70/-.70$ & $.87/.90$ & $8.46/\mathbf{0.14}$ & $1.23/1.98$ \\
    S\&PTopK & $\mathbf{.99}/\mathbf{.99}$ & $0.97/0.86$ & $-.54/-.68$ & $\mathbf{.94}/\mathbf{.94}$ & $8.32/1.73$ & $1.08/0.35$ \\
    DiffMean & $\underline{.97}/.97$ & $0.92/0.75$ & $-.82/-.82$ & $\underline{.91}/.89$ & $6.73/2.42$ & $0.14/0.07$ \\
    \hspace{5pt}\includegraphics[width=6pt]{figures_new/subdirectory_arrow_right.png} SAE & $\underline{.97}/.97$ & $0.89/0.73$ & $-.84/-.84$ & $.89/.88$ & $6.21/1.42$ & $0.13/0.05$ \\
    DiffMedian & $\underline{.97}/.97$ & $0.95/0.77$ & $-.76/-.76$ & $\underline{.91}/.90$ & $7.35/2.04$ & $\underline{0.01}/0.02$ \\
    \hspace{5pt}\includegraphics[width=6pt]{figures_new/subdirectory_arrow_right.png} SAE & $.95/.97$ & $0.82/0.86$ & $-.66/-.66$ & $.86/.90$ & $10.6/2.55$ & $0.15/0.05$ \\
    FastCAV & $\underline{.97}/.97$ & $0.92/0.75$ & $-.82/-.82$ & $\underline{.91}/.89$ & $6.73/2.42$ & $0.06/0.05$ \\
    \hspace{5pt}\includegraphics[width=6pt]{figures_new/subdirectory_arrow_right.png} SAE & $\underline{.97}/.97$ & $0.89/0.73$ & $-.84/-.84$ & $.89/.88$ & $6.21/1.42$ & $0.29/0.05$ \\
    PatCAV & $\underline{.97}/.97$ & $0.92/0.75$ & $-.82/-.82$ & $\underline{.91}/.89$ & $6.73/2.42$ & $\mathbf{0.00}/\mathbf{0.00}$ \\
    AurA & $\underline{.97}/.96$ & $1.00/\underline{1.00}$ & $\mathbf{+.04}/\mathbf{+.04}$ & $.90/.90$ & $\mathbf{1.10}/\underline{0.21}$ & $1.40/0.78$ \\
    \hspace{5pt}\includegraphics[width=6pt]{figures_new/subdirectory_arrow_right.png} SAE & $.93/.96$ & $\mathbf{1.04}/\mathbf{1.01}$ & $+.56/+.56$ & $.84/.84$ & $\underline{5.25}/1.31$ & $26.0/24.2$ \\
    PCA & $.77/.48$ & $\underline{1.02}/0.92$ & $\underline{-.18}/\underline{-.18}$ & $.71/.20$ & $13.3/0.24$ & $0.07/\underline{0.01}$ \\
    LAT & $.57/.53$ & $\underline{1.02}/\mathbf{1.01}$ & $+.19/+.19$ & $.56/.35$ & $5.49/\underline{0.21}$ & $0.02/0.02$ \\
    \hline
    \end{tblr}
\end{table}

\begin{table}[h]
    \centering
    \small
    \caption{\textbf{Evaluating results of steering on CLIP on ImageNet-W husky-cat watermark class pair in three data setups.}
    The results are presented for three CAV training setups: Vision class paired, where samples were drawn from \textit{Cat} class samples; All Classes Paired, from random samples from all 10 classes; and All Classes Unpaired, where we broke counterfactuality by applying a watermark to a different set of images. Columns report Downstream Concept Detection (F1) and Steering Disparity (SD). Values are represented with the mean $\pm$ two standard errors across 5 seeds, where the best method (row) in each evaluation metric (column) is in \textbf{bold} and the second best is \underline{underlined}. 
    We don't display CD as only PCA, DiffMean, FastCAV, DiffMedian (with their SAE version) reported deviation from 0 with LAT on the unpaired one.}
    \label{tbl:results_imagenetw_husky_clip}
    \vspace{0.5em}
    \begin{tblr}{
      colspec = {lccccccccc},
      column{1} = {leftsep=0pt},
      vline{2,5,8} = {dashed},
      hline{8,17} = {dashed},
      row{5,9,11,13,16} = {bg=black!10},
      rowsep=1.5pt,
      colsep = 2.6pt
    }
    & \SetCell[c=3]{c} \textbf{Vision Class Paired} & & & \SetCell[c=3]{c} \textbf{All Classes Paired} & & & \SetCell[c=3]{c} \textbf{All Classes Unpaired} & & \\
    \textbf{Method} & AUC ($\uparrow$) & F1 ($\uparrow$) & SD ($\downarrow$) & AUC ($\uparrow$) & F1 ($\uparrow$) & SD ($\downarrow$) & AUC ($\uparrow$) & F1 ($\uparrow$) & SD ($\downarrow$) \\
    \hline
    Linear SVM & $\mathbf{1.0}_{\pm.00}$ & $.91_{\pm.03}$ & $.57_{\pm.36}$ & $\mathbf{.99}_{\pm.00}$ & $.91_{\pm.04}$ & $.83_{\pm.12}$ & $\mathbf{.99}_{\pm.00}$ & $\underline{.94}_{\pm.02}$ & $.63_{\pm.16}$ \\
    LR & $\mathbf{1.0}_{\pm.00}$ & $.92_{\pm.02}$ & $.58_{\pm.42}$ & $\mathbf{.99}_{\pm.00}$ & $\underline{.94}_{\pm.01}$ & $.67_{\pm.28}$ & $\mathbf{.99}_{\pm.01}$ & $.91_{\pm.02}$ & $.44_{\pm.21}$ \\
    \hspace{5pt}\includegraphics[width=6pt]{figures_new/subdirectory_arrow_right.png} SAE & $\underline{.99}_{\pm.01}$ & $.82_{\pm.03}$ & $.71_{\pm.00}$ & $\underline{.98}_{\pm.02}$ & $.78_{\pm.10}$ & $.69_{\pm.02}$ & $\underline{.98}_{\pm.01}$ & $.77_{\pm.11}$ & $.74_{\pm.08}$ \\
    SAS & $\underline{.99}_{\pm.01}$ & $.90_{\pm.06}$ & $\mathbf{.05}_{\pm.00}$ & $\mathbf{.99}_{\pm.00}$ & $.91_{\pm.03}$ & $.23_{\pm.08}$ & $\mathbf{.99}_{\pm.01}$ & $.92_{\pm.01}$ & $\underline{.16}_{\pm.06}$ \\
    S\&PTopK & $\underline{.99}_{\pm.01}$ & $\underline{.95}_{\pm.02}$ & $.55_{\pm.20}$ & $\mathbf{.99}_{\pm.00}$ & $.92_{\pm.04}$ & $.75_{\pm.06}$ & $\mathbf{.99}_{\pm.00}$ & $\underline{.94}_{\pm.04}$ & $.73_{\pm.02}$ \\
    DiffMean & $\underline{.99}_{\pm.00}$ & $.84_{\pm.01}$ & $.41_{\pm.06}$ & $\underline{.98}_{\pm.00}$ & $.88_{\pm.02}$ & $.40_{\pm.09}$ & $\underline{.98}_{\pm.01}$ & $.90_{\pm.02}$ & $.36_{\pm.13}$ \\
    \hspace{5pt}\includegraphics[width=6pt]{figures_new/subdirectory_arrow_right.png} SAE & $\underline{.99}_{\pm.00}$ & $.81_{\pm.01}$ & $1.1_{\pm.04}$ & $\underline{.98}_{\pm.00}$ & $.86_{\pm.02}$ & $.95_{\pm.00}$ & $\underline{.98}_{\pm.01}$ & $.87_{\pm.02}$ & $1.0_{\pm.09}$ \\
    DiffMedian & $\underline{.99}_{\pm.00}$ & $.85_{\pm.01}$ & $.41_{\pm.07}$ & $\underline{.98}_{\pm.00}$ & $.89_{\pm.02}$ & $.30_{\pm.12}$ & $\underline{.98}_{\pm.00}$ & $.87_{\pm.01}$ & $.36_{\pm.13}$ \\
    \hspace{5pt}\includegraphics[width=6pt]{figures_new/subdirectory_arrow_right.png} SAE & $\underline{.99}_{\pm.00}$ & $.72_{\pm.01}$ & $1.5_{\pm.08}$ & $.97_{\pm.00}$ & $.80_{\pm.03}$ & $1.3_{\pm.07}$ & $.97_{\pm.01}$ & $.78_{\pm.01}$ & $1.4_{\pm.09}$ \\
    FastCAV & $\underline{.99}_{\pm.00}$ & $.84_{\pm.01}$ & $.41_{\pm.06}$ & $\underline{.98}_{\pm.00}$ & $.88_{\pm.02}$ & $.40_{\pm.09}$ & $\underline{.98}_{\pm.01}$ & $.90_{\pm.02}$ & $.36_{\pm.13}$ \\
    \hspace{5pt}\includegraphics[width=6pt]{figures_new/subdirectory_arrow_right.png} SAE & $\underline{.99}_{\pm.00}$ & $.81_{\pm.01}$ & $1.1_{\pm.04}$ & $\underline{.98}_{\pm.00}$ & $.86_{\pm.02}$ & $.95_{\pm.00}$ & $\underline{.98}_{\pm.01}$ & $.87_{\pm.02}$ & $1.0_{\pm.09}$ \\
    PatCAV & $\underline{.99}_{\pm.00}$ & $.84_{\pm.01}$ & $.41_{\pm.06}$ & $\underline{.98}_{\pm.00}$ & $.88_{\pm.02}$ & $.40_{\pm.09}$ & $\underline{.98}_{\pm.01}$ & $.90_{\pm.02}$ & $.36_{\pm.13}$ \\
    AurA & $\underline{.99}_{\pm.00}$ & $.84_{\pm.01}$ & $.24_{\pm.00}$ & $\underline{.98}_{\pm.00}$ & $.92_{\pm.02}$ & $.23_{\pm.02}$ & $\underline{.98}_{\pm.00}$ & $.91_{\pm.04}$ & $.23_{\pm.02}$ \\
    \hspace{5pt}\includegraphics[width=6pt]{figures_new/subdirectory_arrow_right.png} SAE & $\underline{.99}_{\pm.01}$ & $\mathbf{.97}_{\pm.00}$ & $.30_{\pm.02}$ & $\mathbf{.99}_{\pm.00}$ & $\mathbf{.96}_{\pm.01}$ & $.33_{\pm.03}$ & $\underline{.98}_{\pm.00}$ & $\mathbf{.96}_{\pm.02}$ & $.51_{\pm.06}$ \\
    PCA & $.86_{\pm.10}$ & $.49_{\pm.19}$ & $1.1_{\pm.22}$ & $.86_{\pm.04}$ & $.69_{\pm.02}$ & $\underline{.18}_{\pm.25}$ & $.91_{\pm.03}$ & $.79_{\pm.07}$ & $\mathbf{.05}_{\pm.04}$ \\
    LAT & $.73_{\pm.03}$ & $.42_{\pm.03}$ & $\underline{.06}_{\pm.02}$ & $.83_{\pm.01}$ & $.50_{\pm.01}$ & $\mathbf{.00}_{\pm.00}$ & $.55_{\pm.02}$ & $.34_{\pm.06}$ & $1.3_{\pm.39}$ \\
    \hline
    \end{tblr}
\end{table}

\begin{table}[h]
    \centering
    \small
    \caption{\textbf{Evaluating results of steering on SigLIP on ImageNet-W husky-cat watermark class pair in three data setups.} Non-zero CD was observed in SAS, SAE-FastCAV, and all classes in the PCA setup, with unpaired only for LAT.}
    \label{tbl:results_imagenetw_husky_siglip}
    \vspace{0.5em}
    \begin{tblr}{
      colspec = {lccccccccc},
      column{1} = {leftsep=0pt},
      vline{2,5,8} = {dashed},
      hline{8,17} = {dashed},
      row{5,9,11,13,16} = {bg=black!10},
      rowsep=1.5pt,
      colsep = 2.6pt
    }
    & \SetCell[c=3]{c} \textbf{Vision Class Paired} & & & \SetCell[c=3]{c} \textbf{All Classes Paired} & & & \SetCell[c=3]{c} \textbf{All Classes Unpaired} & & \\
    \textbf{Method} & AUC ($\uparrow$) & F1 ($\uparrow$) & SD ($\downarrow$) & AUC ($\uparrow$) & F1 ($\uparrow$) & SD ($\downarrow$) & AUC ($\uparrow$) & F1 ($\uparrow$) & SD ($\downarrow$) \\
    \hline
    Linear SVM & $\mathbf{1.0}_{\pm.00}$ & $\mathbf{.92}_{\pm.01}$ & $.16_{\pm.10}$ & $\mathbf{.99}_{\pm.00}$ & $.89_{\pm.05}$ & $.39_{\pm.15}$ & $\mathbf{.99}_{\pm.00}$ & $\underline{.94}_{\pm.02}$ & $.29_{\pm.06}$ \\
    LR & $\mathbf{1.0}_{\pm.00}$ & $\underline{.90}_{\pm.01}$ & $\underline{.09}_{\pm.15}$ & $\mathbf{.99}_{\pm.00}$ & $.88_{\pm.07}$ & $.49_{\pm.27}$ & $\mathbf{.99}_{\pm.00}$ & $.93_{\pm.02}$ & $.24_{\pm.08}$ \\
    \hspace{5pt}\includegraphics[width=6pt]{figures_new/subdirectory_arrow_right.png} SAE & $\underline{.99}_{\pm.00}$ & $.87_{\pm.04}$ & $.39_{\pm.19}$ & $\underline{.95}_{\pm.04}$ & $.71_{\pm.22}$ & $.34_{\pm.09}$ & $\underline{.99}_{\pm.00}$ & $.85_{\pm.05}$ & $.34_{\pm.05}$ \\
    SAS & $\underline{.99}_{\pm.01}$ & $.85_{\pm.05}$ & $.16_{\pm.25}$ & $\underline{.98}_{\pm.01}$ & $.85_{\pm.16}$ & $.32_{\pm.28}$ & $\mathbf{.99}_{\pm.00}$ & $.90_{\pm.05}$ & $.24_{\pm.20}$ \\
    S\&PTopK & $\underline{.99}_{\pm.00}$ & $.89_{\pm.02}$ & $.29_{\pm.06}$ & $\mathbf{.99}_{\pm.00}$ & $\underline{.90}_{\pm.02}$ & $.65_{\pm.20}$ & $\mathbf{.99}_{\pm.00}$ & $.90_{\pm.01}$ & $.49_{\pm.14}$ \\
    DiffMean & $\underline{.99}_{\pm.00}$ & $.78_{\pm.01}$ & $.11_{\pm.00}$ & $\underline{.98}_{\pm.00}$ & $.88_{\pm.01}$ & $.11_{\pm.00}$ & $\underline{.98}_{\pm.00}$ & $.89_{\pm.02}$ & $\underline{.11}_{\pm.00}$ \\
    \hspace{5pt}\includegraphics[width=6pt]{figures_new/subdirectory_arrow_right.png} SAE & $\underline{.99}_{\pm.00}$ & $.68_{\pm.01}$ & $.37_{\pm.08}$ & $\mathbf{.99}_{\pm.00}$ & $.86_{\pm.01}$ & $.24_{\pm.00}$ & $\underline{.98}_{\pm.00}$ & $.86_{\pm.02}$ & $.32_{\pm.15}$ \\
    DiffMedian & $\underline{.99}_{\pm.00}$ & $.75_{\pm.01}$ & $.11_{\pm.00}$ & $\underline{.98}_{\pm.00}$ & $.87_{\pm.02}$ & $\underline{.09}_{\pm.05}$ & $\underline{.98}_{\pm.00}$ & $.89_{\pm.04}$ & $\mathbf{.09}_{\pm.05}$ \\
    \hspace{5pt}\includegraphics[width=6pt]{figures_new/subdirectory_arrow_right.png} SAE & $\underline{.99}_{\pm.00}$ & $.66_{\pm.01}$ & $.32_{\pm.06}$ & $\mathbf{.99}_{\pm.00}$ & $.82_{\pm.01}$ & $.72_{\pm.19}$ & $\mathbf{.99}_{\pm.00}$ & $.85_{\pm.02}$ & $.59_{\pm.19}$ \\
    FastCAV & $\underline{.99}_{\pm.00}$ & $.78_{\pm.01}$ & $.11_{\pm.00}$ & $\underline{.98}_{\pm.00}$ & $.88_{\pm.01}$ & $.11_{\pm.00}$ & $\underline{.98}_{\pm.00}$ & $.89_{\pm.02}$ & $\underline{.11}_{\pm.00}$ \\
    \hspace{5pt}\includegraphics[width=6pt]{figures_new/subdirectory_arrow_right.png} SAE & $\underline{.99}_{\pm.00}$ & $.68_{\pm.01}$ & $.37_{\pm.08}$ & $\mathbf{.99}_{\pm.00}$ & $.86_{\pm.01}$ & $.24_{\pm.00}$ & $\underline{.98}_{\pm.00}$ & $.86_{\pm.02}$ & $.32_{\pm.15}$ \\
    PatCAV & $\underline{.99}_{\pm.00}$ & $.78_{\pm.01}$ & $.11_{\pm.00}$ & $\underline{.98}_{\pm.00}$ & $.88_{\pm.01}$ & $.11_{\pm.00}$ & $\underline{.98}_{\pm.00}$ & $.89_{\pm.02}$ & $\underline{.11}_{\pm.00}$ \\
    AurA & $\underline{.99}_{\pm.00}$ & $\underline{.90}_{\pm.00}$ & $.11_{\pm.00}$ & $\mathbf{.99}_{\pm.00}$ & $\mathbf{.95}_{\pm.00}$ & $.11_{\pm.00}$ & $\underline{.98}_{\pm.00}$ & $\mathbf{.95}_{\pm.01}$ & $\underline{.11}_{\pm.00}$ \\
    \hspace{5pt}\includegraphics[width=6pt]{figures_new/subdirectory_arrow_right.png} SAE & $.99_{\pm.00}$ & $.84_{\pm.01}$ & $.24_{\pm.00}$ & $\underline{.98}_{\pm.00}$ & $.89_{\pm.01}$ & $.24_{\pm.00}$ & $\underline{.98}_{\pm.00}$ & $.90_{\pm.02}$ & $.24_{\pm.00}$ \\
    PCA & $.66_{\pm.02}$ & $.53_{\pm.02}$ & $.87_{\pm.00}$ & $.60_{\pm.02}$ & $.46_{\pm.04}$ & $3.6_{\pm1.0}$ & $.59_{\pm.02}$ & $.41_{\pm.04}$ & $3.5_{\pm.67}$ \\
    LAT & $.91_{\pm.01}$ & $.38_{\pm.01}$ & $\mathbf{.01}_{\pm.05}$ & $.81_{\pm.01}$ & $.51_{\pm.01}$ & $\mathbf{.04}_{\pm.06}$ & $.54_{\pm.02}$ & $.35_{\pm.05}$ & $3.1_{\pm1.1}$ \\
    \hline
    \end{tblr}
\end{table}

\begin{table}[h]
    \centering
    \small
    \caption{
    \textbf{Evaluating results of steering on CLIP on ImageNet-W church-goldfish watermark class pair in three data setups.} Non-zero CD was observed in PCA, DiffMean, PatCAV, FastCAV, and DiffMedian (including their SAE variants). Additionally, for the LAT in the unpaired version.
    }
    \label{tbl:results_imagenetw_church_clip}
    \vspace{0.5em}
    \begin{tblr}{
      colspec = {lccccccccc},
      column{1} = {leftsep=0pt},
      vline{2,5,8} = {dashed},
      hline{8,17} = {dashed},
      row{5,9,11,13,16} = {bg=black!10},
      rowsep=1.5pt,
      colsep = 2.6pt
    }
    & \SetCell[c=3]{c} \textbf{Vision Class Paired} & & & \SetCell[c=3]{c} \textbf{All Classes Paired} & & & \SetCell[c=3]{c} \textbf{All Classes Unpaired} & & \\
    \textbf{Method} & AUC ($\uparrow$) & F1 ($\uparrow$) & SD ($\downarrow$) & AUC ($\uparrow$) & F1 ($\uparrow$) & SD ($\downarrow$) & AUC ($\uparrow$) & F1 ($\uparrow$) & SD ($\downarrow$) \\
    \hline
    Linear SVM & $\mathbf{1.0}_{\pm.00}$ & $\underline{.97}_{\pm.02}$ & $\underline{.08}_{\pm.00}$ & $\mathbf{1.0}_{\pm.00}$ & $\mathbf{.98}_{\pm.02}$ & $.43_{\pm.18}$ & $\mathbf{1.0}_{\pm.00}$ & $\mathbf{.98}_{\pm.01}$ & $.35_{\pm.08}$ \\
    LR & $\mathbf{1.0}_{\pm.00}$ & $.95_{\pm.01}$ & $\mathbf{.00}_{\pm.00}$ & $\mathbf{1.0}_{\pm.00}$ & $\underline{.97}_{\pm.01}$ & $.32_{\pm.23}$ & $\mathbf{1.0}_{\pm.00}$ & $\underline{.97}_{\pm.01}$ & $.23_{\pm.14}$ \\
    \hspace{5pt}\includegraphics[width=6pt]{figures_new/subdirectory_arrow_right.png} SAE & $\mathbf{1.0}_{\pm.00}$ & $.75_{\pm.09}$ & $.15_{\pm.06}$ & $.99_{\pm.00}$ & $.81_{\pm.11}$ & $.53_{\pm.04}$ & $.99_{\pm.00}$ & $.92_{\pm.03}$ & $.53_{\pm.11}$ \\
    SAS & $\mathbf{1.0}_{\pm.00}$ & $.70_{\pm.06}$ & $\mathbf{.00}_{\pm.00}$ & $.97_{\pm.01}$ & $.89_{\pm.06}$ & $\underline{.07}_{\pm.06}$ & $.97_{\pm.01}$ & $.88_{\pm.08}$ & $\mathbf{.05}_{\pm.07}$ \\
    S\&PTopK & $\mathbf{1.0}_{\pm.00}$ & $.87_{\pm.07}$ & $.37_{\pm.12}$ & $\underline{.99}_{\pm.00}$ & $.96_{\pm.01}$ & $.60_{\pm.16}$ & $\underline{.99}_{\pm.00}$ & $.96_{\pm.01}$ & $.52_{\pm.03}$ \\
    DiffMean & $\mathbf{1.0}_{\pm.00}$ & $.92_{\pm.01}$ & $\underline{.08}_{\pm.00}$ & $\underline{.99}_{\pm.00}$ & $.94_{\pm.00}$ & $.17_{\pm.00}$ & $\underline{.99}_{\pm.00}$ & $.95_{\pm.00}$ & $.15_{\pm.03}$ \\
    \hspace{5pt}\includegraphics[width=6pt]{figures_new/subdirectory_arrow_right.png} SAE & $\mathbf{1.0}_{\pm.00}$ & $.66_{\pm.01}$ & $.23_{\pm.03}$ & $\underline{.99}_{\pm.00}$ & $.93_{\pm.00}$ & $.38_{\pm.07}$ & $\underline{.99}_{\pm.00}$ & $.93_{\pm.01}$ & $.35_{\pm.10}$ \\
    DiffMedian & $\mathbf{1.0}_{\pm.00}$ & $.91_{\pm.00}$ & $\underline{.08}_{\pm.00}$ & $\underline{.99}_{\pm.00}$ & $.95_{\pm.00}$ & $.17_{\pm.00}$ & $\underline{.99}_{\pm.00}$ & $.95_{\pm.00}$ & $.15_{\pm.03}$ \\
    \hspace{5pt}\includegraphics[width=6pt]{figures_new/subdirectory_arrow_right.png} SAE & $\mathbf{1.0}_{\pm.00}$ & $.60_{\pm.01}$ & $1.9_{\pm.26}$ & $.97_{\pm.00}$ & $.88_{\pm.02}$ & $.93_{\pm.13}$ & $.97_{\pm.01}$ & $.88_{\pm.03}$ & $.93_{\pm.34}$ \\
    FastCAV & $\mathbf{1.0}_{\pm.00}$ & $.92_{\pm.01}$ & $\underline{.08}_{\pm.00}$ & $\underline{.99}_{\pm.00}$ & $.94_{\pm.00}$ & $.17_{\pm.00}$ & $\underline{.99}_{\pm.00}$ & $.95_{\pm.00}$ & $.15_{\pm.03}$ \\
    \hspace{5pt}\includegraphics[width=6pt]{figures_new/subdirectory_arrow_right.png} SAE & $\mathbf{1.0}_{\pm.00}$ & $.66_{\pm.01}$ & $.23_{\pm.03}$ & $\underline{.99}_{\pm.00}$ & $.93_{\pm.00}$ & $.38_{\pm.07}$ & $\underline{.99}_{\pm.00}$ & $.93_{\pm.01}$ & $.35_{\pm.10}$ \\
    PatCAV & $\mathbf{1.0}_{\pm.00}$ & $.92_{\pm.01}$ & $\underline{.08}_{\pm.00}$ & $\underline{.99}_{\pm.00}$ & $.94_{\pm.00}$ & $.17_{\pm.00}$ & $\underline{.99}_{\pm.00}$ & $.95_{\pm.00}$ & $.15_{\pm.03}$ \\
    AurA & $\mathbf{1.0}_{\pm.00}$ & $.96_{\pm.00}$ & $\underline{.08}_{\pm.00}$ & $\underline{.99}_{\pm.00}$ & $.96_{\pm.01}$ & $.08_{\pm.00}$ & $\underline{.99}_{\pm.00}$ & $.96_{\pm.00}$ & $\underline{.08}_{\pm.00}$ \\
    \hspace{5pt}\includegraphics[width=6pt]{figures_new/subdirectory_arrow_right.png} SAE & $\mathbf{1.0}_{\pm.00}$ & $\mathbf{.99}_{\pm.00}$ & $.50_{\pm.00}$ & $\underline{.99}_{\pm.00}$ & $.96_{\pm.01}$ & $.50_{\pm.00}$ & $\underline{.99}_{\pm.00}$ & $\underline{.97}_{\pm.01}$ & $.50_{\pm.00}$ \\
    PCA & $\mathbf{1.0}_{\pm.00}$ & $.81_{\pm.01}$ & $\mathbf{.00}_{\pm.00}$ & $.98_{\pm.00}$ & $.94_{\pm.01}$ & $.13_{\pm.04}$ & $.98_{\pm.00}$ & $.87_{\pm.10}$ & $.17_{\pm.00}$ \\
    LAT & $.55_{\pm.06}$ & $.17_{\pm.08}$ & $.20_{\pm.07}$ & $.86_{\pm.00}$ & $.62_{\pm.01}$ & $\mathbf{.00}_{\pm.00}$ & $.53_{\pm.01}$ & $.23_{\pm.10}$ & $3.6_{\pm.56}$ \\
    \hline
    \end{tblr}
\end{table}

\begin{table}[h]
    \centering
    \small
    \caption{
    \textbf{Evaluating results of steering on SigLIP on ImageNet-W church-goldfish watermark class pair in three data setups.} Non-zero CD was observed across three settings. For Vision Class Paired, this occurred with SAE-FastCAV, SAE-DiffMean, and SAE-DiffMedian. For All Classes Paired, non-zero CD was observed in PCA, as well as the DiffMean, PatCAV, FastCAV, and DiffMedian methods (including their SAE variants). Finally, for All Classes Unpaired, non-zero CD was seen with PCA, LAT, and the aforementioned methods (including their SAE variants).
    }
    \label{tbl:results_imagenetw_church_siglip}
    \vspace{0.5em}
    \begin{tblr}{
      colspec = {lccccccccc},
      column{1} = {leftsep=0pt},
      vline{2,5,8} = {dashed},
      hline{8,17} = {dashed},
      row{5,9,11,13,16} = {bg=black!10},
      rowsep=1.5pt,
      colsep = 2.6pt
    }
    & \SetCell[c=3]{c} \textbf{Vision Class Paired} & & & \SetCell[c=3]{c} \textbf{All Classes Paired} & & & \SetCell[c=3]{c} \textbf{All Classes Unpaired} & & \\
    \textbf{Method} & AUC ($\uparrow$) & F1 ($\uparrow$) & SD ($\downarrow$) & AUC ($\uparrow$) & F1 ($\uparrow$) & SD ($\downarrow$) & AUC ($\uparrow$) & F1 ($\uparrow$) & SD ($\downarrow$) \\
    \hline
    Linear SVM & $\mathbf{1.0}_{\pm.00}$ & $\underline{.93}_{\pm.03}$ & $.49_{\pm.19}$ & $\mathbf{1.0}_{\pm.00}$ & $\mathbf{.97}_{\pm.01}$ & $.69_{\pm.07}$ & $\mathbf{1.0}_{\pm.00}$ & $\mathbf{.98}_{\pm.02}$ & $.69_{\pm.07}$ \\
    LR & $\mathbf{1.0}_{\pm.00}$ & $.80_{\pm.13}$ & $.25_{\pm.20}$ & $\mathbf{1.0}_{\pm.00}$ & $\underline{.96}_{\pm.01}$ & $.66_{\pm.00}$ & $\mathbf{1.0}_{\pm.00}$ & $\mathbf{.98}_{\pm.02}$ & $.69_{\pm.07}$ \\
    \hspace{5pt}\includegraphics[width=6pt]{figures_new/subdirectory_arrow_right.png} SAE & $\mathbf{1.0}_{\pm.00}$ & $.71_{\pm.03}$ & $.32_{\pm.15}$ & $\underline{.99}_{\pm.00}$ & $.86_{\pm.07}$ & $.73_{\pm.08}$ & $\underline{.99}_{\pm.00}$ & $.87_{\pm.09}$ & $.66_{\pm.00}$ \\
    SAS & $\mathbf{1.0}_{\pm.00}$ & $.66_{\pm.06}$ & $\mathbf{.02}_{\pm.07}$ & $.98_{\pm.01}$ & $.83_{\pm.03}$ & $.49_{\pm.36}$ & $.98_{\pm.01}$ & $.76_{\pm.08}$ & $.56_{\pm.14}$ \\
    S\&PTopK & $\mathbf{1.0}_{\pm.00}$ & $.87_{\pm.07}$ & $.66_{\pm.00}$ & $\underline{.99}_{\pm.00}$ & $\underline{.96}_{\pm.01}$ & $.66_{\pm.00}$ & $\underline{.99}_{\pm.00}$ & $\underline{.96}_{\pm.01}$ & $.66_{\pm.00}$ \\
    DiffMean & $\mathbf{1.0}_{\pm.00}$ & $.55_{\pm.01}$ & $\underline{.15}_{\pm.00}$ & $.98_{\pm.00}$ & $.77_{\pm.04}$ & $\underline{.19}_{\pm.07}$ & $.98_{\pm.00}$ & $.71_{\pm.10}$ & $\underline{.22}_{\pm.08}$ \\
    \hspace{5pt}\includegraphics[width=6pt]{figures_new/subdirectory_arrow_right.png} SAE & $\mathbf{1.0}_{\pm.00}$ & $.41_{\pm.01}$ & $.53_{\pm.07}$ & $.97_{\pm.00}$ & $.63_{\pm.02}$ & $.56_{\pm.08}$ & $.97_{\pm.01}$ & $.66_{\pm.11}$ & $.63_{\pm.07}$ \\
    DiffMedian & $\mathbf{1.0}_{\pm.00}$ & $.54_{\pm.01}$ & $\underline{.15}_{\pm.00}$ & $.98_{\pm.00}$ & $.80_{\pm.01}$ & $\mathbf{.15}_{\pm.00}$ & $.97_{\pm.00}$ & $.71_{\pm.11}$ & $\mathbf{.15}_{\pm.00}$ \\
    \hspace{5pt}\includegraphics[width=6pt]{figures_new/subdirectory_arrow_right.png} SAE & $\underline{.99}_{\pm.00}$ & $.37_{\pm.03}$ & $.83_{\pm.11}$ & $.97_{\pm.00}$ & $.68_{\pm.03}$ & $.53_{\pm.07}$ & $.97_{\pm.01}$ & $.69_{\pm.03}$ & $.56_{\pm.08}$ \\
    FastCAV & $\mathbf{1.0}_{\pm.00}$ & $.55_{\pm.01}$ & $\underline{.15}_{\pm.00}$ & $.98_{\pm.00}$ & $.77_{\pm.04}$ & $\underline{.19}_{\pm.07}$ & $.98_{\pm.00}$ & $.71_{\pm.10}$ & $\underline{.22}_{\pm.08}$ \\
    \hspace{5pt}\includegraphics[width=6pt]{figures_new/subdirectory_arrow_right.png} SAE & $\mathbf{1.0}_{\pm.00}$ & $.41_{\pm.01}$ & $.53_{\pm.07}$ & $.97_{\pm.00}$ & $.63_{\pm.02}$ & $.56_{\pm.08}$ & $.97_{\pm.01}$ & $.66_{\pm.11}$ & $.63_{\pm.07}$ \\
    PatCAV & $\mathbf{1.0}_{\pm.00}$ & $.55_{\pm.01}$ & $\underline{.15}_{\pm.00}$ & $.98_{\pm.00}$ & $.77_{\pm.04}$ & $\underline{.19}_{\pm.07}$ & $.98_{\pm.00}$ & $.71_{\pm.10}$ & $\underline{.22}_{\pm.08}$ \\
    AurA & $\mathbf{1.0}_{\pm.00}$ & $.62_{\pm.01}$ & $.49_{\pm.00}$ & $.98_{\pm.00}$ & $.89_{\pm.02}$ & $.49_{\pm.00}$ & $.98_{\pm.00}$ & $.87_{\pm.06}$ & $.49_{\pm.00}$ \\
    \hspace{5pt}\includegraphics[width=6pt]{figures_new/subdirectory_arrow_right.png} SAE & $\mathbf{1.0}_{\pm.00}$ & $\mathbf{.98}_{\pm.00}$ & $.66_{\pm.00}$ & $\underline{.99}_{\pm.00}$ & $\underline{.96}_{\pm.00}$ & $.66_{\pm.00}$ & $\underline{.99}_{\pm.00}$ & $.95_{\pm.01}$ & $.66_{\pm.00}$ \\
    PCA & $.98_{\pm.00}$ & $.46_{\pm.00}$ & $\underline{.15}_{\pm.00}$ & $.55_{\pm.03}$ & $.20_{\pm.01}$ & $2.6_{\pm.71}$ & $.56_{\pm.05}$ & $.21_{\pm.05}$ & $3.6_{\pm1.9}$ \\
    LAT & $.75_{\pm.01}$ & $.35_{\pm.01}$ & $.42_{\pm.08}$ & $.74_{\pm.01}$ & $.54_{\pm.01}$ & $\mathbf{.15}_{\pm.00}$ & $.52_{\pm.02}$ & $.24_{\pm.08}$ & $2.2_{\pm.86}$ \\
    \hline
    \end{tblr}
\end{table}

\begin{table}[h]
    \centering
    \small
    \caption{
    \textbf{Evaluating steering representations on ImageNet-C \emph{Shot Noise} and \emph{Glass Blur} corruptions on CLIP.}
    The best method (row) in each evaluation metric (column) is in \textbf{bold} and the second best is \underline{underlined}.
    }
    \label{tbl:results_imagenet_c_clip}
    \vspace{0.5em}
    \begin{tblr}{
      colspec = {lcccccccccc},
      column{1} = {leftsep=0pt},
      vline{2,7} = {dashed},
      hline{8,17} = {dashed},
      row{5,9,11,13,16} = {bg=black!10},
      rowsep=1.5pt,
      colsep = 2.3pt
    }
    & \SetCell[c=5]{c} \textbf{Shot Noise} & & & & & \SetCell[c=5]{c} \textbf{Glass Blur} & & & & \\
    \textbf{Method} & AUC ($\uparrow$) & MS ($\downarrow$) & F1 ($\uparrow$) & CD ($\downarrow$) & SD ($\downarrow$) & AUC ($\uparrow$) & MS ($\downarrow$) & F1 ($\uparrow$) & CD ($\downarrow$) & SD ($\downarrow$) \\
    \hline
    Linear SVM & $\mathbf{1.0}_{\pm.00}$ & $\underline{.98}_{\pm.00}$ & $\mathbf{1.0}_{\pm.00}$ & $\mathbf{.00}_{\pm.00}$ & $.86_{\pm.03}$ & $\mathbf{1.0}_{\pm.00}$ & $.74_{\pm.01}$ & $\mathbf{1.0}_{\pm.00}$ & $\mathbf{.00}_{\pm.00}$ & $1.16_{\pm.05}$ \\
    LR & $\mathbf{1.0}_{\pm.00}$ & $1.0_{\pm.00}$ & $\mathbf{1.0}_{\pm.00}$ & $\mathbf{.00}_{\pm.00}$ & $.86_{\pm.02}$ & $\mathbf{1.0}_{\pm.00}$ & $.75_{\pm.00}$ & $\underline{.99}_{\pm.04}$ & $\mathbf{.00}_{\pm.00}$ & $\underline{1.13}_{\pm.03}$ \\
    \hspace{5pt}\includegraphics[width=6pt]{figures_new/subdirectory_arrow_right.png} SAE & $\mathbf{1.0}_{\pm.00}$ & $\mathbf{.96}_{\pm.04}$ & $\underline{.99}_{\pm.32}$ & $\mathbf{.00}_{\pm.00}$ & $1.1_{\pm.03}$ & $\mathbf{1.0}_{\pm.00}$ & $.74_{\pm.05}$ & $\underline{.99}_{\pm.10}$ & $\mathbf{.00}_{\pm.00}$ & $1.19_{\pm.01}$ \\
    SAS & $\mathbf{1.0}_{\pm.00}$ & $\underline{.98}_{\pm.03}$ & $\underline{.99}_{\pm.15}$ & $\mathbf{.00}_{\pm.00}$ & $.98_{\pm.16}$ & $\mathbf{1.0}_{\pm.00}$ & $.77_{\pm.03}$ & $\underline{.99}_{\pm.18}$ & $\mathbf{.00}_{\pm.00}$ & $1.28_{\pm.03}$ \\
    S\&PTopK & $\mathbf{1.0}_{\pm.00}$ & $\underline{.98}_{\pm.01}$ & $\underline{.99}_{\pm.23}$ & $\mathbf{.00}_{\pm.00}$ & $.91_{\pm.03}$ & $\mathbf{1.0}_{\pm.00}$ & $.81_{\pm.03}$ & $\underline{.99}_{\pm.05}$ & $\mathbf{.00}_{\pm.00}$ & $1.15_{\pm.03}$ \\
    DiffMean & $\mathbf{1.0}_{\pm.00}$ & $1.0_{\pm.00}$ & $\mathbf{1.0}_{\pm.00}$ & $\mathbf{.00}_{\pm.00}$ & $.82_{\pm.02}$ & $\mathbf{1.0}_{\pm.00}$ & $\underline{.73}_{\pm.00}$ & $\underline{.99}_{\pm.05}$ & $\mathbf{.00}_{\pm.00}$ & $1.15_{\pm.02}$ \\
    \hspace{5pt}\includegraphics[width=6pt]{figures_new/subdirectory_arrow_right.png} SAE & $\mathbf{1.0}_{\pm.00}$ & $1.0_{\pm.00}$ & $\underline{.99}_{\pm.00}$ & $\mathbf{.00}_{\pm.00}$ & $.83_{\pm.02}$ & $\mathbf{1.0}_{\pm.00}$ & $.83_{\pm.00}$ & $\underline{.99}_{\pm.05}$ & $\mathbf{.00}_{\pm.00}$ & $1.21_{\pm.03}$ \\
    DiffMedian & $\mathbf{1.0}_{\pm.00}$ & $.99_{\pm.00}$ & $\underline{.99}_{\pm.04}$ & $\mathbf{.00}_{\pm.00}$ & $.85_{\pm.03}$ & $\mathbf{1.0}_{\pm.00}$ & $\underline{.73}_{\pm.00}$ & $\underline{.99}_{\pm.04}$ & $\mathbf{.00}_{\pm.00}$ & $1.17_{\pm.01}$ \\
    \hspace{5pt}\includegraphics[width=6pt]{figures_new/subdirectory_arrow_right.png} SAE & $\mathbf{1.0}_{\pm.00}$ & $.99_{\pm.00}$ & $\underline{.99}_{\pm.05}$ & $\mathbf{.00}_{\pm.00}$ & $\underline{.79}_{\pm.01}$ & $\mathbf{1.0}_{\pm.00}$ & $.82_{\pm.01}$ & $\underline{.99}_{\pm.05}$ & $\mathbf{.00}_{\pm.00}$ & $1.15_{\pm.02}$ \\
    FastCAV & $\mathbf{1.0}_{\pm.00}$ & $1.0_{\pm.00}$ & $\mathbf{1.0}_{\pm.00}$ & $\mathbf{.00}_{\pm.00}$ & $.82_{\pm.02}$ & $\mathbf{1.0}_{\pm.00}$ & $\underline{.73}_{\pm.00}$ & $\underline{.99}_{\pm.05}$ & $\mathbf{.00}_{\pm.00}$ & $1.15_{\pm.02}$ \\
    \hspace{5pt}\includegraphics[width=6pt]{figures_new/subdirectory_arrow_right.png} SAE & $\mathbf{1.0}_{\pm.00}$ & $1.0_{\pm.00}$ & $\underline{.99}_{\pm.00}$ & $\mathbf{.00}_{\pm.00}$ & $.83_{\pm.02}$ & $\mathbf{1.0}_{\pm.00}$ & $.83_{\pm.00}$ & $\underline{.99}_{\pm.05}$ & $\mathbf{.00}_{\pm.00}$ & $1.21_{\pm.03}$ \\
    PatCAV & $\mathbf{1.0}_{\pm.00}$ & $1.0_{\pm.00}$ & $\mathbf{1.0}_{\pm.00}$ & $\mathbf{.00}_{\pm.00}$ & $.82_{\pm.02}$ & $\mathbf{1.0}_{\pm.00}$ & $\underline{.73}_{\pm.00}$ & $\underline{.99}_{\pm.05}$ & $\mathbf{.00}_{\pm.00}$ & $1.15_{\pm.02}$ \\
    AurA & $\mathbf{1.0}_{\pm.00}$ & $1.0_{\pm.00}$ & $\mathbf{1.0}_{\pm.00}$ & $\mathbf{.00}_{\pm.00}$ & $.94_{\pm.02}$ & $\mathbf{1.0}_{\pm.00}$ & $.76_{\pm.00}$ & $\underline{.99}_{\pm.04}$ & $\mathbf{.00}_{\pm.00}$ & $1.18_{\pm.01}$ \\
    \hspace{5pt}\includegraphics[width=6pt]{figures_new/subdirectory_arrow_right.png} SAE & $\mathbf{1.0}_{\pm.00}$ & $1.0_{\pm.00}$ & $\underline{.99}_{\pm.00}$ & $\mathbf{.00}_{\pm.00}$ & $.94_{\pm.00}$ & $\mathbf{1.0}_{\pm.00}$ & $.89_{\pm.00}$ & $\underline{.99}_{\pm.04}$ & $\mathbf{.00}_{\pm.00}$ & $1.15_{\pm.01}$ \\
    PCA & $\mathbf{1.0}_{\pm.00}$ & $1.0_{\pm.00}$ & $\underline{.99}_{\pm.04}$ & $\mathbf{.00}_{\pm.00}$ & $\mathbf{.70}_{\pm.03}$ & $\mathbf{1.0}_{\pm.00}$ & $\mathbf{.72}_{\pm.01}$ & $\underline{.99}_{\pm.07}$ & $\mathbf{.00}_{\pm.00}$ & $\mathbf{1.12}_{\pm.03}$ \\
    LAT & $.76_{\pm.04}$ & $.97_{\pm.02}$ & $.72_{\pm.21}$ & $\mathbf{.00}_{\pm.00}$ & $1.4_{\pm.21}$ & $.67_{\pm.07}$ & $.81_{\pm.04}$ & $.71_{\pm.82}$ & $.04_{\pm.08}$ & $2.67_{\pm.23}$ \\
    \hline
    \end{tblr}
\end{table}

\begin{table}[h]
    \centering
    \small
    \caption{
    \textbf{Evaluating steering representations on ImageNet-C \emph{Shot Noise} and \emph{Glass Blur} corruptions DINOv2.}
    The best method (row) in each evaluation metric (column) is in \textbf{bold} and the second best is \underline{underlined}.
    }
    \label{tbl:results_imagenet_c_dinov2}
    \vspace{0.5em}
    \begin{tblr}{
      colspec = {lcccccccccc},
      column{1} = {leftsep=0pt},
      vline{2,7} = {dashed},
      hline{8,17} = {dashed},
      row{5,9,11,13,16} = {bg=black!10},
      rowsep=1.5pt,
      colsep = 2.3pt
    }
    & \SetCell[c=5]{c} \textbf{Shot Noise} & & & & & \SetCell[c=5]{c} \textbf{Glass Blur} & & & & \\
    \textbf{Method} & AUC ($\uparrow$) & MS ($\downarrow$) & F1 ($\uparrow$) & CD ($\downarrow$) & SD ($\downarrow$) & AUC ($\uparrow$) & MS ($\downarrow$) & F1 ($\uparrow$) & CD ($\downarrow$) & SD ($\downarrow$) \\
    \hline
    Linear SVM & $\mathbf{1.0}_{\pm.00}$ & $.91_{\pm.03}$ & $\mathbf{1.0}_{\pm.00}$ & $\mathbf{.00}_{\pm.00}$ & $1.1_{\pm.02}$ & $\mathbf{1.0}_{\pm.00}$ & $.64_{\pm.03}$ & $\mathbf{1.0}_{\pm.00}$ & $\mathbf{.00}_{\pm.00}$ & $1.03_{\pm.03}$ \\
    LR & $\mathbf{1.0}_{\pm.00}$ & $.94_{\pm.02}$ & $\underline{.99}_{\pm.00}$ & $\mathbf{.00}_{\pm.00}$ & $1.1_{\pm.01}$ & $\mathbf{1.0}_{\pm.00}$ & $.66_{\pm.02}$ & $\underline{.99}_{\pm.00}$ & $\mathbf{.00}_{\pm.00}$ & $1.03_{\pm.02}$ \\
    \hspace{5pt}\includegraphics[width=6pt]{figures_new/subdirectory_arrow_right.png} SAE & $\mathbf{1.0}_{\pm.00}$ & $\mathbf{.72}_{\pm.03}$ & $.98_{\pm.01}$ & $\mathbf{.00}_{\pm.00}$ & $1.0_{\pm.07}$ & $\mathbf{1.0}_{\pm.00}$ & $\mathbf{.48}_{\pm.04}$ & $\underline{.99}_{\pm.00}$ & $\mathbf{.00}_{\pm.00}$ & $\underline{0.95}_{\pm.05}$ \\
    SAS & $.97_{\pm.00}$ & $.96_{\pm.02}$ & $.91_{\pm.01}$ & $\mathbf{.00}_{\pm.00}$ & $1.1_{\pm.01}$ & $.98_{\pm.00}$ & $.74_{\pm.04}$ & $.95_{\pm.00}$ & $\mathbf{.00}_{\pm.00}$ & $0.97_{\pm.00}$ \\
    S\&PTopK & $\underline{.99}_{\pm.00}$ & $.92_{\pm.02}$ & $.96_{\pm.01}$ & $\mathbf{.00}_{\pm.00}$ & $1.0_{\pm.01}$ & $\underline{.99}_{\pm.00}$ & $.72_{\pm.04}$ & $.95_{\pm.01}$ & $\mathbf{.00}_{\pm.00}$ & $\mathbf{0.94}_{\pm.01}$ \\
    DiffMean & $\mathbf{1.0}_{\pm.00}$ & $.96_{\pm.00}$ & $.97_{\pm.00}$ & $\mathbf{.00}_{\pm.00}$ & $\mathbf{.97}_{\pm.03}$ & $\mathbf{1.0}_{\pm.00}$ & $.63_{\pm.00}$ & $.98_{\pm.00}$ & $\mathbf{.00}_{\pm.00}$ & $1.03_{\pm.01}$ \\
    \hspace{5pt}\includegraphics[width=6pt]{figures_new/subdirectory_arrow_right.png} SAE & $\underline{.99}_{\pm.00}$ & $.95_{\pm.00}$ & $.95_{\pm.00}$ & $\mathbf{.00}_{\pm.00}$ & $\underline{.99}_{\pm.04}$ & $\mathbf{1.0}_{\pm.00}$ & $.76_{\pm.01}$ & $.97_{\pm.00}$ & $\mathbf{.00}_{\pm.00}$ & $1.09_{\pm.02}$ \\
    DiffMedian & $\mathbf{1.0}_{\pm.00}$ & $.95_{\pm.00}$ & $.97_{\pm.00}$ & $\mathbf{.00}_{\pm.00}$ & $1.0_{\pm.05}$ & $\mathbf{1.0}_{\pm.00}$ & $.64_{\pm.00}$ & $.98_{\pm.00}$ & $\mathbf{.00}_{\pm.00}$ & $1.09_{\pm.02}$ \\
    \hspace{5pt}\includegraphics[width=6pt]{figures_new/subdirectory_arrow_right.png} SAE & $.94_{\pm.00}$ & $1.0_{\pm.00}$ & $.87_{\pm.00}$ & $\mathbf{.00}_{\pm.00}$ & $1.2_{\pm.00}$ & $.96_{\pm.00}$ & $1.0_{\pm.00}$ & $.91_{\pm.00}$ & $\mathbf{.00}_{\pm.00}$ & $\underline{0.95}_{\pm.00}$ \\
    FastCAV & $\mathbf{1.0}_{\pm.00}$ & $.96_{\pm.00}$ & $.97_{\pm.00}$ & $\mathbf{.00}_{\pm.00}$ & $\mathbf{.97}_{\pm.03}$ & $\mathbf{1.0}_{\pm.00}$ & $.63_{\pm.00}$ & $.98_{\pm.00}$ & $\mathbf{.00}_{\pm.00}$ & $1.03_{\pm.01}$ \\
    \hspace{5pt}\includegraphics[width=6pt]{figures_new/subdirectory_arrow_right.png} SAE & $\underline{.99}_{\pm.00}$ & $.95_{\pm.00}$ & $.95_{\pm.00}$ & $\mathbf{.00}_{\pm.00}$ & $\underline{.99}_{\pm.04}$ & $\mathbf{1.0}_{\pm.00}$ & $.76_{\pm.01}$ & $.97_{\pm.00}$ & $\mathbf{.00}_{\pm.00}$ & $1.09_{\pm.02}$ \\
    PatCAV & $\mathbf{1.0}_{\pm.00}$ & $.96_{\pm.00}$ & $.97_{\pm.00}$ & $\mathbf{.00}_{\pm.00}$ & $\mathbf{.97}_{\pm.03}$ & $\mathbf{1.0}_{\pm.00}$ & $.63_{\pm.00}$ & $.98_{\pm.00}$ & $\mathbf{.00}_{\pm.00}$ & $1.03_{\pm.01}$ \\
    AurA & $\underline{.99}_{\pm.00}$ & $.96_{\pm.00}$ & $.96_{\pm.00}$ & $\mathbf{.00}_{\pm.00}$ & $1.0_{\pm.03}$ & $\mathbf{1.0}_{\pm.00}$ & $.68_{\pm.00}$ & $.98_{\pm.00}$ & $\mathbf{.00}_{\pm.00}$ & $1.04_{\pm.01}$ \\
    \hspace{5pt}\includegraphics[width=6pt]{figures_new/subdirectory_arrow_right.png} SAE & $\underline{.99}_{\pm.00}$ & $1.0_{\pm.00}$ & $.95_{\pm.00}$ & $\mathbf{.00}_{\pm.00}$ & $1.1_{\pm.00}$ & $.99_{\pm.00}$ & $.99_{\pm.00}$ & $.96_{\pm.00}$ & $\mathbf{.00}_{\pm.00}$ & $0.98_{\pm.00}$ \\
    PCA & $.84_{\pm.06}$ & $\underline{.79}_{\pm.13}$ & $.82_{\pm.05}$ & $.40_{\pm.52}$ & $2.3_{\pm.42}$ & $.98_{\pm.00}$ & $\underline{.51}_{\pm.07}$ & $.94_{\pm.01}$ & $\mathbf{.00}_{\pm.00}$ & $1.29_{\pm.07}$ \\
    LAT & $.68_{\pm.05}$ & $.90_{\pm.01}$ & $.64_{\pm.05}$ & $\underline{.12}_{\pm.10}$ & $1.1_{\pm.07}$ & $.54_{\pm.04}$ & $.72_{\pm.03}$ & $.51_{\pm.10}$ & $\underline{.20}_{\pm.13}$ & $0.98_{\pm.08}$ \\
    \hline
    \end{tblr}
\end{table}

\begin{table}[h]
    \centering
    \small
    \caption{
    \textbf{Evaluating steering representations on ImageNet-C \emph{Shot Noise} and \emph{Glass Blur} corruptions SigLIP.}
    The best method (row) in each evaluation metric (column) is in \textbf{bold} and the second best is \underline{underlined}.
    }
    \label{tbl:results_imagenet_c_siglip}
    \vspace{0.5em}
    \begin{tblr}{
      colspec = {lcccccccccc},
      column{1} = {leftsep=0pt},
      vline{2,7} = {dashed},
      hline{8,17} = {dashed},
      row{5,9,11,13,16} = {bg=black!10},
      rowsep=1.5pt,
      colsep = 2.3pt
    }
    & \SetCell[c=5]{c} \textbf{Shot Noise} & & & & & \SetCell[c=5]{c} \textbf{Glass Blur} & & & & \\
    \textbf{Method} & AUC ($\uparrow$) & MS ($\downarrow$) & F1 ($\uparrow$) & CD ($\downarrow$) & SD ($\downarrow$) & AUC ($\uparrow$) & MS ($\downarrow$) & F1 ($\uparrow$) & CD ($\downarrow$) & SD ($\downarrow$) \\
    \hline
    Linear SVM & $\mathbf{1.0}_{\pm.00}$ & $.99_{\pm.00}$ & $\mathbf{1.0}_{\pm.00}$ & $\mathbf{.00}_{\pm.00}$ & $1.1_{\pm.02}$ & $\mathbf{1.0}_{\pm.00}$ & $\underline{.69}_{\pm.01}$ & $\mathbf{1.0}_{\pm.00}$ & $\mathbf{.00}_{\pm.01}$ & $1.0_{\pm.03}$ \\
    LR & $\mathbf{1.0}_{\pm.00}$ & $.99_{\pm.00}$ & $\mathbf{1.0}_{\pm.00}$ & $\mathbf{.00}_{\pm.00}$ & $.99_{\pm.02}$ & $\mathbf{1.0}_{\pm.00}$ & $.75_{\pm.00}$ & $\mathbf{1.0}_{\pm.00}$ & $\mathbf{.00}_{\pm.00}$ & $.97_{\pm.04}$ \\
    \hspace{5pt}\includegraphics[width=6pt]{figures_new/subdirectory_arrow_right.png} SAE & $\mathbf{1.0}_{\pm.00}$ & $\mathbf{.92}_{\pm.05}$ & $\mathbf{1.0}_{\pm.00}$ & $\mathbf{.00}_{\pm.00}$ & $1.1_{\pm.03}$ & $\mathbf{1.0}_{\pm.00}$ & $\mathbf{.50}_{\pm.11}$ & $.98_{\pm.01}$ & $\mathbf{.00}_{\pm.00}$ & $.96_{\pm.01}$ \\
    SAS & $\mathbf{1.0}_{\pm.00}$ & $.98_{\pm.00}$ & $\mathbf{1.0}_{\pm.00}$ & $\mathbf{.00}_{\pm.00}$ & $.99_{\pm.01}$ & $\mathbf{1.0}_{\pm.00}$ & $.70_{\pm.01}$ & $\mathbf{1.0}_{\pm.00}$ & $\mathbf{.00}_{\pm.00}$ & $.93_{\pm.02}$ \\
    S\&PTopK & $\mathbf{1.0}_{\pm.00}$ & $\underline{.97}_{\pm.02}$ & $\mathbf{1.0}_{\pm.00}$ & $\mathbf{.00}_{\pm.00}$ & $.97_{\pm.02}$ & $\mathbf{1.0}_{\pm.00}$ & $.81_{\pm.03}$ & $.99_{\pm.01}$ & $\mathbf{.00}_{\pm.00}$ & $.95_{\pm.00}$ \\
    DiffMean & $\mathbf{1.0}_{\pm.00}$ & $.99_{\pm.00}$ & $\mathbf{1.0}_{\pm.00}$ & $\mathbf{.00}_{\pm.00}$ & $\underline{.96}_{\pm.06}$ & $\mathbf{1.0}_{\pm.00}$ & $.76_{\pm.00}$ & $\mathbf{1.0}_{\pm.00}$ & $\mathbf{.00}_{\pm.00}$ & $.91_{\pm.02}$ \\
    \hspace{5pt}\includegraphics[width=6pt]{figures_new/subdirectory_arrow_right.png} SAE & $\mathbf{1.0}_{\pm.00}$ & $.99_{\pm.00}$ & $\mathbf{1.0}_{\pm.00}$ & $\mathbf{.00}_{\pm.00}$ & $.99_{\pm.05}$ & $\mathbf{1.0}_{\pm.00}$ & $.81_{\pm.00}$ & $\mathbf{1.0}_{\pm.00}$ & $\mathbf{.00}_{\pm.00}$ & $.88_{\pm.02}$ \\
    DiffMedian & $\mathbf{1.0}_{\pm.00}$ & $.99_{\pm.00}$ & $\mathbf{1.0}_{\pm.00}$ & $\mathbf{.00}_{\pm.00}$ & $\underline{.96}_{\pm.05}$ & $\mathbf{1.0}_{\pm.00}$ & $.77_{\pm.00}$ & $\mathbf{1.0}_{\pm.00}$ & $\mathbf{.00}_{\pm.00}$ & $.91_{\pm.00}$ \\
    \hspace{5pt}\includegraphics[width=6pt]{figures_new/subdirectory_arrow_right.png} SAE & $\mathbf{1.0}_{\pm.00}$ & $.98_{\pm.01}$ & $\mathbf{1.0}_{\pm.00}$ & $\mathbf{.00}_{\pm.00}$ & $\underline{.96}_{\pm.01}$ & $\mathbf{1.0}_{\pm.00}$ & $.81_{\pm.01}$ & $.99_{\pm.01}$ & $\mathbf{.00}_{\pm.00}$ & $\underline{.87}_{\pm.02}$ \\
    FastCAV & $\mathbf{1.0}_{\pm.00}$ & $.99_{\pm.00}$ & $\mathbf{1.0}_{\pm.00}$ & $\mathbf{.00}_{\pm.00}$ & $\underline{.96}_{\pm.06}$ & $\mathbf{1.0}_{\pm.00}$ & $.76_{\pm.00}$ & $\mathbf{1.0}_{\pm.00}$ & $\mathbf{.00}_{\pm.00}$ & $.91_{\pm.02}$ \\
    \hspace{5pt}\includegraphics[width=6pt]{figures_new/subdirectory_arrow_right.png} SAE & $\mathbf{1.0}_{\pm.00}$ & $.99_{\pm.00}$ & $\mathbf{1.0}_{\pm.00}$ & $\mathbf{.00}_{\pm.00}$ & $.99_{\pm.05}$ & $\mathbf{1.0}_{\pm.00}$ & $.81_{\pm.00}$ & $\mathbf{1.0}_{\pm.00}$ & $\mathbf{.00}_{\pm.00}$ & $.88_{\pm.02}$ \\
    PatCAV & $\mathbf{1.0}_{\pm.00}$ & $.99_{\pm.00}$ & $\mathbf{1.0}_{\pm.00}$ & $\mathbf{.00}_{\pm.00}$ & $\underline{.96}_{\pm.06}$ & $\mathbf{1.0}_{\pm.00}$ & $.76_{\pm.00}$ & $\mathbf{1.0}_{\pm.00}$ & $\mathbf{.00}_{\pm.00}$ & $.91_{\pm.02}$ \\
    AurA & $\mathbf{1.0}_{\pm.00}$ & $.99_{\pm.00}$ & $\mathbf{1.0}_{\pm.00}$ & $\mathbf{.00}_{\pm.00}$ & $\underline{.96}_{\pm.03}$ & $\mathbf{1.0}_{\pm.00}$ & $.77_{\pm.00}$ & $\mathbf{1.0}_{\pm.00}$ & $\mathbf{.00}_{\pm.00}$ & $\mathbf{.81}_{\pm.01}$ \\
    \hspace{5pt}\includegraphics[width=6pt]{figures_new/subdirectory_arrow_right.png} SAE & $\mathbf{1.0}_{\pm.00}$ & $.99_{\pm.00}$ & $\mathbf{1.0}_{\pm.00}$ & $\mathbf{.00}_{\pm.00}$ & $.98_{\pm.01}$ & $\mathbf{1.0}_{\pm.00}$ & $.84_{\pm.00}$ & $\mathbf{1.0}_{\pm.00}$ & $\mathbf{.00}_{\pm.00}$ & $.96_{\pm.00}$ \\
    PCA & $\mathbf{1.0}_{\pm.00}$ & $.99_{\pm.00}$ & $\mathbf{1.0}_{\pm.00}$ & $\mathbf{.00}_{\pm.00}$ & $\mathbf{.93}_{\pm.04}$ & $\mathbf{1.0}_{\pm.00}$ & $.80_{\pm.01}$ & $\mathbf{1.0}_{\pm.00}$ & $\mathbf{.00}_{\pm.00}$ & $.91_{\pm.04}$ \\
    LAT & $.55_{\pm.06}$ & $.98_{\pm.01}$ & $.75_{\pm.05}$ & $.36_{\pm.23}$ & $1.6_{\pm.07}$ & $.50_{\pm.05}$ & $.80_{\pm.03}$ & $.73_{\pm.04}$ & $.60_{\pm.72}$ & $1.5_{\pm.16}$ \\
    \hline
    \end{tblr}
\end{table}

%%%%%%%%%%

\clearpage
\subsection{Counteranimals Results}
\label{app:counteranimal_results}

In prior benchmarks such as CelebA and ISIC, models can exploit spurious correlations between concepts and class labels, effectively using concepts as shortcuts for classification. This behavior arises due to strong class–concept co-occurrence in the training data.
In the CounterAnimal dataset, this shortcut phenomenon is further amplified. Here, specific animals not only co-occur with particular backgrounds but are deterministically associated with them. 
As a result, models may tend to strongly overfit to background cues rather than learning the underlying object representations, leading to severe bias and poor generalization to novel contexts.
To better assess debiasing in this setting, we modify the evaluation protocol as follows.

First, we train a classifier on the original training set and evaluate it on the unmodified test set. Next, we repeat the procedure using data in which all identified concepts have been orthogonalized from the training set. 
This setup aims to eliminate the model’s reliance on spurious concept-based signals, beyond simply orthogonalizing concepts during inference.
The results are summarized in Table~\ref{tbl:results_counteranimals_downstream}.
The first three columns report standard CAV-based metrics used throughout this work. The columns ``Diff Acc'' and ``Diff F1'' denote the performance differences between models trained on orthogonalized data and those trained on the original data.
Overall, most methods yield improvements in downstream classification performance after debiasing. 
However, perfect concept separability---as achieved by Linear SVM and Logistic Regression---does not necessarily translate into the largest gains in classification accuracy or F1 score.
This strengthens our assumption that concept detectability alone is insufficient for effective debiasing.

\begin{table}[h]
    \centering
    \small
    \caption{
    \textbf{Evaluating debiasing on CLIP for Counteranimal dataset.} 
    Columns ``Diff Acc'' and ``Diff F1'' refer to the difference in the classification task between the model that operated on data orthogonalized to all concepts, and the model using the original data.
    }
    \label{tbl:results_counteranimals_downstream}
    \vspace{0.5em}
    \begin{tblr}{
      colspec = {lcccccc},
      column{1} = {leftsep=0pt},
      vline{2} = {dashed},
      hline{7,16} = {dashed},
      row{4,8,10,12,15} = {bg=black!10},
      rowsep=1.5pt
    }
    \textbf{Method} & AUC ($\uparrow$) & CCR ($\uparrow$) & MS ($\downarrow$) & Diff Acc ($\uparrow$) & Diff F1 ($\uparrow$) & Time (s) \\
    \hline
    Linear SVM & $\mathbf{1.0}_{\pm.00}$ & $\mathbf{1.0}_{\pm.00}$ & $\mathbf{-.08}_{\pm.04}$ & $1.9$ & $1.9$ & $1.4_{\pm.59}$ \\
    LR & $\mathbf{1.0}_{\pm.00}$ & $\mathbf{1.0}_{\pm.00}$ & $\underline{-.07}_{\pm.03}$ & $1.6$ & $\underline{2.0}$ & $1.2_{\pm.48}$ \\
    \hspace{5pt}\includegraphics[width=6pt]{figures_new/subdirectory_arrow_right.png} SAE & $\underline{.99}_{\pm.01}$ & $\underline{.99}_{\pm.01}$ & $-.01_{\pm.08}$ & $\underline{2.1}$ & $1.8$ & $2.1_{\pm.50}$ \\
    SAS & $.91_{\pm.07}$ & $.95_{\pm.07}$ & $.07_{\pm.03}$ & $.24$ & $.48$ & $5.2_{\pm3.8}$ \\
    S\&PTopK & $.97_{\pm.03}$ & $\underline{.99}_{\pm.01}$ & $.16_{\pm.08}$ & $1.1$ & $.81$ & $.88_{\pm.25}$ \\
    DiffMean & $.97_{\pm.02}$ & $.98_{\pm.02}$ & $.00_{\pm.08}$ & $1.4$ & $1.5$ & $\mathbf{.00}_{\pm.01}$ \\
    \hspace{5pt}\includegraphics[width=6pt]{figures_new/subdirectory_arrow_right.png} SAE & $.97_{\pm.02}$ & $.98_{\pm.02}$ & $.00_{\pm.08}$ & $.94$ & $.84$ & $.26_{\pm.09}$ \\
    DiffMedian & $.96_{\pm.02}$ & $.98_{\pm.02}$ & $.00_{\pm.08}$ & $\underline{2.2}$ & $\mathbf{2.3}$ & $\underline{.01}_{\pm.01}$ \\
    \hspace{5pt}\includegraphics[width=6pt]{figures_new/subdirectory_arrow_right.png} SAE & $.87_{\pm.10}$ & $.95_{\pm.06}$ & $.15_{\pm.16}$ & $-.56$ & $-1.2$ & $.26_{\pm.07}$ \\
    FastCAV & $.97_{\pm.02}$ & $.98_{\pm.02}$ & $.00_{\pm.08}$ & $1.4$ & $1.5$ & $\mathbf{.00}_{\pm.00}$ \\
    \hspace{5pt}\includegraphics[width=6pt]{figures_new/subdirectory_arrow_right.png} SAE & $.97_{\pm.02}$ & $.98_{\pm.02}$ & $.00_{\pm.08}$ & $.92$ & $.81$ & $.25_{\pm.07}$ \\
    PatCAV & $.97_{\pm.02}$ & $.98_{\pm.02}$ & $.00_{\pm.08}$ & $1.4$ & $1.5$ & $\mathbf{.00}_{\pm.00}$ \\
    AurA & $.97_{\pm.02}$ & $\underline{.99}_{\pm.01}$ & $.33_{\pm.05}$ & $\mathbf{2.4}$ & $\underline{2.2}$ & $.90_{\pm.03}$ \\
    \hspace{5pt}\includegraphics[width=6pt]{figures_new/subdirectory_arrow_right.png} SAE & $.96_{\pm.03}$ & $.96_{\pm.04}$ & $.70_{\pm.01}$ & $1.4$ & $1.5$ & $28_{\pm.91}$ \\
    PCA & $.73_{\pm.09}$ & $.90_{\pm.11}$ & $.58_{\pm.14}$ & $-.66$ & $-.71$ & $.08_{\pm.04}$ \\
    LAT & $.60_{\pm.10}$ & $.87_{\pm.11}$ & $.59_{\pm.24}$ & $-1.8$ & $-1.8$ & $.05_{\pm.05}$ \\
    \hline
    \end{tblr}
\end{table}

%%%%%%%%%%%%%%%%%
\clearpage
\subsection{Critical Difference Analysis}
\label{app:cdd}
\begin{figure}[h]
    \centering
    \includegraphics[width=\linewidth]{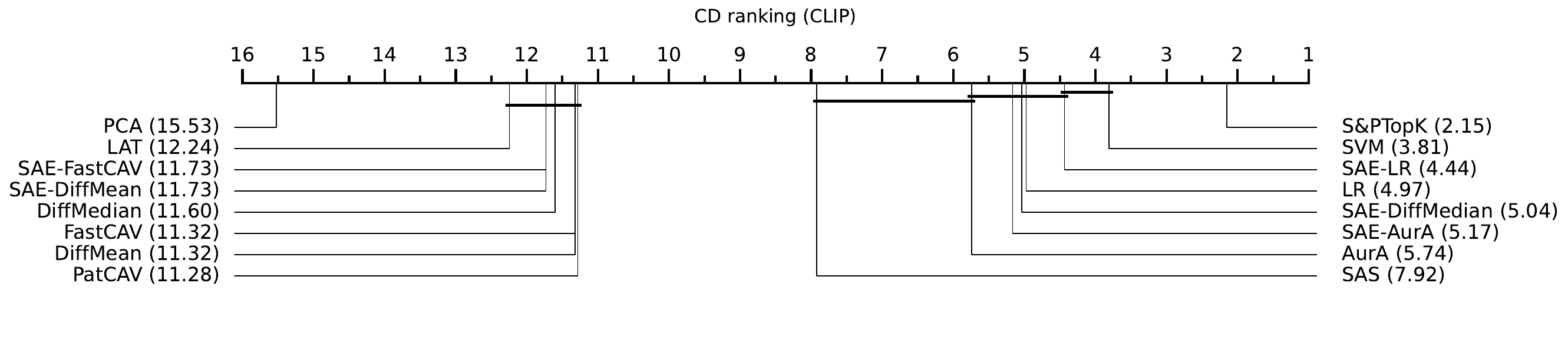}
    \medskip
    \includegraphics[width=\linewidth]{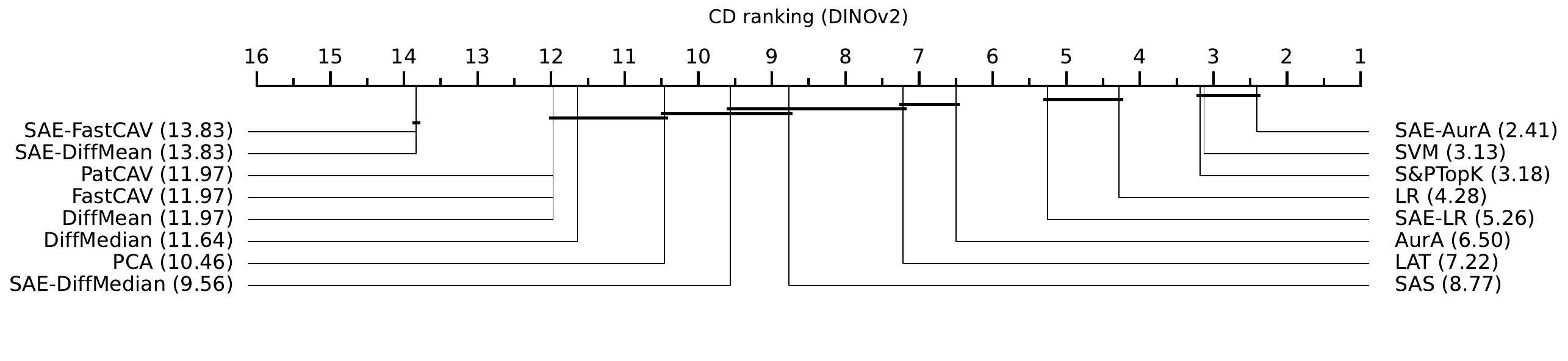}
    \medskip
    \includegraphics[width=\linewidth]{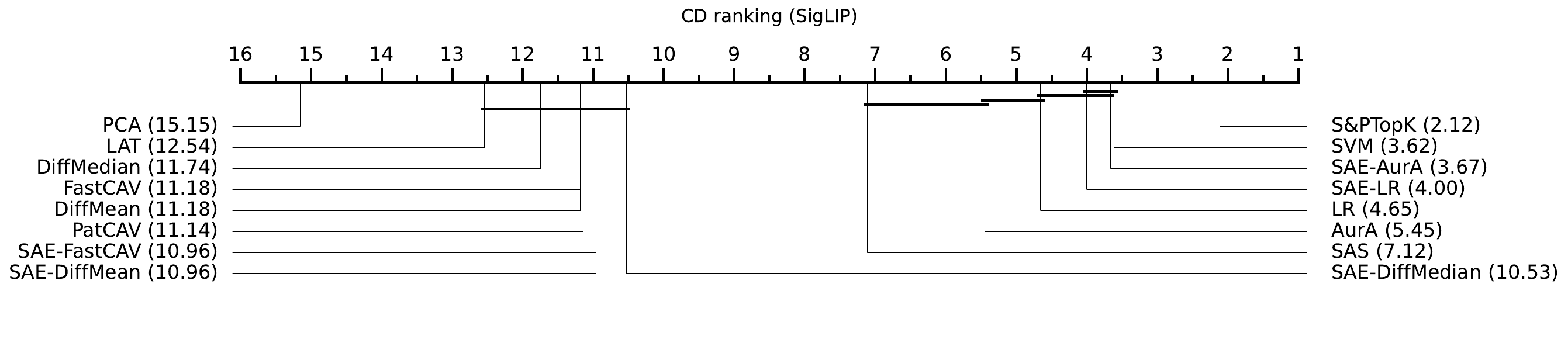}
    \caption{\textbf{Critical difference analysis of collateral damage.} 
    The plots display results for CLIP (top), DINOv2 (middle), and SigLIP (bottom). 
    Across all backbones, optimization-based methods consistently outperform statistical and environment-specific approaches, with the notable exception of AurA and SAE-AurA.}
    \label{fig:cdd_heatmap}
\end{figure}

In \Cref{fig:cdd_heatmap}, we analyze the critical differences between CAV methods regarding collateral damage (CD) on the CelebA dataset. The results show that S\&PTopK achieves the lowest CD on two out of the three backbones. The slightly worse results on DINOv2 may be attributed to the SAE's weak performance on this specific backbone (see \Cref{tab:sae_performance}). SVM and SAE-AurA follow closely as top-performing methods. Excluding AurA, we see that on all backbones, statistical methods often lag behind optimization-based approaches.

%%%%%%%%%%
\clearpage
\section{Extended Ablation on Sample Efficiency}
\label{app:extended_ablation_samples}
\begin{figure}[h]
    \centering
    \begin{subfigure}{\linewidth}
        \centering
        \includegraphics[width=1\linewidth]{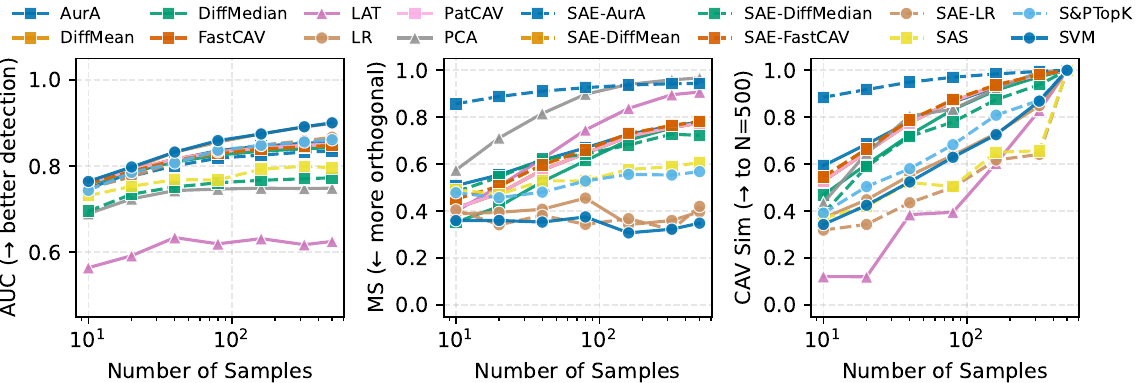}
        \end{subfigure}
    \par\bigskip % 
    \begin{subfigure}{\linewidth}
        \centering
        \includegraphics[width=1\linewidth]{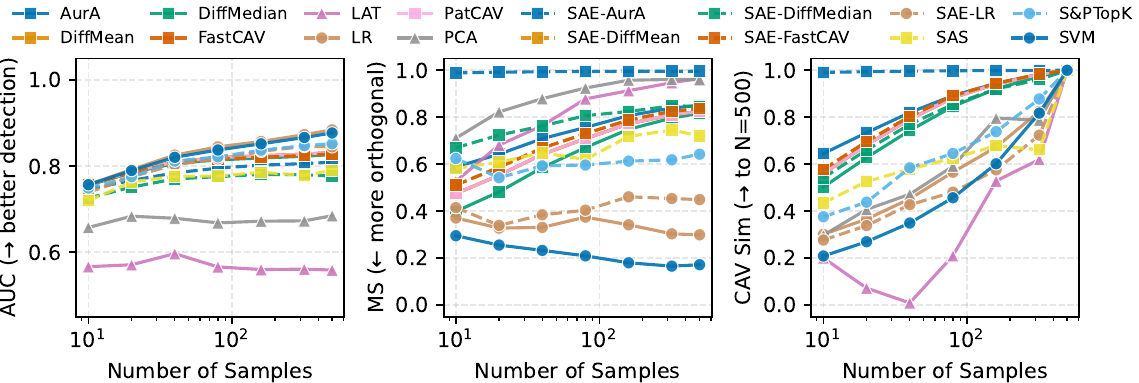}
    \end{subfigure}
    \par\bigskip % 
    \begin{subfigure}{\linewidth}
        \centering
        \includegraphics[width=1\linewidth]{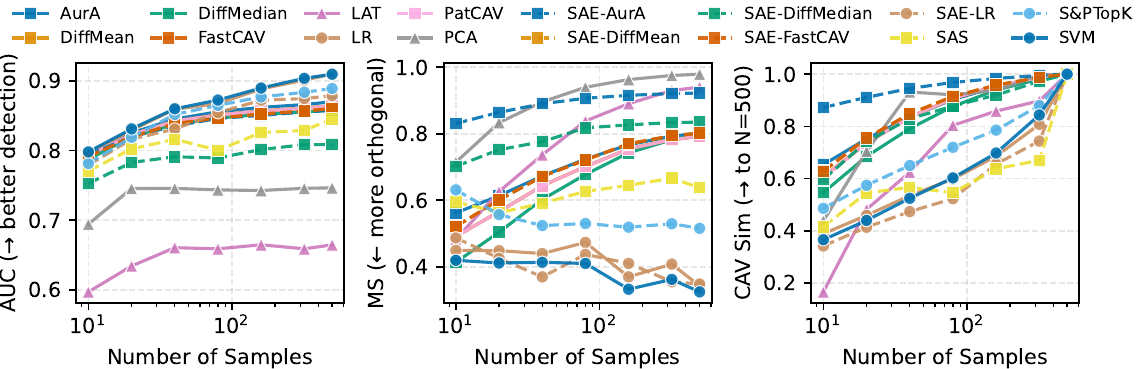}
    \end{subfigure}
    \caption{
    \textbf{Extended sample efficiency ablations.} 
    Complementing \Cref{fig:samples_ablations}, we report metrics as a function of training sample size for CelebA with CLIP (\emph{top}), CelebA with DINOv2 (\emph{middle}), and CelebA with SigLIP (\emph{bottom}). 
    Convergence trends remain consistent across all backbones, with the exception of SAE-AurA on DINOv2. The DINOv2 SAE-AurA has already converged at just 10 samples, demonstrating only marginal subsequent changes, as indicated by an increase in AUC.}
    \label{fig:extended_samples}
\end{figure}

\begin{figure}[h]
    \centering
    \begin{subfigure}{\linewidth}
        \centering
        \includegraphics[width=1\linewidth]{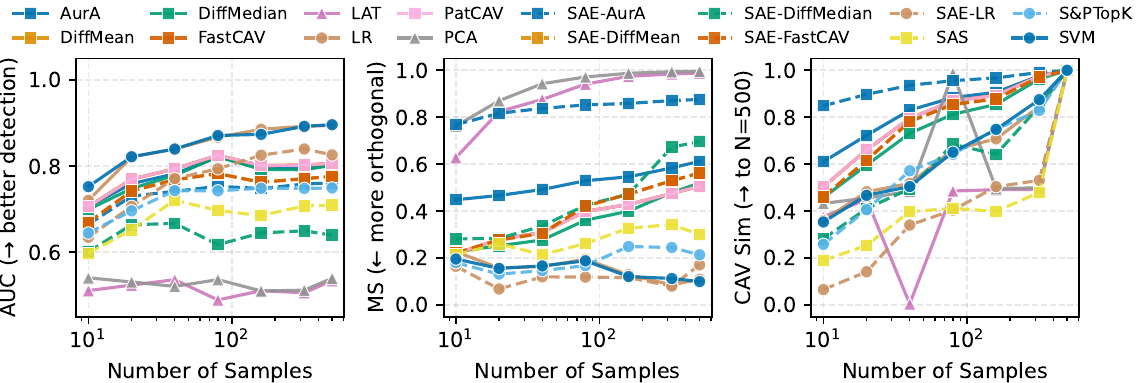}
        \end{subfigure}
    \par\bigskip % 
    \begin{subfigure}{\linewidth}
        \centering
        \includegraphics[width=1\linewidth]{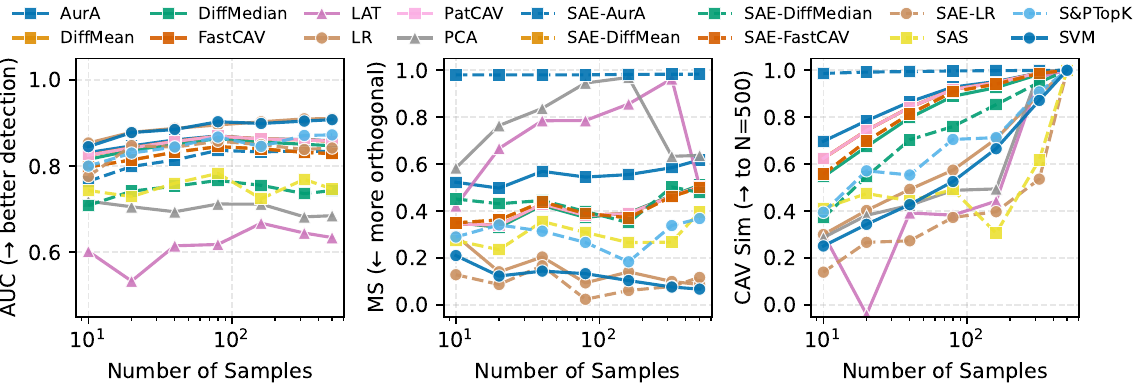}
    \end{subfigure}
    \par\bigskip % 
    \begin{subfigure}{\linewidth}
        \centering
        \includegraphics[width=1\linewidth]{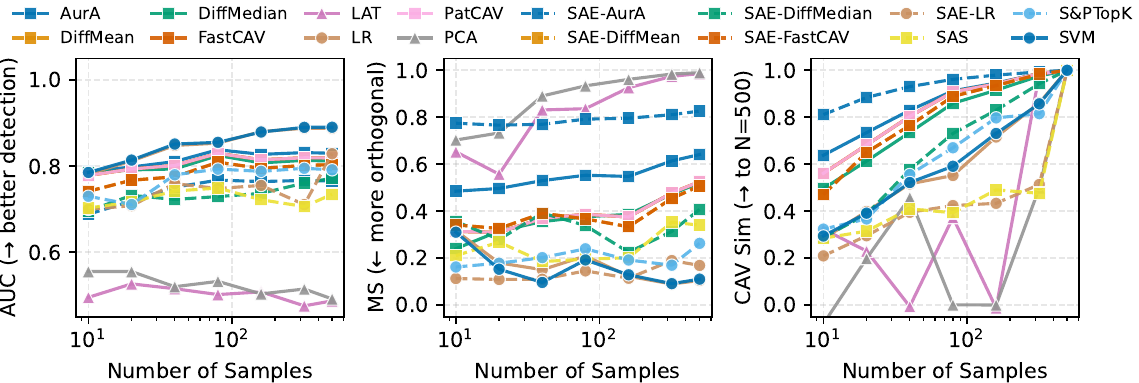}
    \end{subfigure}
    \caption{
    \textbf{Sample efficiency ablations for ISIC.} 
    We report metrics as a function of training sample size for: ISIC with CLIP (\emph{top}); ISIC with DINOv2 (\emph{middle}), and ISIC with SigLIP (\emph{bottom}).}
    \label{fig:samples_ablations_isic}
\end{figure}

Extending the ablations from \Cref{sec:results}, \Cref{fig:extended_samples} presents sample efficiency ablations for additional backbones on CelebA (CLIP, DINOv2) and \Cref{fig:samples_ablations_isic} the ISIC dataset on all backbones. These results reinforce our primary finding: methods consistently cluster into three distinct groups: optimization-based ({$\bullet$}), statistical ({$\blacksquare$}), and ``environment-specific'' ({$\blacktriangle$}), regardless of the backbone or dataset.

\begin{figure}[h]
    \centering
    \includegraphics[width=\linewidth]{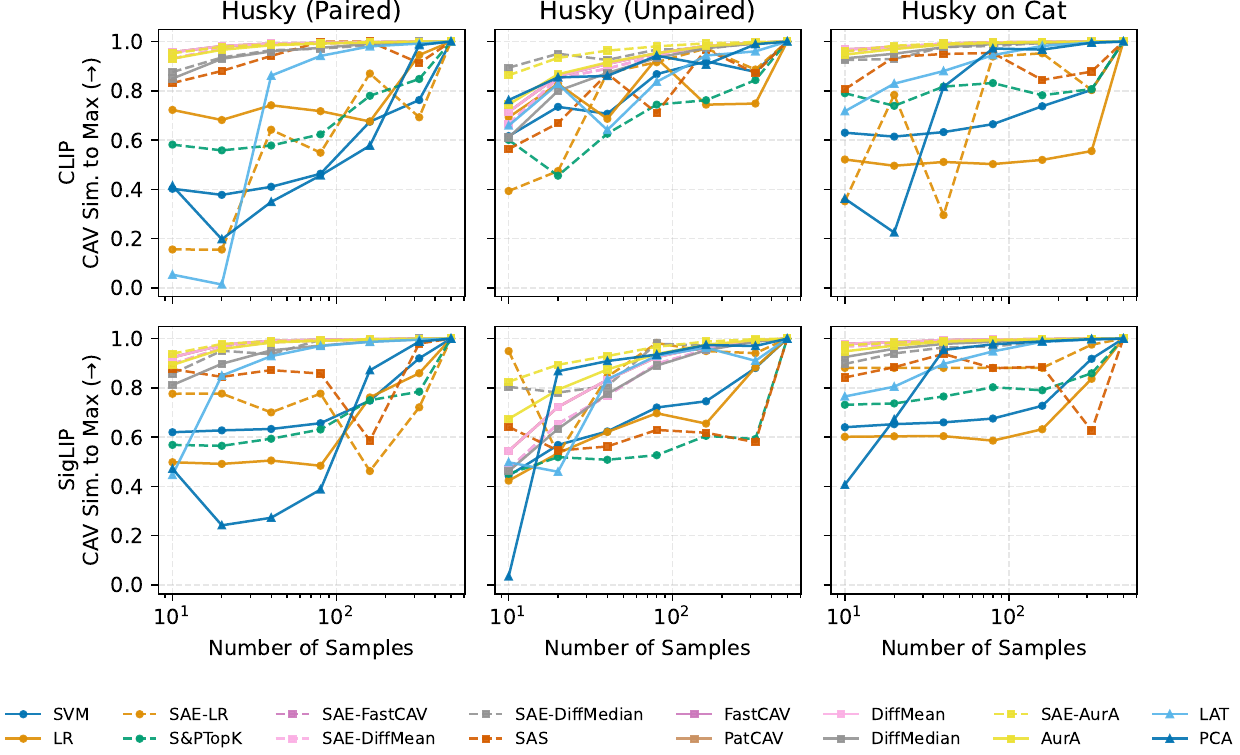}
    \caption{\textbf{Sample efficiency on ImageNet-W (husky/cat Watermark).} We compare SigLIP (top row) and CLIP (bottom row) across three setups: watermark on all classes, unpaired samples, and paired (cat-only) watermarks. Results indicate that paired setups require the least amount of data to converge. Notably, optimization-based methods ({$\bullet$}) utilize the full range of samples to optimize, whereas statistical ({$\blacksquare$}) methods stabilize much earlier.}
    \label{fig:imagenetw_samples}
\end{figure}

Furthermore, \Cref{fig:imagenetw_samples} examines the ImageNet-W dataset (\emph{husky} watermark on \emph{cat} class scenario). 
We observe that data requirements depend heavily on the specific training data setup. Paired setups require significantly fewer samples to establish a stable direction. 
Consistently, statistical methods demonstrate superior data efficiency, saturating early, whereas optimization-based methods continue to refine their trajectory up to the maximum $N=500$ samples.

%%%%%%%%%%%%%%%%%%%%
\clearpage
\section{Performance Comparison Between Linear and Nonlinear Methods}
\label{app:linear_vs_nonlinear}
This section validates the hypothesis from Finding~\ref{finding1}: that top-performing linear CAVs like linear SVM and LR (but not SAE-LR) achieve separation accuracy (AUC) comparable to complex nonlinear probes. We visualize the win-ratio between linear and nonlinear methods for the CelebA and ISIC datasets. (Waterbirds is excluded due to insufficient concept count, and synthetic datasets are excluded due to ceiling effects where AUC $\approx$ 1.0).

\begin{figure}[h]
    \centering
    \includegraphics[width=1.0\linewidth]{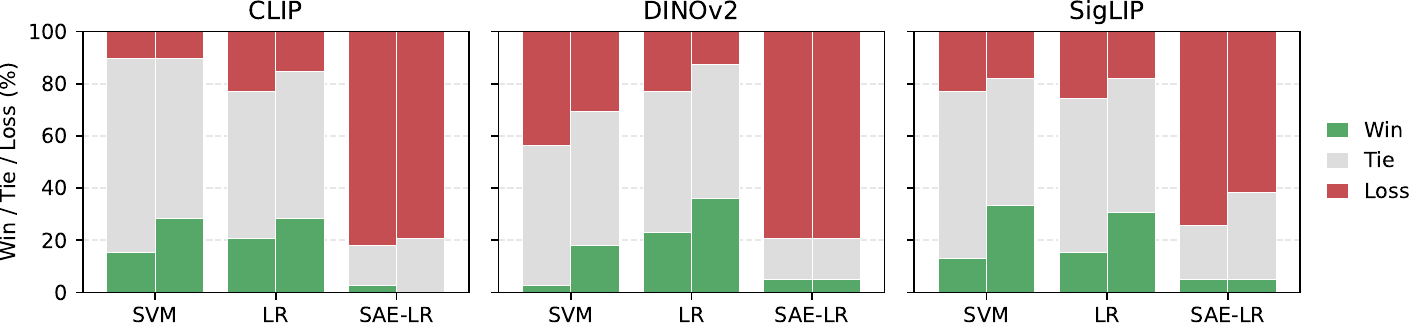}
    \caption{\textbf{Win-rate comparison on CelebA: linear CAVs vs. nonlinear probes.} We present the win-rates of linear methods (linear SVM, LR, SAE-LR) against nonlinear baselines \emph{left} TabPFN), (\emph{right} SVM-RBF. Results are rounded to 2 significant digits. While linear SVM and LR maintain competitive performance (high tie/win rates), SAE-LR demonstrates a high loss ratio. Among nonlinear probes, TabPFN generally outperforms SVM-RBF.}
    \label{fig:celeba_winrates}
\end{figure}

\begin{figure}[h]
    \centering
    \includegraphics[width=1.0\linewidth]{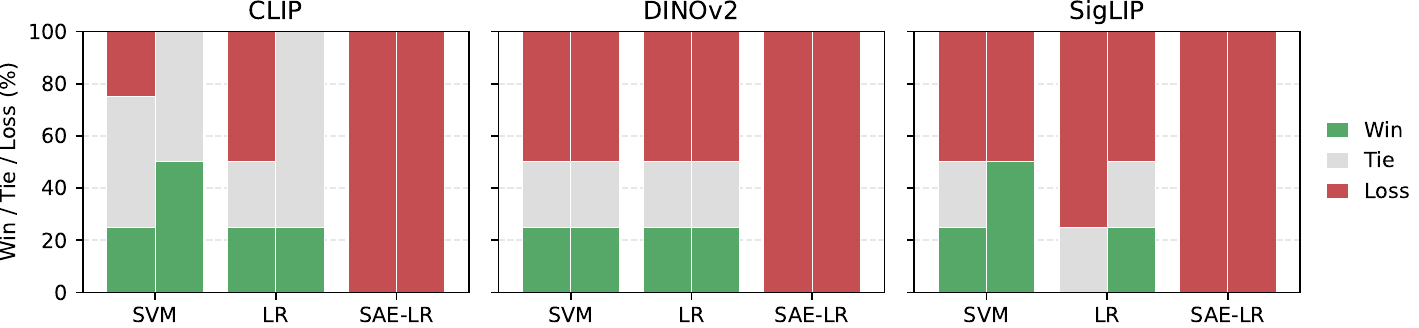}
    \caption{\textbf{Win-rate comparison on ISIC: linear CAVs vs. nonlinear probes.} Similar to CelebA Figure~\ref{fig:celeba_winrates}, we report win-rates against \emph{left} TabPFN, \emph{right} SVM-RBF on all backbones. Linear SVM and LR perform comparably to nonlinear methods, whereas SAE-LR significantly underperforms.}
    \label{fig:isic_winrates}
\end{figure}

Although raw win rates in \Cref{fig:celeba_winrates,fig:isic_winrates} suggest a marginal advantage for nonlinear probes (TabPFN, SVM-RBF), a two-sided Wilcoxon signed-rank test (\Cref{tab:pvalues_tblr}), paired by concept, shows that for linear SVM and LR the null hypothesis of no difference cannot be rejected ($p > 0.05$). In contrast, SAE-LR on ISIC frequently yields a significant result ($p < 0.05$), and its win rates indicate that SAE-based adaptation consistently underperforms relative to linear probes.

\begin{table}[t]
    \centering
    \caption{\textbf{Statistical significance of performance differences (two-sided Wilcoxon signed-rank test).}
    We compare the AUC distributions of each method against nonlinear probes (TabPFN and SVM-RBF). Values are $p$-values; $p < 0.05$ (\textbf{bold}) indicates a statistically significant difference. 
    High $p$-values for SVM and LR suggest no significant gap relative to nonlinear probes, whereas SAE-LR frequently on CelebA shows significant underperformance (see also win rates in \Cref{fig:celeba_winrates,fig:isic_winrates}).}
    \label{tab:pvalues_tblr}
    \vspace{0.5em}
    \begin{tabular}{l l cc cc}
    \toprule
    \multirow{2}{*}{\textbf{Backbone}} & \multirow{2}{*}{\textbf{CAV}} & \multicolumn{2}{c}{\textbf{ISIC}} & \multicolumn{2}{c}{\textbf{CelebA}} \\
    \cmidrule(lr){3-4} \cmidrule(lr){5-6}
     & & TabPFN & SVM-RBF & TabPFN & SVM-RBF \\
    \midrule
    \multirow{3}{*}{SigLIP} 
        & SVM    & 0.250 & 0.875 & 0.512 & 0.054 \\
        & LR     & 0.125 & 1.000 & 0.634 & 0.104 \\
        & SAE-LR & 0.125 & 0.125 & \textbf{0.000} & \textbf{0.000} \\
    \midrule
    \multirow{3}{*}{DINOv2} 
        & SVM    & 0.625 & 0.875 & \textbf{0.000} & 0.329 \\
        & LR     & 0.875 & 0.875 & 0.595 & \textbf{0.001} \\
        & SAE-LR & 0.125 & 0.125 & \textbf{0.000} & \textbf{0.000} \\
    \midrule
    \multirow{3}{*}{CLIP} 
        & SVM    & 0.875 & 0.875 & 0.070 & \textbf{0.007} \\
        & LR     & 0.875 & 0.875 & 0.918 & 0.127 \\
        & SAE-LR & 0.125 & 0.125 & \textbf{0.000} & \textbf{0.000} \\
    \bottomrule
    \end{tabular}
\end{table}

%%%%%%%%%%%%%%%%%%%%
\clearpage
\section{Extended Metrics Analysis}
\label{app:metric_analysis}
In this section, we provide detailed visualizations and statistical analyses to substantiate Finding~\ref{finding4}, demonstrating the relationships and the critical distinctiveness between vector and steering metrics.

\begin{figure}[h]
    \centering
    \begin{subfigure}{0.33\columnwidth}
        \centering
        \includegraphics[width=\linewidth]{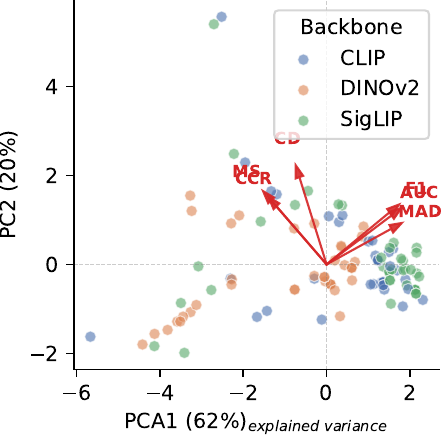}
    \end{subfigure}
    \begin{subfigure}{0.33\columnwidth}
        \centering
        \includegraphics[width=\linewidth]{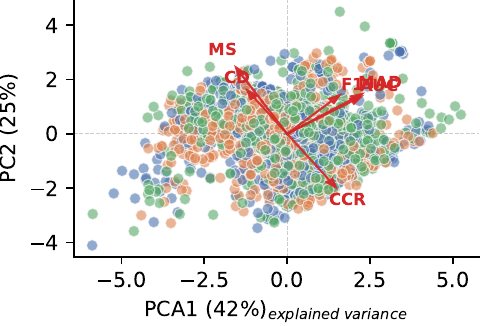}
    \end{subfigure}\hfill
    \begin{subfigure}{0.33\columnwidth}
        \centering
        \includegraphics[width=\linewidth]{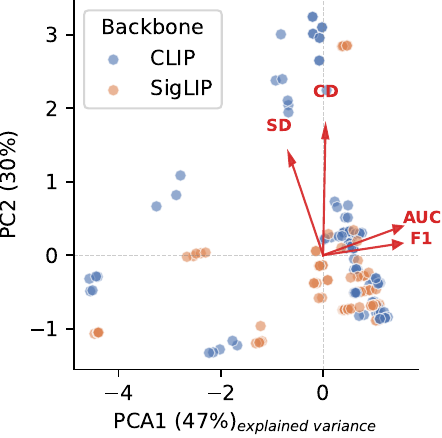}
    \end{subfigure}
    \caption{\textbf{PCA biplots of metric correlations.} We project all evaluation metrics onto the first two principal components across all backbones for Waterbirds (left), CelebA (middle), and ImageNet-W (right). The arrows indicate the direction of each metric. Predictive metrics (AUC, F1, MAD) consistently overlap, while MS and CCR form distinct correlated directions.}
    \label{fig:biplot}
\end{figure}

\begin{figure}[h]
    \centering
    \includegraphics[width=\linewidth]{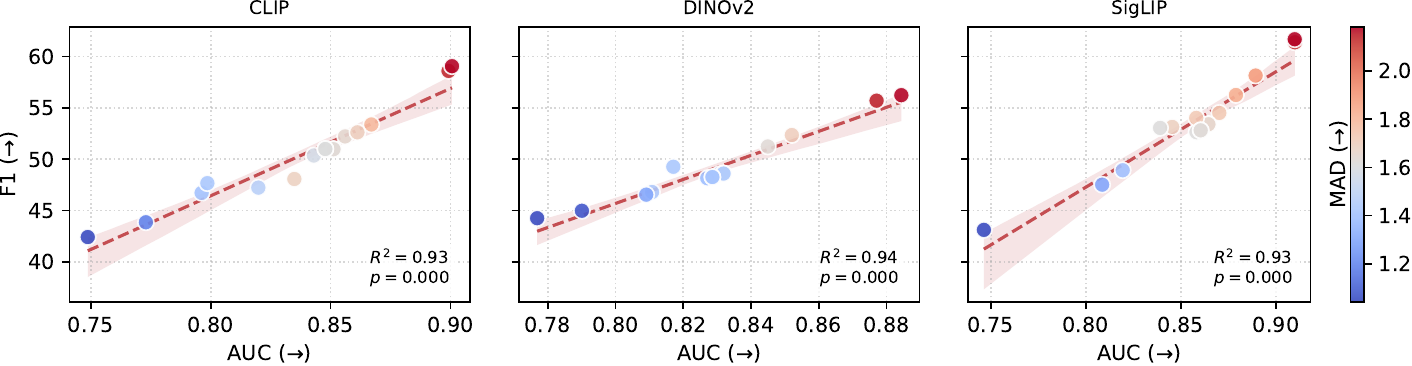}
    \caption{\textbf{Correlation between AUC and F1 on CelebA.} We report linear regression statistics ($R^2$ and $p$-value) for CLIP (left), DINOv2 (middle), and SigLIP (right). All backbones show a statistically significant positive correlation ($p \approx 0, R^2 > 0.9$), confirming that intrinsic concept detection (AUC) reliably predicts downstream performance (F1).}
    \label{fig:celeba_auc_f1_mad}
\end{figure}

\begin{figure}[h]
    \centering
    \includegraphics[width=\linewidth]{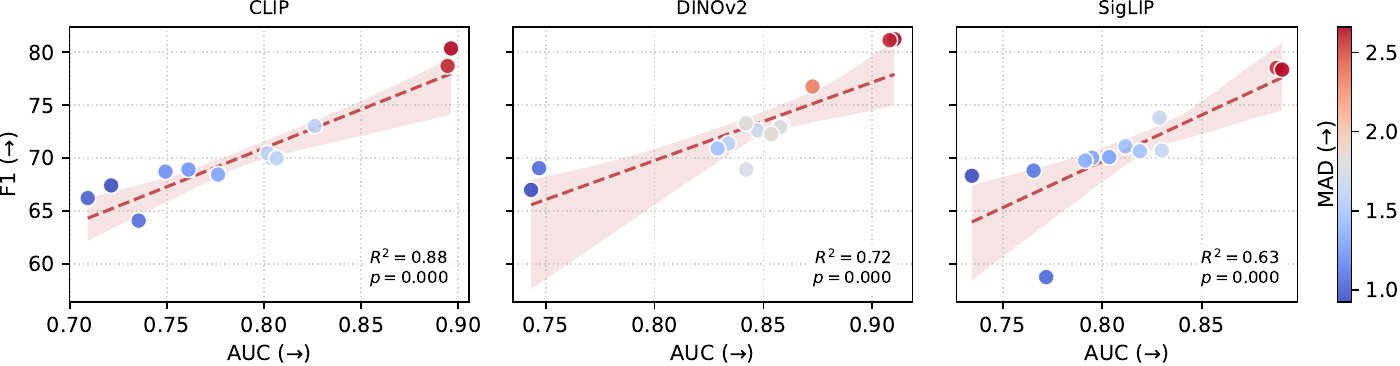}
    \caption{\textbf{Correlation between AUC and F1 on ISIC.} 
    Across all backbones, we observe a strong correlation ($p \approx 0, R^2 > 0.6$) between the F1 and AUC metrics. MAD shows a clear and strong correlation with both of AUC and F1, which is why we exclude it from our results.}
    \label{fig:isic_auc_f1_mad}
\end{figure}

\begin{figure}[h]
    \centering
    \includegraphics[width=\linewidth]{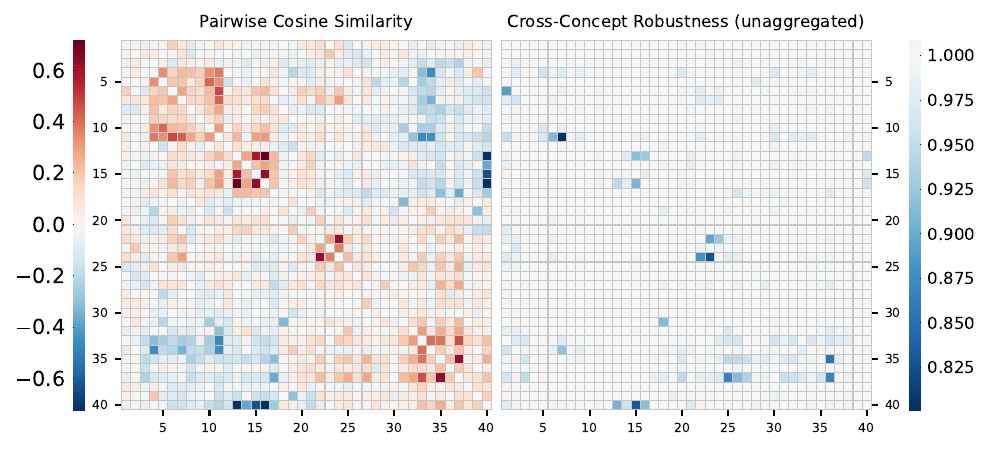}
    %{figures/heatmap_celeba.pdf}
    \caption{\textbf{Heatmap comparison of MS vs CCR on CelebA (using linear SVM).} Left: Pairwise Cosine Similarity between concept vectors. Right: Pairwise Cross Concept Robustness (CCR). We choose linear SVM as CAVs from this method have the highest disentanglement. Although the CCR heatmap is sparser, the non-zero off-diagonal elements largely overlap with high-similarity regions in the left plot. This indicates that high vector similarity (entanglement) drives low robustness (interference), justifying the use of Maximum Similarity (MS) as a worst-case proxy for robustness.}
    \label{fig:ccr_ms_heatmap}
\end{figure}

\begin{figure}[h]
    \centering
    \begin{subfigure}{0.4\columnwidth}
        \centering
        \includegraphics[width=\columnwidth]{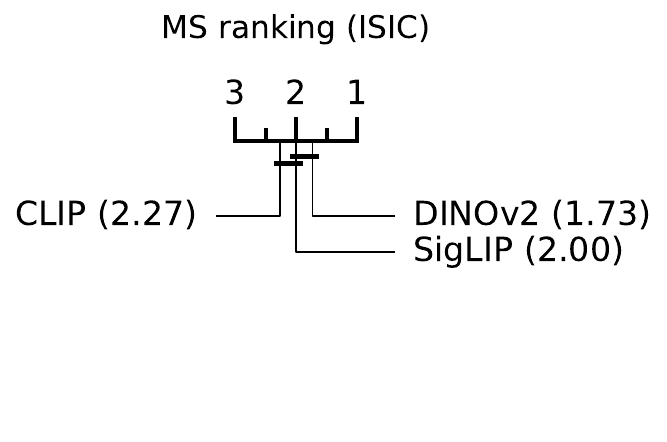}
        \vspace{-4.5em}
        %{figures/ms_ranks_isic_1.pdf}
    \end{subfigure}\hfill
    \begin{subfigure}{0.4\columnwidth}
        \centering
        \includegraphics[width=\columnwidth]{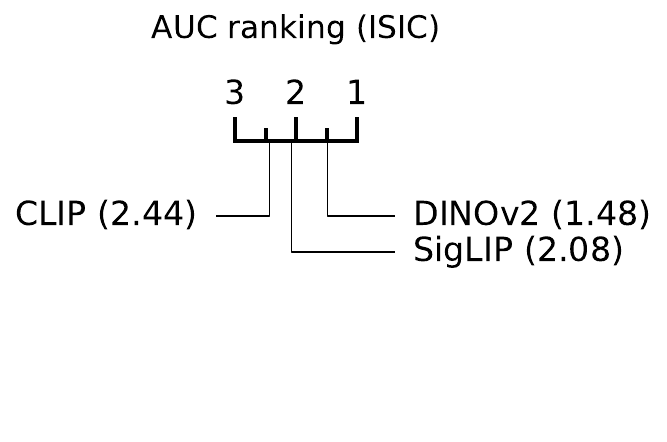}
        \vspace{-4.5em}
        %{figures/auc_ranks_isic_1.pdf}
    \end{subfigure}
    \medskip
    \begin{subfigure}{0.4\columnwidth}
        \centering
        \includegraphics[width=\columnwidth]{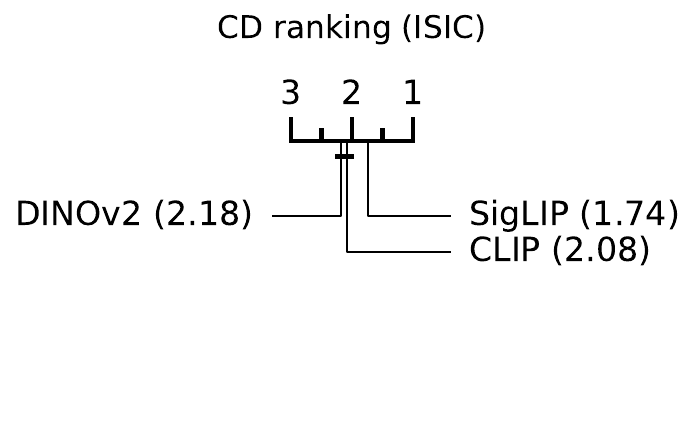}
        \vspace{-4.5em}
        %{figures/cd_ranks_isic_1.pdf}
    \end{subfigure}\hfill
    \begin{subfigure}{0.4\columnwidth}
        \centering
        \includegraphics[width=\columnwidth]{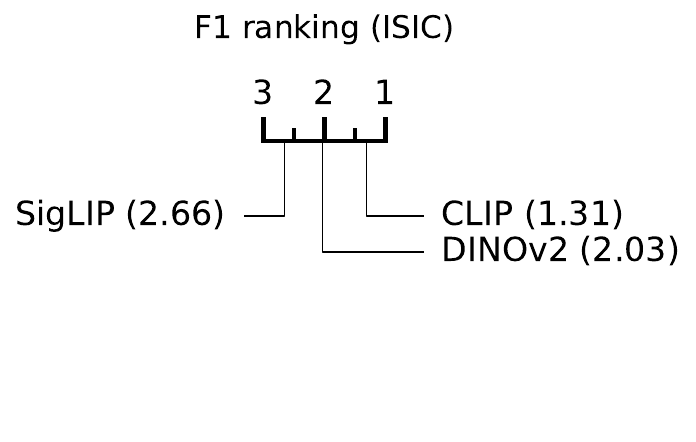}
        \vspace{-4.5em}
        %{figures/f1_ranks_isic_1.pdf}
    \end{subfigure}
    \caption{\textbf{Critical difference analysis of backbone performance on ISIC} (Wilcoxon's test, $\alpha=0.1$). 
    We report average rank comparisons for concept similarity (MS, top left), detection performance (AUC, top right), collateral damage (CD, bottom left), and F1 detection score (bottom right). 
    On this dataset, we observe that the optimal backbone varies depending on the specific metric being evaluated. For MS, the performance of all backbones is statistically indistinguishable.}
    \label{fig:cd_plots_isic}
\end{figure}

\begin{figure}[h]
    \centering
    \begin{subfigure}[b]{0.4\linewidth}
        \centering
        \includegraphics[width=\linewidth]{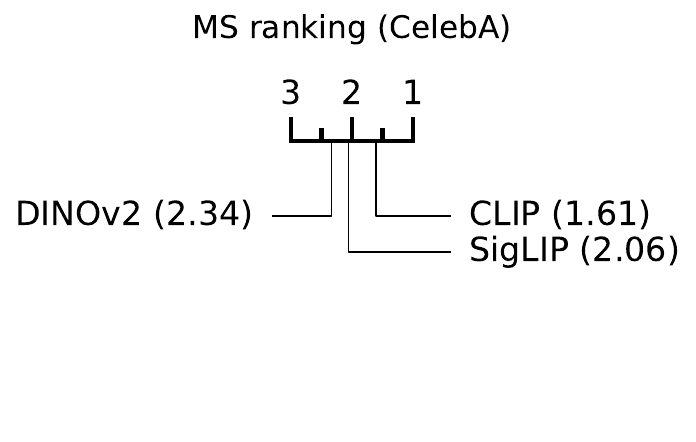}
        \vspace{-4.5em}
        %{figures/ms_ranks_waterbirds.pdf}
    \end{subfigure}
    \hfill
    \begin{subfigure}[b]{0.4\linewidth}
        \centering
        \includegraphics[width=\linewidth]{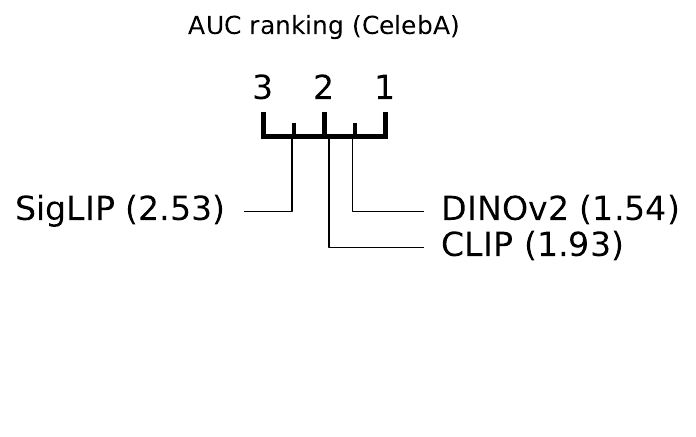}
        \vspace{-4.5em}
        %{figures/rocauc_ranks.pdf}
    \end{subfigure}
    \medskip
    \begin{subfigure}[b]{0.4\linewidth}
        \centering
        \includegraphics[width=\linewidth]{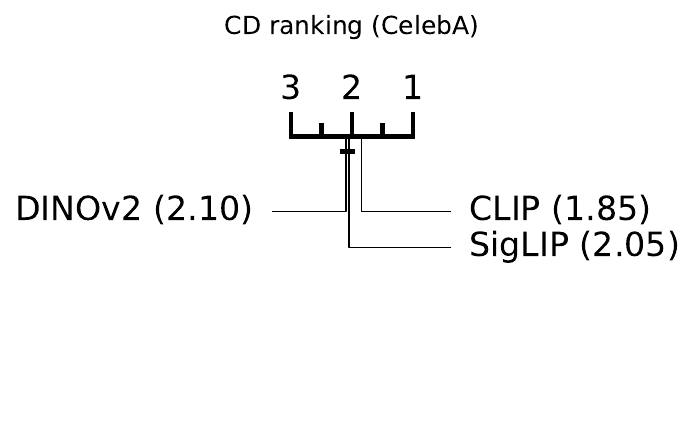}
        \vspace{-4.5em}
        %{figures/cd_ranks_waterbirds.pdf}
    \end{subfigure}
    \hfill
    \begin{subfigure}[b]{0.4\linewidth}
        \centering
        \includegraphics[width=\linewidth]{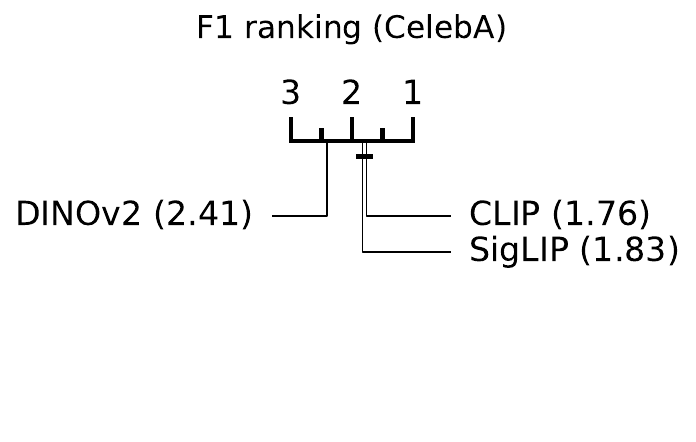}
        \vspace{-4.5em}
        %{figures/f1_ranks_waterbirds.pdf}
    \end{subfigure}
    \caption{\textbf{Critical difference analysis of backbone performance on CelebA} (Wilcoxon's test, $\alpha=0.1$). 
    We report average rank comparisons for concept similarity (MS, top left), detection performance (AUC, top right), collateral damage (CD, bottom left), and F1 detection score (bottom right). 
    Although SigLIP is the top-performing backbone for overall concept detection on both vector and downstream tasks, CLIP performs significantly better on MS.}
    \label{fig:cd_plots_celeba_extended}
\end{figure}

To analyze the interplay among the metrics defined in \Cref{sec:evaluation}, we employ a PCA biplot visualization in \Cref{fig:biplot}, extending the previous analysis to cover multiple datasets. 
This allows us to project the multi-dimensional metric space onto two principal axes for CelebA, Waterbirds, and ImageNet-W. 
From the biplots, we find 5 key observations:
\begin{itemize}
    \item \textbf{Direction 1: predictive fidelity (AUC, F1, MAD).} The biplots reveal that MAD, F1 score, and AUC are highly collinear. This strong correlation implies that MAD provides redundant information when AUC or F1 is available. We further confirm this in \Cref{fig:celeba_auc_f1_mad,fig:isic_auc_f1_mad}, which demonstrate a high correlation between these metrics across all backbones.
    
    \item \textbf{Direction 2: disentanglement \& robustness (MS, CCR).} We observe a strong correlation between Cross Concept Robustness (CCR) and Maximum Similarity (MS). For the CelebA task (\Cref{fig:biplot}, middle), these vectors point in opposite directions, which is intuitive as lower MS (more orthogonal vectors) should correspond to higher robustness (higher CCR) against interference from other concepts. In the Waterbirds task (\Cref{fig:biplot}, left), CCR and MS appear positively correlated (same direction). As detailed in \Cref{tbl:results_waterbirds_clip} this is an artifact of the dataset's concept binary structure: the ``land'' and ``water'' concepts are essentially inverse vectors. Thus, high negative similarity to one implies lower robustness.
    
    \item \textbf{Orthogonality of steering metrics.} Crucially, side-effect metrics (CD and CCR) and Disparity appear orthogonal to the predictive cluster (AUC/F1) \Cref{fig:main,fig:biplot}. This lack of correlation, further supported by the quantitative results in \Cref{tbl:results_dinov2,tbl:results_watermark}, confirms that high predictive performance does not imply steering effects.

    \item \textbf{Impact of CAVs similarity and their robustness.} 
    Strong similarity between two CAVs (dark red or blue color in \Cref{fig:ccr_ms_heatmap}, left) is required for observing a loss in predictive performance after orthogonalization of one CAV against the other (dark blue in \Cref{fig:ccr_ms_heatmap}, right).
    Notably, the Cross-Concept Robustness is not symmetric, i.e., one CAV might be more robust to orthogonalization than the other in a pair. 

    \item \textbf{Impact of vision models.}
    The overall performance of the CAV extraction methods is strongly influenced by the vision model that serves as the backbone for all computations. The superiority of a backbone depends on the specific task, which is presented in Figures~\ref{fig:cd_plots_isic}--\ref{fig:cd_plots_celeba_extended}. Notably, DINOv2 serves as the best backbone for the ISIC task, although SigLIP incurs lower collateral damage. For CelebA, SigLIP is the dominant model, trailing CLIP only on the MS metric.
    
\end{itemize}

\clearpage
\section{Broader Impact}
\label{app:broader_impact}

\ourbenchmarkx{} contributes to interpretability, controllability, and fairness research on vision models, motivated by high-stakes settings where representation-level shortcuts cause harm: dermatology classifiers latching onto acquisition artifacts rather than lesion morphology~\citep{bissoto2020debiasing,wu2023discover}, demographic correlates contaminating face attribute predictions~\citep{sagawa2020distributionally}, and contrastive backbones relying on context rather than the object itself~\citep{wang2024sober}.
By explicitly measuring \emph{collateral damage} and \emph{cross-concept robustness}, our benchmark discourages a common failure mode in this area: declaring success on separability or downstream accuracy alone, while leaving unrelated capabilities silently degraded.

Steering is, however, fundamentally dual-use. 
Any method that subtracts a concept can, with a sign flip or tuned coefficient, amplify or inject one: propagating sensitive attributes, weakening refusal behaviors, or covertly altering predictions along demographic or commercial axes. 
We further caution that strong linear erasure performance does not guarantee that information has been removed from the representation; it removes only the linearly accessible component along the chosen direction~\citep{belrose2023leace,ravfogel2020null}, and our own Findings~3--4 show that vector and downstream metrics can disagree in non-obvious ways. 
\textsc{SwordBench} scores are evidence about a method, not a fairness certificate for any deployed system.
To mitigate these risks, we release \textsc{SwordBench} as an open benchmark and invite contributions of new concepts, datasets, and metrics.
We prominently report negative results to discourage selective benchmarking.

\end{document}